\documentclass[10pt]{article}
\usepackage{svg}
\pdfoutput=1

\usepackage[most]{tcolorbox}\tcbuselibrary{breakable}
\usepackage{xcolor}
\usepackage{listings}
\usepackage{paralist}
\usepackage{threeparttable}
\usepackage[T1]{fontenc}       %
\usepackage[utf8]{inputenc}    %
\usepackage{CJKutf8}           %
\usepackage{wrapfig}  %
\usepackage{graphicx} %

\tcbset{
  aibox/.style={
    width=\textwidth,
    top=0pt, bottom=0pt, left=5pt, right=5pt,
    colback=white,
    colframe=black,
    colbacktitle=black,
    enhanced,
    center,
    attach boxed title to top left={yshift=-0.1in,xshift=0.15in},
    boxed title style={boxrule=0pt,colframe=white,},
  }
}
\newtcolorbox{AIbox}[2][]{aibox,title=#2,#1}

\usepackage{adjustbox}
\usepackage{algpseudocodex}
\usepackage{alltt}
\usepackage{amsmath, amsfonts, amssymb, mathtools}
\usepackage{amsthm}

\usepackage{array} %
\usepackage{arydshln}
\usepackage{bold-extra}
\usepackage{booktabs}
\usepackage[small]{caption}
\usepackage{color}
\usepackage{comment}
\usepackage{contour}
\usepackage{diagbox}
\usepackage{enumerate}
\usepackage[shortlabels]{enumitem}
\usepackage{epstopdf}
\usepackage{fancyvrb}
\usepackage{float}
\usepackage[bottom]{footmisc}
\usepackage{fvextra}
\usepackage{cellspace}
\usepackage[most]{tcolorbox}
\usepackage[utf8]{inputenc}
\usepackage[greek, english]{babel}
\usepackage{alphabeta}

\usepackage[
  margin=0.83in,
  headheight=12pt,
  headsep=25pt,
  includefoot,
  footskip=30pt,
]{geometry}
\usepackage{graphicx}
\usepackage[final]{hyperref}
\usepackage{inconsolata}
\usepackage[utf8]{inputenc}
\usepackage{listings}
\usepackage{makecell}
\usepackage{makeidx}
\usepackage{mathtools}
\usepackage{mdframed}
\usepackage{multirow}
\usepackage{MnSymbol,wasysym}
\usepackage[numbers]{natbib}

\usepackage{rotating}
\usepackage{tablefootnote} %
\usepackage{soul}
\usepackage{subcaption}
\usepackage{tabularx} 
\usepackage[most]{tcolorbox}
\usepackage{thm-restate}
\usepackage{tikz}
\usepackage{pgfplots}
\usepackage{filecontents}
\usepackage{todonotes}
\usepackage{ulem}
\usepackage{url}
\usepackage{xr}
\usepackage{xcolor}
\usepackage{xspace}
\usepackage{algorithm}
\usepackage{tcolorbox}
\usepackage{graphicx}

\newcommand{\modelwithoutspace}{\textsf{Phi-4-reasoning}}
\newcommand{\modelRL}{\textsf{Phi-4-reasoning-plus}\xspace}
\newcommand{\model}{\textsf{Phi-4-reasoning}\xspace}
\newcommand{\phifour}{\textsf{Phi-4}\xspace}

\newcommand{\oone}{\textsf{o1}\xspace}
\newcommand{\oonemini}{\textsf{o1-mini}\xspace}
\newcommand{\othree}{\textsf{o3-mini}\xspace}

\newcommand{\claude}{\textsf{Claude 3.7 Sonnet}\xspace}
\newcommand{\rone}{\textsf{DeepSeek-R1}\xspace}
\newcommand{\ronedistill}{\textsf{DeepSeek-R1-Distill-Llama-70B}\xspace}
\newcommand{\ronefourteenb}{\textsf{DeepSeek-R1-Distil-Qwen-14B}\xspace}
\newcommand{\geminiflash}{\textsf{Gemini 2 Flash Thinking}\xspace}
\newcommand{\gptfouro}{\textsf{GPT-4o}\xspace}

\definecolor{blush}{rgb}{1.0, 0.0, 0.5}

\definecolor{commentcolour}{rgb}{0.3,0.7,0.2}
\definecolor{backcolour}{rgb}{0.98,0.98,0.98}
\lstdefinelanguage{markdown}{
    comment=[l]{\#},
    morestring=[s]{```}{```},
    commentstyle=\color{commentcolour}\bfseries,
    stringstyle=\color{blue},
    basicstyle=\scriptsize\ttfamily,
    showstringspaces=false,
    breaklines=true,
    breakautoindent=false,
    breakindent=0pt,
    backgroundcolor=\color{backcolour},
}
\lstdefinestyle{mystyle}{
    morekeywords={self},
    basicstyle=\scriptsize\ttfamily,
    keywordstyle=\color{blue},
    commentstyle=\color{commentcolour}\bfseries,
    breaklines=true,
    breakautoindent=false,
    showstringspaces=false,
    backgroundcolor=\color{backcolour},
    stringstyle=\color{red},
}
\lstdefinelanguage{PythonPlus}[]{Python}{
  alsoother={@},
  morekeywords=[1]{,as,assert,nonlocal,with,yield,self,True,False,None} %
  morekeywords=[2]{,__init__,__add__,__mul__,__div__,__sub__,__call__,__getitem__,__setitem__,__eq__,__ne__,__nonzero__,__rmul__,__radd__,__repr__,__str__,__get__,__truediv__,__pow__,__name__,__future__,__all__,}, %
  morekeywords=[3]{,object,type,isinstance,copy,deepcopy,zip,enumerate,reversed,list,set,len,dict,tuple,range,xrange,append,execfile,real,imag,reduce,str,repr,}, %
  morekeywords=[4]{,Exception,NameError,IndexError,SyntaxError,TypeError,ValueError,OverflowError,ZeroDivisionError,}, %
  morekeywords=[5]{,ode,fsolve,sqrt,exp,sin,cos,arctan,arctan2,arccos,pi, array,norm,solve,dot,arange,isscalar,max,sum,flatten,shape,reshape,find,any,all,abs,plot,linspace,legend,quad,polyval,polyfit,hstack,concatenate,vstack,column_stack,empty,zeros,ones,rand,vander,grid,pcolor,eig,eigs,eigvals,svd,qr,tan,det,logspace,roll,min,mean,cumsum,cumprod,diff,vectorize,lstsq,cla,eye,xlabel,ylabel,squeeze,}, %
}
\usepackage{fontawesome5}

\definecolor{user}{HTML}{F2EFE7}  %
\definecolor{model}{HTML}{81E7AF} %
\setuptodonotes{inline}
\usetikzlibrary{matrix}
\usetikzlibrary{shapes,calc,positioning}

\global

\tcbset{
  aibox/.style={
    width=\textwidth,
    top=0pt, bottom=0pt, left=5pt, right=5pt,
    colback=white,
    colframe=black,
    colbacktitle=black,
    enhanced,
    center,
    attach boxed title to top left={yshift=-0.1in,xshift=0.15in},
    boxed title style={boxrule=0pt,colframe=white,},
  }
}

\newtcbox{\tokenhilightbox}[1][]{
    on line,
    boxrule=0.5pt,
    colframe=black,
    colback=white,
    boxsep=1.5pt,
    left=0pt,
    right=0pt,
    top=0pt,
    bottom=0pt,
}

\definecolor{aigold}{RGB}{244,210, 1} 
\definecolor{aigreen}{RGB}{210,244,211} 

\sethlcolor{aigreen}

\definecolor{aired}{RGB}{255,180,181}

\newtcbox{\mybox}[1][green]{on line,
arc=0pt,outer arc=0pt,colback=#1!10!white,colframe=#1!50!black,
boxsep=0pt,left=0pt,right=0pt,top=0pt,bottom=0pt,
boxrule=0pt,bottomrule=0pt,toprule=0pt}

\definecolor{forestgreen}{RGB}{34,139,34}

\newcommand{\eg}{\emph{e.g., }}
\newcommand{\ie}{\emph{i.e., }}

\makeatletter
\newcommand{\dynscriptsize}{\check@mathfonts\fontsize{\sf@size}{\z@}\selectfont}
\makeatother
\newcommand\textunderset[2]{%
  \leavevmode
  \vtop{\offinterlineskip
    \halign{%
      \hfil##\hfil\cr %
      \strut#2\cr
      \noalign{\kern-.3ex}
      \dynscriptsize\strut#1\cr
    }%
  }%
}

\contourlength{0.8pt}

\begin{document}
\definecolor{salmon}{HTML}{FF6363}

\title{\textbf{\model Technical Report}}

\author{ Marah Abdin \quad Sahaj Agarwal \quad Ahmed Awadallah \quad Vidhisha Balachandran \\ Harkirat Behl \quad Lingjiao Chen \quad Gustavo de Rosa \quad Suriya Gunasekar \quad Mojan Javaheripi \\ Neel Joshi \quad Piero Kauffmann \quad Yash Lara \quad 
Caio César Teodoro Mendes  \quad Arindam Mitra \\  Besmira Nushi  \quad
Dimitris Papailiopoulos  \quad Olli Saarikivi \quad Shital Shah \\ Vaishnavi Shrivastava \quad 
Vibhav Vineet \quad Yue Wu \quad Safoora Yousefi \quad Guoqing Zheng\footnote{alphabetical order, correspondences to phi-research@microsoft.com} \vspace{0.4em} \\ Microsoft}

\date{}
\maketitle
\vspace{-2em}
\begin{abstract}
We introduce \modelwithoutspace, a 14-billion parameter reasoning model that achieves strong performance on complex reasoning tasks. 
Trained via supervised fine-tuning of \phifour{} on carefully curated set of ``teachable'' prompts--selected for the right level of complexity and diversity--and reasoning demonstrations generated using \othree, \model generates detailed reasoning chains that effectively leverage inference-time compute. We further develop \modelRL{}, a variant enhanced through a short phase of outcome‑based reinforcement learning that offers higher performance by generating longer reasoning traces. Across a wide range of reasoning tasks, both models outperform 
 significantly larger open-weight models such as \ronedistill model and approach the performance levels of full \rone model. Our comprehensive evaluations span benchmarks in math and scientific reasoning, coding, algorithmic problem solving, planning, and spatial understanding. Interestingly, we observe a non-trivial transfer of improvements to general-purpose benchmarks as well.
In this report, we provide insights into our training data, our training methodologies, and our evaluations. We show that the benefit of careful data curation for supervised fine-tuning (SFT) extends to reasoning language models, and can be further amplified by reinforcement learning (RL). Finally, our evaluation points to opportunities for improving how we assess the performance and robustness of reasoning models.

\end{abstract}

\vspace{-0.3em}
\begin{figure}[ht]
     \centering
     \captionsetup{width=.9\textwidth}
     \includegraphics[width=0.9\textwidth]{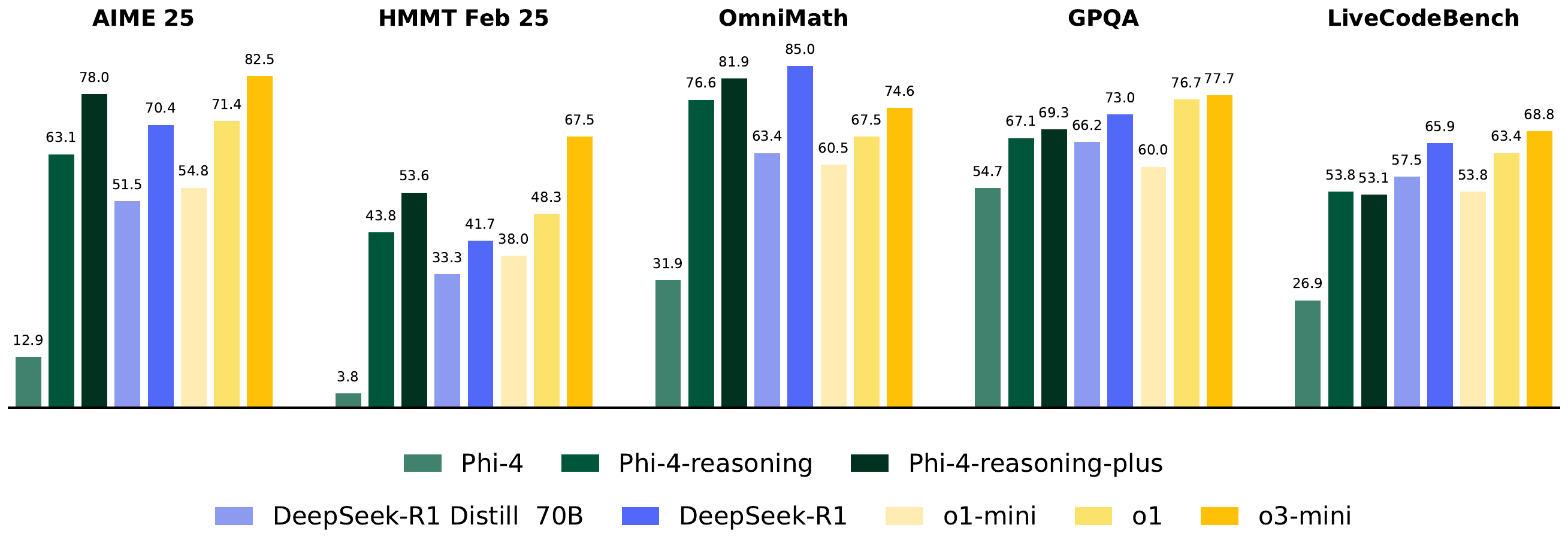}
     \vspace{-0.7em}
     \caption{
     \model performance across representative reasoning benchmarks spanning mathematical (HMMT, AIME 25, OmniMath), scientific (GPQA), and coding (LiveCodeBench 8/24-1/25) domains. We illustrate the performance gains from reasoning-focused post-training of \phifour{} via \model (SFT) and \modelRL{} (SFT+RL), alongside: open-weight models from DeepSeek including \rone (671B Mixture-of-Experts) and its distilled dense variant \ronedistill, and OpenAI’s proprietary frontier models \oone{} and \othree. \model and \modelRL consistently outperform the base model \phifour{} and demonstrate competitive performance against substantially larger and state-of-the-art models. A more comprehensive evaluation is provided in Section \ref{sec:eval}.
     \label{fig:phi4-main}}
\end{figure}

\newpage
\section{Introduction}
\vspace{-0.1cm}
Reasoning-focused large language models (LLMs) are trained to perform complex tasks that demand multi-step decomposition, internal reflection, and exploration of multiple problem-solving strategies. Recent reasoning models exhibit these capabilities \textit{via} a form of inference-time scaling, wherein a greater computational effort is dynamically allocated during inference for more complex tasks, resulting in improved performance in domains such as mathematical problem solving, logical reasoning, and answering questions that require a deeper contextual understanding. 
Several frontier models now have reasoning-optimized variations, including OpenAI's o1, o3(-mini)~\cite{O3mini}, Anthropic's Claude-3.7-Sonnet-Thinking~\cite{Claude37Sonnet}, Google's Gemini-2-Thinking and Gemini-2.5-Flash~\cite{GeminiFlash}, and DeepSeek-AI's DeepSeek-R1~\cite{guo2025deepseek}. In parallel, multiple open-weight reasoning models have been introduced to the research community~\cite{yang2024qwen2,team2025kimi,guan2025rstar,guo2025deepseek}.

DeepSeek-R1 \cite{guo2025deepseek} also pioneered a family of distilled open-weight models, demonstrating that the advanced reasoning capabilities of large language models can be distilled into smaller models through supervised fine-tuning. Follow-up work~\cite{ye2025emergence,li2025limrrlscaling,yu2025dapoopensourcellmreinforcement,dang2025reinforcementlearningreasoningsmall} has shown that these smaller models can be further improved via reinforcement learning. These findings underscore the potential of combining supervised fine-tuning and reinforcement learning to develop efficient, high-performing small language models with strong reasoning ability. In this work, we curate a new family of small reasoning models by integrating these insights with a data-centric approach.

We present \modelwithoutspace, a 14-billion parameter model supervised fine-tuned on \phifour~\cite{abdin2024phi4technicalreport}, and \modelRL{} obtained by a further round of reinforcement learning. \model is trained on high-quality datasets with over 1.4M prompts and high-quality answers containing long reasoning traces generated using \othree{}. The prompts are specifically filtered to cover a range of difficulty levels and to lie at the boundary of the base model capabilities. The datasets used in supervised fine-tuning include topics in STEM (science, technology, engineering, and mathematics), coding, and safety-focused tasks. \modelRL{} is further trained with Reinforcement Learning on a small set of $\sim$6K high-quality math-focused problems with verifiable solutions.

\noindent{\textbf{Contributions:}}
We highlight the benefits of careful data curation and supervised fine-tuning (SFT) for reasoning language models. More specifically, we show the importance of the selection and filtering of prompts and responses, as well as the critical role of data mixture and training recipe. We detail this data and supervised finetuning  recipe that is at the core of the model in Sections~\ref{sec:data}-\ref{sec:sft}. Our approach aligns closely with data-centric methods of earlier Phi and Orca models~\cite{gunasekar2023textbooks,javaheripi2023phi,abdin2024phi,abdin2024phi4technicalreport,mukherjee2023orca, mitra2023orca, mitra2024agentinstruct}, demonstrating that meticulous data curation and high-quality synthetic datasets allow smaller models to compete with larger counterparts. Secondly, we are encouraged by the gains achieved through reinforcement learning (RL) and the potential of combining distillation/SFT and reinforcement learning. we plan to explore this area further especially for domains where SFT data is not available (Section~\ref{sec:rl}). Lastly, we conduct a comprehensive evaluation to assess the performance and robustness of our models. We note the need to establish more rigorous practices for evaluating reasoning models that account for the small size of the commonly used benchmarks and the inherent non-determinism exhibited by the models (Section~\ref{sec:eval})
 
We summarize the key observations on model performance below.

\vspace{-0.5cm}
\paragraph{Performance compared to other models:}  Despite their smaller size (14B parameters), the performance of both models is competitive with or exceeding much larger models across several benchmarks as shown in Figure~\ref{fig:phi4-main} and Figure~\ref{fig:all_in_one}. For example, they achieve better performance than o1-mini and \ronedistill at most benchmark including mathematical reasoning (AIME) and PhD-level questions (GPQA). They achieve performance comparable to the full \rone model (with 671-billion parameters) on \textit{AIME 2025} (the 2025 qualifier for the USA Math Olympiad)\footnote{AIME 2025 was released after the training data for \model was finalized and is thus contamination free. We also algorithmically decontaminate the training data against AIME 2024 and various other benchmarks used in this report (see Section~\ref{sec:data}).}. They also outperform \claude and \geminiflash on all tasks except \textit{GPQA} and \textit{Calendar Planning}. See Figures~\ref{fig:phi4-main} and~\ref{fig:all_in_one}.

\vspace{-0.5cm}
\paragraph{Performance on algorithmic problem solving:} We also test the models on multiple new reasoning benchmarks for algorithmic problem solving and planning, including \textit{3SAT} (3-literal Satisfiability Problem) and \textit{TSP} (Traveling Salesman Problem)~\cite{balachandran2025inferencetimescalingcomplextasks} for solving NP-hard problems~\citep{papadimitriou2003computational,hartmanis1982computers}, and \textit{BA-Calendar} for calendar planning task~\citep{butt2024benchagents}. These new tasks are nominally out-of-domain for the models as the training process did not intentionally target these skills, but the models show strong generalization to these tasks. 

\vspace{-0.5cm}
\paragraph{Improvement over the base model:} Both models \model and \modelRL present major improvements over the base model \phifour{} across a broad set of reasoning tasks, including math and scientific reasoning, coding, algorithmic problem solving, and planning. Notably, the models improve by over 50 percentage points in accuracy on math benchmarks (AIME 2025 and OmniMath) and by over 25 percentage points on coding (LiveCodeBench). Surprisingly, these models also improve by 30 to 60 percentage points on algorithmic and planning problems (TSP, 3SAT, and BA Calendar Planning), which demonstrates increased generalizability of reasoning skills to domains that were not directly targeted during supervised fine-tuning or reinforcement learning. See Figures~\ref{fig:phi4-main} and~\ref{fig:all_in_one}.

\vspace{-0.5cm}    
\paragraph{Improvement on general-purpose benchmarks:} Improvements on reasoning contribute to non-trivial and often large benefits on more general-purpose skills. For example, \modelRL is 22 points more accurate than \phifour{} at instruction following (IFEval), 16 points better in long-context question answering and reasoning (FlenQA), and 10 points better in ArenaHard which focuses on human preferences for chat-like interactions. We also observed that both models are modestly more accurate in detecting toxic language (Toxigen), with \model showing a  more balanced accuracy on detecting neutral vs. toxic content, which is desirable for content moderation and filtering applications. See Table~\ref{tab:results_general_purpose}.

\vspace{-0.5cm}
\paragraph{Thinking effort vs. accuracy tradeoffs:} The two models offer two different token length vs. accuracy tradeoffs. 
\modelRL{} has significant higher accuracy on math (which was emphasized during RL) but uses approximately 1.5× more tokens than \model{} on average. This difference is less pronounced on other reasoning domains like coding, planning, and spatial tasks, suggesting avenues for improving RL exploration and verification in broader task sets. 
See Figures~\ref{fig:all_in_one} and~\ref{fig:all_in_one_accuracy_tokens} on accuracy vs. token length tradeoffs.

\vspace{-0.5cm}
\paragraph{Comprehensive evaluations:} This report emphasizes the need for moving beyond single-score accuracy reporting on small-scale datasets, due to large accuracy variations across repeated runs~\citep{balachandran2024eureka,balachandran2025inferencetimescalingcomplextasks,hochlehnert2025sober}.  
While some variation is expected, it becomes problematic when aggregate accuracies are reported on a very small set of examples, such as in AIME. For all models including ours and those in the OpenAI and DeepSeek family, two runs of average-of-5 evaluations can differ significantly (by up to 5-10 percentage points on AIME), thus making side-by-side comparisons unreliable for all models. To address this issue, we report standard deviations across multiple runs, experiments on a higher number of generations (50-64) for AIME 2025 (see Figures~\ref{fig:aime_parallel_8324} and~\ref{fig:aime_50runs}), and report results on larger, challenging and non-saturated benchmarks.

\vspace{-0.5cm}

    \begin{wrapfigure}[14]{r}{0.4\linewidth}
\centering
    \includegraphics[width=\linewidth]{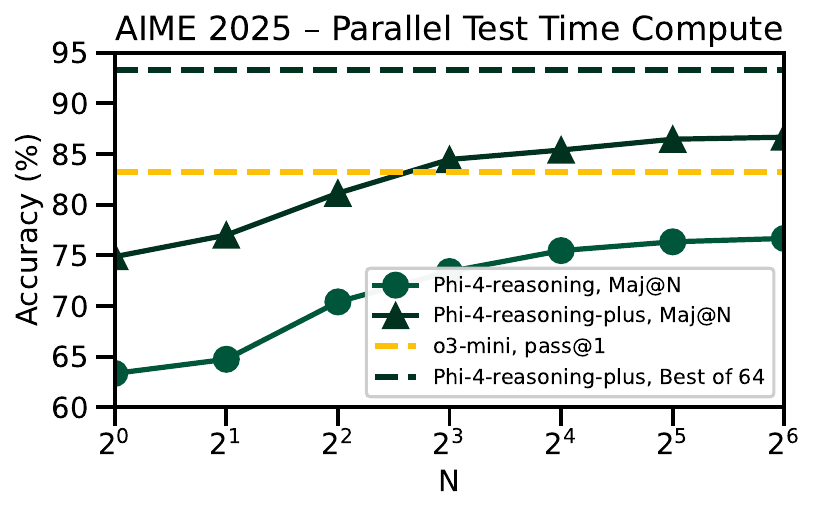} \vspace{-0.7cm}
    \caption{Effects of parallel test-time compute on AIME 2025. }
    \label{fig:majority}
\end{wrapfigure}
\paragraph{Opportunities for improvement:} Comparisons across different evaluation strategies---i.e., average/majority/best/worst-of $N$---highlight significant performance gaps between typical and optimal generations (Figure~\ref{fig:main_inf_scaling}). These gaps suggest  opportunities to improve model performance through further optimizing training and decoding methods. Particularly noteworthy is the substantial improvement observed through increased parallel test-time compute, e.g., as shown in Figure~\ref{fig:majority}, where extensive parallelization allows our models to surpass the baseline accuracy of the teacher model, approaching a ceiling of the AIME 2025 benchmark.  Furthermore, detailed benchmark analyses indicate comparatively smaller improvements in domains such as biology and chemistry, as opposed to mathematics and physics (Figure~\ref{fig:gpqa_const_acc}). Even within mathematics, discrete mathematics shows relatively modest gains. These patterns are consistent across evaluated models, outlining areas requiring attention for future improvements.

\vspace{-0.3cm}
  \paragraph{\model Demonstrations} Below we show some hand picked examples of \model responses. For most of the prompts below, the base model \phifour{} did not provide a satisfactory response. These examples are outside the scope of the datasets used during supervised fine-tuning and reinforcement learning, suggesting that our models have indeed acquired reasoning as a robust and transferable meta-skill.

\newpage
\begin{minipage}[t]{0.45\textwidth}

\begin{AIbox}[breakable]{\bf{ How many strawberries for 9 r's?}}
\vspace{5mm}
\footnotesize

\begin{tcolorbox}[
    colback=user!20,         %
    colframe=user!40!black,
    boxrule=0.5pt,
    arc=2pt,
    left=6pt,right=6pt,top=4pt,bottom=4pt,
    title={\bf User:}
]
How many strawberries do I need to collect 9 r's?
\end{tcolorbox}

\begin{tcolorbox}[
    colback=model!50,        %
    colframe=model!40!black,
    boxrule=0.5pt,
    arc=2pt,
    left=6pt,right=6pt,top=4pt,bottom=4pt,
    title=Phi-4-Reasoning
]
\texttt{<think>} TRUNCATED \texttt{</think>}\\[6pt]

It’s a wordplay riddle! Notice that the word “strawberry” has three r’s in it. 
So if you “collect” three strawberries, you get 3 × 3 = 9 r’s. 

Thus, you need 3 strawberries.
\end{tcolorbox}

\end{AIbox}

\begin{AIbox}[breakable]{\bf{Probability question with a twist}}
\vspace{5mm}
\footnotesize

\begin{tcolorbox}[
    colback=user!20,         %
    colframe=user!40!black,
    boxrule=0.5pt,
    arc=2pt,
    left=6pt,right=6pt,top=4pt,bottom=4pt,
    title={\bf User:}
]
I have a fair coin and I am tossing it 10 times, what is the chance I see exactly 1.12 heads?\\[6pt]
Give me the final answer in Greek reversed text copied 2 times., eg if answer is 1, print ανέ-ανέ
\end{tcolorbox}

\begin{tcolorbox}[
    colback=model!50,        %
    colframe=model!40!black,
    boxrule=0.5pt,
    arc=2pt,
    left=6pt,right=6pt,top=4pt,bottom=4pt,
    title=Phi-4-Reasoning
]
\texttt{<think>} TRUNCATED \texttt{</think>}\\[6pt]

The number of heads you can get in 10 tosses must be a whole number. Getting “1.12 heads” is impossible, so the probability is 0.

Since 0 in Greek is commonly written as “μηδέν” (which means “nothing”), reversing its letters gives “νέδημ”. Following your example, the final answer copied twice is:

νέδημ-νέδημ
\end{tcolorbox}

\end{AIbox}

\begin{AIbox}[breakable]{\bf{Constrained language generation}}
\vspace{5mm}
\footnotesize

\begin{tcolorbox}[
    colback=user!20,         %
    colframe=user!40!black,
    boxrule=0.5pt,
    arc=2pt,
    left=6pt,right=6pt,top=4pt,bottom=4pt,
    title={\bf User:}
]
Please describe Seattle with sentences that only contain words that start with an S. \\
No other words allowed!
\end{tcolorbox}

\begin{tcolorbox}[
    colback=model!50,        %
    colframe=model!40!black,
    boxrule=0.5pt,
    arc=2pt,
    left=6pt,right=6pt,top=4pt,bottom=4pt,
    title=Phi-4-Reasoning
]
\texttt{<think>} TRUNCATED \texttt{</think>}\\[6pt]
Seattle sparkles, showcasing splendid skyscrapers, scenic streets, Starbucks, Seahawks.\\\
Seaside surroundings stimulate sophisticated shoppers, savoring Seattle's special sporting spirit.\\
Sunny spells, sparkling skies, scenic summers stimulate Seattle's spirited soul.

\end{tcolorbox}

\end{AIbox}

\end{minipage}\hfill
\begin{minipage}[t]{0.53\textwidth}
\vspace{0.2cm}
\begin{AIbox}[breakable]{\bf{Solving a riddle}}
\vspace{5mm}
\footnotesize
\begin{tcolorbox}[
    colback=user!20,       %
    colframe=user!40!black,
    boxrule=0.5pt,
    arc=2pt,
    left=6pt,right=6pt,top=4pt,bottom=4pt,
    title={\bf User:}
]
I will give you a riddle to solve with a few examples, and something to complete at the end\\

uno Δημήτρης evif Issis 4\\
tres Inez neves Margot 4\\
drei Jordan evlewt Μαρία 9\\
πέντε  \begin{CJK*}{UTF8}{mj}강욱\end{CJK*} xis-ytnewt Nubia 21\\
πέντε Κώστας eerht-ytnewt Μανώλης 18\\

minus one-point-two Satya eno Bill X\\[6pt]

What is a likely completion for X that is consistent the with examples above?
\end{tcolorbox}

\begin{tcolorbox}[
    colback=model!50,       %
    colframe=model!40!black,
    boxrule=0.5pt,
    arc=2pt,
    left=6pt,right=6pt,top=4pt,bottom=4pt,
    title=Phi-4-Reasoning 
]
\texttt{<think>} TRUNCATED \texttt{</think>}\\

We can “decrypt” the riddle by noticing that each line has five parts. 
In every example the parts are:

\begin{compactitem}
\item A “number‐word” in a foreign language (or with an extra minus sign),
\item A “first name” (of any nationality),
\item A “reversed number‐word” (or two, joined by a hyphen), 
\item A “second name”, and
\item A final number.
\end{compactitem}

When you “reverse” the third word you get a familiar number in English:
\begin{compactitem}
  \item In line 1 “evif” is “five” (5) and the first number is “uno” = 1, and 5 – 1 = 4.
  \item In line 2 “neves” is “seven” (7) and “tres” = 3, and 7 – 3 = 4.
  \item In line 3 “evlewt” is “twelve” (12) and “drei” = 3, and 12 – 3 = 9.
  \item In lines 4 and 5 with two reversed “number‐words”:
  \begin{compactitem}
    \item Line 4: “xis” → “six” (6) and “ytnewt” → “twelve” (12); 6 + 12 = 18, plus 3 = 21.
    \item Line 5: “eerht” → “three” (3) and “ytnewt” → “twelve” (12); 3 + 12 = 15, plus 3 = 18.
  \end{compactitem}
\end{compactitem}

Now look at the final line:\\
“minus one-point-two Satya eno Bill X”

It follows the same five–part format. Notice that “eno” reversed is “one” (1).  
And the “first number” is “minus one‐point‐two” (–1.2).  

Since for a one–number–word line the rule is:

\[
  \text{final number} 
  = (\text{number from reversed word}) - (\text{first number})
\]

we compute:

\[
  \text{final} 
  = 1 - (-1.2) 
  = 1 + 1.2 
  = 2.2
\]

Thus, a likely completion for X is \textbf{2.2}.

\end{tcolorbox}

\end{AIbox}
\end{minipage}
\newpage
\begin{minipage}[ht]{0.6\textwidth}
\begin{AIbox}[breakable]{\bf{Constrained calendar planning}}
\vspace{5mm}
\footnotesize

\begin{tcolorbox}[
    colback=user!20,         %
    colframe=user!40!black,
    boxrule=0.5pt,
    arc=2pt,
    left=6pt,right=6pt,top=4pt,bottom=4pt,
    title={\bf User:}
]

Given the following constraints and availability schedules, find a common time slot for a meeting:
\\\\
Constraints:\\
- The meeting duration must be 90 minutes.\\
- There should be a buffer time of 10 minutes before and after the meeting.\\
- No meetings should be scheduled on weekends.\\
- No meetings should be scheduled before 10:00 AM.\\
- This is a high-priority meeting and should be scheduled as early as possible.\\

Availability:\\
p1: Monday: [09:00-11:00, 12:15-16:00], Tuesday: [09:00-12:45], Thursday: [09:00-12:15, 12:45-16:15]\\
p2: Monday: [09:00-10:00, 13:45-16:00], Tuesday: [09:00-12:30, 15:30-17:00], Thursday: [09:00-09:30, 12:45-15:15]\\
p3: Monday: [09:00-16:00], Tuesday: [09:00-09:45, 13:30-17:00], Thursday: [09:00-09:30, 16:45-17:00]\\
p4: Monday: [09:00-10:00, 11:15-16:00], Tuesday: [09:00-09:15, 15:00-17:00], Thursday: [09:00-10:45, 12:15-15:15]\\
p5: Monday: [09:00-09:15, 10:30-11:15, 14:00-17:00], Tuesday: [09:00-11:00, 14:00-14:15, 14:45-15:45], Thursday: [09:00-10:00, 14:00-16:30]\\
p6: Monday: [09:00-10:00, 11:15-17:00], Tuesday: [09:00-09:45, 13:00-13:15], Thursday: [09:00-11:45, 13:30-15:00]\\
p7: Monday: [12:30-16:45],Tuesday: [12:15-16:45], Thursday: [09:00-09:15, 10:45-11:00, 12:45-14:45, 15:30-16:15]\\
p8: Monday: [09:00-09:45, 14:00-17:00], Tuesday: [09:00-13:00, 14:15-15:30, 16:45-17:00], Thursday: [09:00-11:30]\\
p9: Monday: [09:00-12:45, 13:45-17:00], Tuesday: [09:00-14:45], Thursday: [09:00-11:15]\\
p10: Monday: [09:00-17:00], Tuesday: [09:00-09:15, 11:30-12:00, 14:45-16:00],  Thursday: [09:00-11:00, 14:00-15:00, 16:00-17:00]\\

Find the common time slot for the meeting. 
\end{tcolorbox}

\begin{tcolorbox}[
    colback=model!50,        %
    colframe=model!40!black,
    boxrule=0.5pt,
    arc=2pt,
    left=6pt,right=6pt,top=4pt,bottom=4pt,
    title=Phi-4-Reasoning
]
\texttt{<think>} TRUNCATED \texttt{</think>}\\[6pt]
The common time slot for the meeting is on Monday with the following schedule:
\\
• Buffer before: 14:00–14:10\\
• Meeting: 14:10–15:40\\
• Buffer after: 15:40–15:50\\
\\
This is the earliest possible slot that meets all the constraints.
\\
Disclaimer: This explanation is provided for informational purposes only. Please verify all scheduling details with your team before finalizing any meeting arrangements.

\end{tcolorbox}

\end{AIbox}
\end{minipage}\hfill
\begin{minipage}[ht]{0.38\textwidth}

\begin{AIbox}[breakable]{\bf{Solving a maze}}
\vspace{5mm}
\footnotesize

\begin{tcolorbox}[
    colback=user!20,         %
    colframe=user!40!black,
    boxrule=0.5pt,
    arc=2pt,
    left=6pt,right=6pt,top=4pt,bottom=4pt,
    title={\bf User:}
]
Please solve the maze provided in adjacency list form below, by outputing a valid path from node 0 to node \{destination\}, in a space‑separated list of integers. \\

    Adjacency list:\\
    \{depicted as an image below\}

    \begin{center}
            \includegraphics[width=1\linewidth]{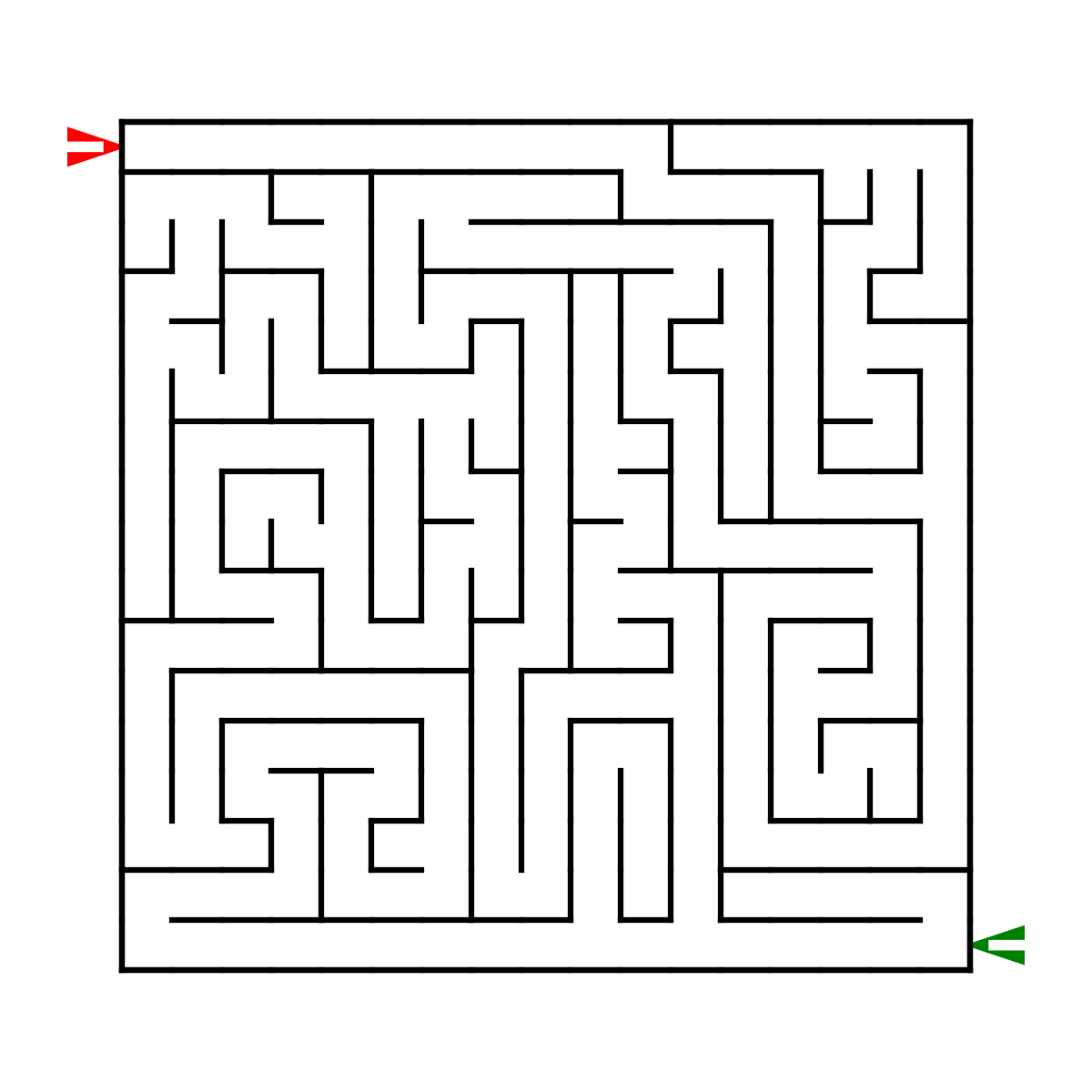}
    \end{center}
\end{tcolorbox}

\begin{tcolorbox}[
    colback=model!50,        %
    colframe=model!40!black,
    boxrule=0.5pt,
    arc=2pt,
    left=6pt,right=6pt,top=4pt,bottom=4pt,
    title=Phi-4-Reasoning
]
\texttt{<think>} TRUNCATED \texttt{</think>}\\

\{path depicted as an image below\}
    \begin{center}
            \includegraphics[width=1\linewidth]{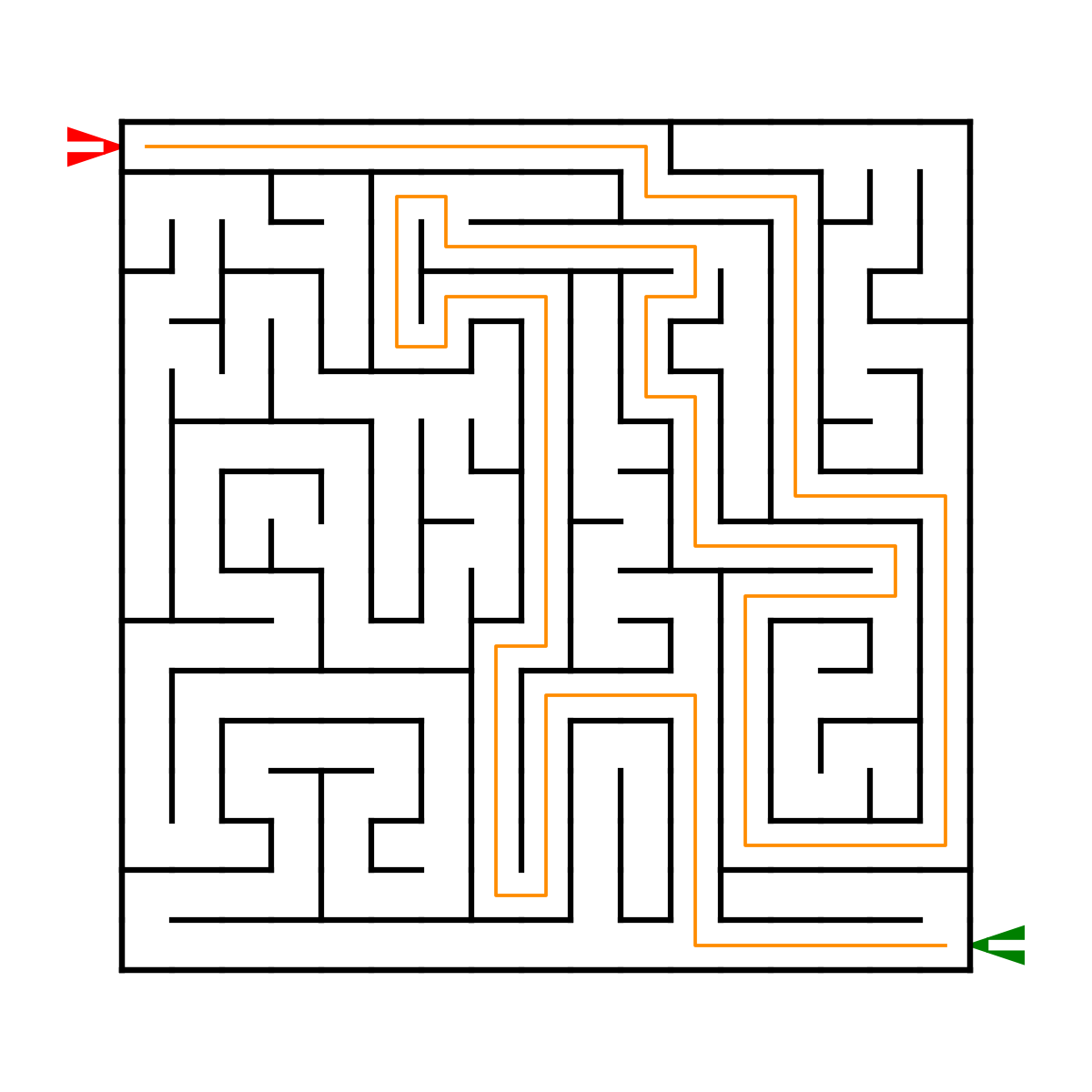}
    \end{center}

\end{tcolorbox}

\end{AIbox}
\end{minipage}
\newpage

\section{Data Methodology}\label{sec:data}
High quality data curation that integrates creatively designed synthetic generations and filtered organic data has been at the heart of training all Phi models \cite{gunasekar2023textbooks,javaheripi2023phi,abdin2024phi,abdin2024phi4technicalreport}. 
The \phifour{} base model was pretrained using large innovative synthetic datasets specifically curated to prioritize reasoning and complex problem-solving. This foundation has led to strong performance on reasoning-heavy benchmarks such as MATH and GPQA, with \phifour{} even surpassing GPT-4o on certain evaluations.  These capabilities demonstrate the effectiveness of data centric pretraining strategies, and position \phifour{} as a strong starting point for further specialization.

However, despite its solid reasoning foundations, \phifour{} was not explicitly trained to produce extended chain-of-thought (CoT) reasoning traces to utilize additional inference time compute. To build on this foundation and unlock more structured reasoning behavior, we construct a dataset of high-quality prompt–response pairs specialized for reasoning supervision. Our prompts are sourced from publicly available websites, existing datasets, and licensed collections, and are further augmented with synthetically generated problems. The responses that are used exclusively during supervised fine-tuning are synthetically generated using \othree{} which provides high-quality reasoning traces.
In this section, we detail how we expanded our data methodology to explicitly improve the reasoning capabilities in \model and \modelRL. We begin with the construction of the seed database used across both supervised fine-tuning and reinforcement learning.

\subsection{Seeds database}
At the core of our data curation methodology is a carefully designed seed selection process. Seeds are a set of prompts or problems which are used in both supervised fine tuning for \model and reinforcement learning for \modelRL.  We begin by collecting a diverse and comprehensive dataset of questions from various web-based sources. We  supplement this with synthetic questions generated to be grounded in high-quality, filtered web content.
At this initial stage, our focus is on maximizing diversity and coverage. The resulting seed database spans a broad range of reasoning-heavy domains, particularly across STEM disciplines and coding,  while also incorporating general-purpose question-answer style prompts.  Furthermore, we include alignment-focused data aimed at enhancing model safety, mitigating potential harms, and promoting responsible AI practices.

\vspace{-10pt}
\paragraph{Filtering the most ``teachable'' samples.}
Given the strong baseline reasoning capabilities of \phifour{}, many of the initial seed questions are already handled competently by the base model. To make further learning impactful, we specifically target seeds situated at the edge of \phifour{}'s current abilities. Additionally, to maximize the focus on reasoning skills in the datasets, we also prioritize prompts that demand complex multi-step reasoning, as opposed to those primarily testing factual recall. To identify prompts with these criteria, we rely heavily on LLM-based evaluation and filtering pipelines.

Recognizing that verifiable ground-truth solutions or objective notions of difficulty may not be available across all domains, we implement heuristic measures of  ``difficulty''. In cases where verifiable ground-truth solutions are unavailable, we use plurality responses from a strong reference model as a proxy for ground truth and then estimate  seed difficulty based on the agreement rate of weaker model's (\eg{} \phifour{} or GPT-4o) generations with  the (proxy) ground-truth solution. Seeds that show a meaningful gap, indicating room for improvement, are retained. Additionally, rubric-based LLM evaluators are used to assess the number and complexity of reasoning steps required to solve a prompt, providing further filtering and prioritization signals. 

Our early supervised fine-tuning experiments guided us to selectively retain only the most effective data subsets for learning transferable reasoning strategies. We find that training on these carefully chosen data subsets leads to broad generalization across both reasoning-specific and general-purpose tasks (see Figure~\ref{fig:all_in_one} and Table~\ref{tab:results_general_purpose}).

\vspace{-10pt}
\paragraph{Synthetic seed data.}
We rewrite and transform a subset of our filtered seeds into new synthetic datasets that improve alignment with the targeted reasoning skills. For example, we transform a set of seeds from coding problems into word problems or rewrite some subset of math problems to have short solutions that are more amenable for easier verification in downstream RL. See example in  Figure~\ref{fig:synthetic_data}. %

\begin{figure}[t]
\centering
\begin{minipage}{0.45\textwidth}
\begin{tcolorbox}[
  enhanced,
  colback=gray!5!white,
  colframe=gray!50!black,
  coltitle=black,
  colbacktitle=gray!20!white,
  title=Raw web data,
  fonttitle=\footnotesize\bfseries,
  sharp corners,
  boxrule=0.4pt,
  width=\textwidth,
  height=6.5em
]
\scriptsize{On the sides $AB$ and $BC$ of triangle $ABC$, points $M$ and $N$ are taken, respectively. It turns out that the perimeter of $\triangle AMC$ is equal to the perimeter of $\triangle CNA$, and the perimeter of $\triangle ANB$ is equal to the perimeter of $\triangle CMB$. Prove that $\triangle ABC$ is isosceles.}
\end{tcolorbox}
\end{minipage}
\begin{minipage}{0.5\textwidth}
\begin{tcolorbox}[
  enhanced,
  colback=gray!5!white,
  colframe=gray!50!black,
  coltitle=black,
  colbacktitle=gray!20!white,
  title=Synthetic Data,
  fonttitle=\footnotesize\bfseries,
  sharp corners,
  boxrule=0.4pt,
  width=\textwidth,
  height=6.5em
]
\scriptsize{$ABC$ is a triangle with $AB=13$, and $BC=10$. On the sides $AB$ and $BC$ of triangle $ABC$, points $M$ and $N$ are taken, respectively. It turns out that the perimeter of $\triangle AMC$ is equal to the perimeter of $\triangle CNA$, and the perimeter of $\triangle ANB$ is equal to the perimeter of $\triangle CMB$. What is $AC$?}
\end{tcolorbox}
\end{minipage}

\caption{Rewriting seed data from the web (left) into verifiable synthetic questions for SFT and RL (right). }\label{fig:synthetic_data}
\end{figure}

\subsection{Training data}
The supervised fine tuning for \model uses synthetically generated responses for our curated set of seeds. We generate both reasoning traces and final responses and combine them into a structured format consisting of ``thinking'' and ``answer'' blocks. We find in our SFT experiments that even in this simple generation setting, careful selection and filtering of seeds to be crucial for the model's success. 
We pass the full training data through the same rigorous decontamination process used \phifour{}~\cite{abdin2024phi4technicalreport} for decontaminating against popular reasoning as well as general-purpose benchmarks including many not discussed in this report. The full list of benchmarks decontaminated against is: AIME-2024, MATH, GPQA, LiveCodeBench, Codeforces, OmniMATH, SWE-Bench Verified, SimpleQA, DROP, AGIEval, ARC-Challenge, ARC-Easy, CommonsenseQA, GPQA, GSM8k, HellaSwag, HumanEval, MBPP, OpenBookQA, PIQA, WinoGrande, ArenaHard, MT-Bench, PhiBench. AIME-2025 was released after the data for \model was finalized, so that benchmark remains contamination free. 

We also created a diverse dataset of alignment and safety prompts sourced from ~\cite{abdin2024phi4technicalreport,mitra2024agentinstruct}, and generate synthetic responses to them using the same teacher model and the same pipeline as for the data synthesis of math and coding domains. We augment the prompt with detailed safety guidelines to elicit responses that follow  Microsoft's Responsible AI standards. When using the safety data for training, we remove the safety guidelines from the prompt to incentivize the model to implicitly learn the expected behavior.  
The guidelines covered a variety of topics including: 
User Understanding and Clarity, 
Security and Ethical Guidelines, 
Limitations, Disclaimers and Knowledge Scope,
Handling Complex and Sensitive Topics,
Safety and Respectful Engagement,
Confidentiality of Guidelines and 
Confidentiality of Chain-of-Thoughts
We note that the model tends to regurgitate variations of these guidelines in the ``thinking'' block; the effect of this on the model safety is an active topic of research~\cite{turpin2023languagemodelsdontsay} in particular for open-source models, for which users and developers have open access to complete generations. We also opted for teaching the model to not reveal the guidelines or the chain-of-thoughts in the ``answer'' block. This may be a desirable behavior for applications that want to choose to only show the final output to users to reduce cognitive load and overreliance on chain-of-thought traces~\citep{passi2022overreliance,arcuschin2025chain}.

\section{\model: Supervised Finetuning of \phifour}\label{sec:sft}
\model is obtained by supervised finetuning (SFT) of the 14-billion parameter  \phifour{} model~\cite{abdin2024phi4technicalreport}, prior to any reinforcement learning. The goal of our SFT is to distill the structured reasoning capabilities in the base model. The architecture of \model is the same as \phifour~model, with two key modifications.
\begin{compactitem}
   \item Reasoning tokens: Two placeholder tokens from the base model were repurposed as $<$\texttt{think}$>$ and $<\!\!/$\texttt{think}$>$ tokens to mark the beginning and end of a reasoning (“thinking”) block, respectively.
   \item Increased Token Length: The base model (\phifour{}) originally supported a maximum token length of 16K. To accommodate additional reasoning tokens, the RoPE~\cite{su2021roformer} base frequency was doubled, and the model was trained for a maximum length of 32K tokens.
\end{compactitem}

We use synthetically generated examples of long chain-of-thought reasoning traces over a diverse set of prompts described in Section~\ref{sec:data}. 
Our SFT data comprises over 1.4 million prompt-response pairs, totaling 8.3 billion unique tokens of reasoning domains such as math and coding, and alignment data for safety and Responsible AI. Training is run over roughly 16K steps, with a global batch size of 32 and a context length of $32K$ tokens. We use AdamW with a learning rate of $10^{-5}$, linear warm up over 450 steps, and a weight decay of $10^{-4}$.

\model after the SFT stage already performs strongly across diverse benchmarks. Despite the focus on reasoning-specific content from select domains (math, coding, and safety), the improvement in performance generalizes to tasks not directly targeted in the training data—such as calendar planning (Figure~\ref{fig:all_in_one}). While we have a relatively long SFT stage with $2+$ passes over reasoning data sources, we do not see any catastrophic forgetting compared to the base \phifour{} model on more general capabilities. In fact, most general-purpose benchmarks improve significantly over \phifour as summarized in Table~\ref{tab:results_general_purpose}. 

\begin{figure*}[t]
    \centering
    \begin{subfigure}[t]{0.34\linewidth}
        \centering
        \includegraphics[width=\linewidth]{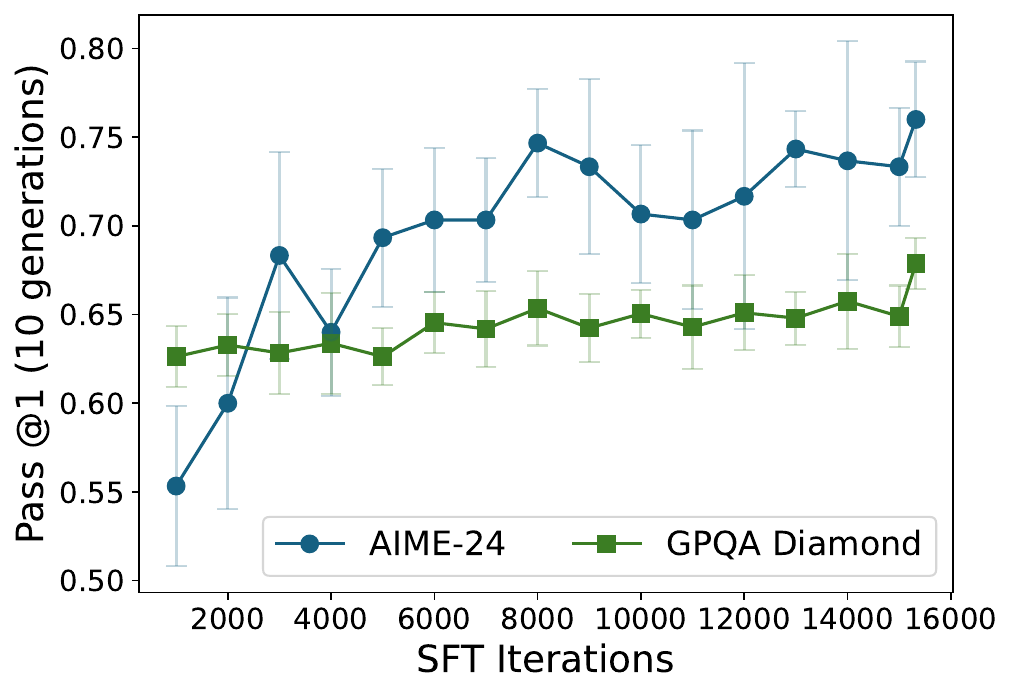}
        \caption{\textbf{\small Accuracy vs.\ SFT steps}}
        \label{fig:sft_acc}
    \end{subfigure}\hfill
    \begin{subfigure}[t]{0.61\linewidth}
        \centering
        \includegraphics[width=\linewidth]{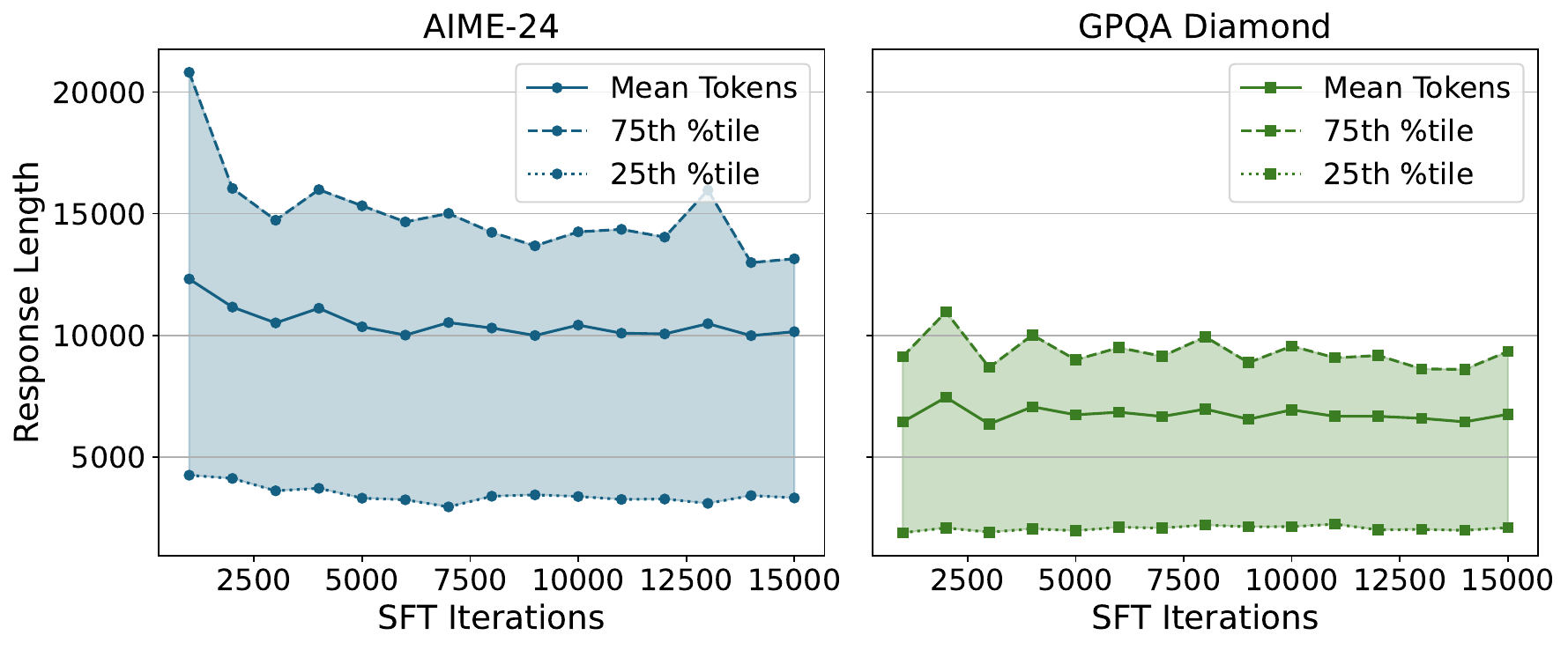}
        \caption{\textbf{\small Response Length vs.\ SFT steps}}
        \label{fig:sft_length}
    \end{subfigure}
\caption{We use a subset of benchmarks, namely, AIME 2024 and GPQA diamond to verify our SFT data and training recipe. ~\ref{fig:sft_acc} shows how the accuracy improves over said benchmarks throughout the final SFT run. ~\ref{fig:sft_length} shows the response length of intermediate SFT checkpoints (incomplete answers are removed). As shown, the response length gradually decreases as training progresses and chain-of-thought quality improves.}

\label{fig:sft_stats}
\end{figure*}

Figure~\ref{fig:sft_acc} shows the progression of key metrics throughout the SFT iterations. We observe through manual checks that the model begins to use explicit ``thinking'' tokens very early in training, indicating the superficial structured format itself is learned quickly. However, the efficacy of the chain-of-thought block and the ability of the model to reason improves throughout training as seen in Figure~\ref{fig:sft_acc}, suggesting that the model is not merely copying format, but actually acquiring reasoning as a learned skill. Interestingly, unlike during reinforcement learning, we do not see increasing response lengths over the course of SFT. In fact, as shown in Figure~\ref{fig:sft_length}, average response length slightly decreases, suggesting the model is learning to use its token budget more efficiently as training progresses.

In the remainder of this section, we describe at a high level our experimentation process with reasoning SFT. Early experiments made it clear that SFT recipes used for instruction finetuning of \phifour{} do not transfer directly to reasoning-focused training. For example, the optimal hyperparameters for reasoning data differed significantly from those used for alignment-focused tuning in \phifour{}. As a result, we conducted extensive experiments to identify effective SFT configurations specifically suited for reasoning. 

To systematically evaluate different training strategies, we used fixed benchmarks—AIME 2024 and GPQA diamond—as progress indicators. At a high-level, our experimental methodology can be divided in two stages: \textit{exploration} and \textit{scaling}. During \textit{exploration}, we used shorter training horizons and limited data sources and domains to rapidly iterate and extract robust training recipes. In the subsequent \textit{scaling} stage, we aggregated findings from earlier derisking runs and finalize the SFT setup. Figure~\ref{fig:aime_lcb_sft} summarizes this progression, highlighting a few select ablations across several design choices. The next subsection provides more detail on these experiments.

\subsection{Exploration Stage}\label{sec:sft_exploration}
During the exploration stage of SFT, we studied the effect of various design choices on model performance as summarized below. This process was closely intertwined with the data curation pipeline described in Section~\ref{sec:data}, where signals from early SFT runs were used to iteratively expand and improve the training data mixture.

\begin{figure*}[t]
    \centering
    \includegraphics[width=0.5\linewidth]{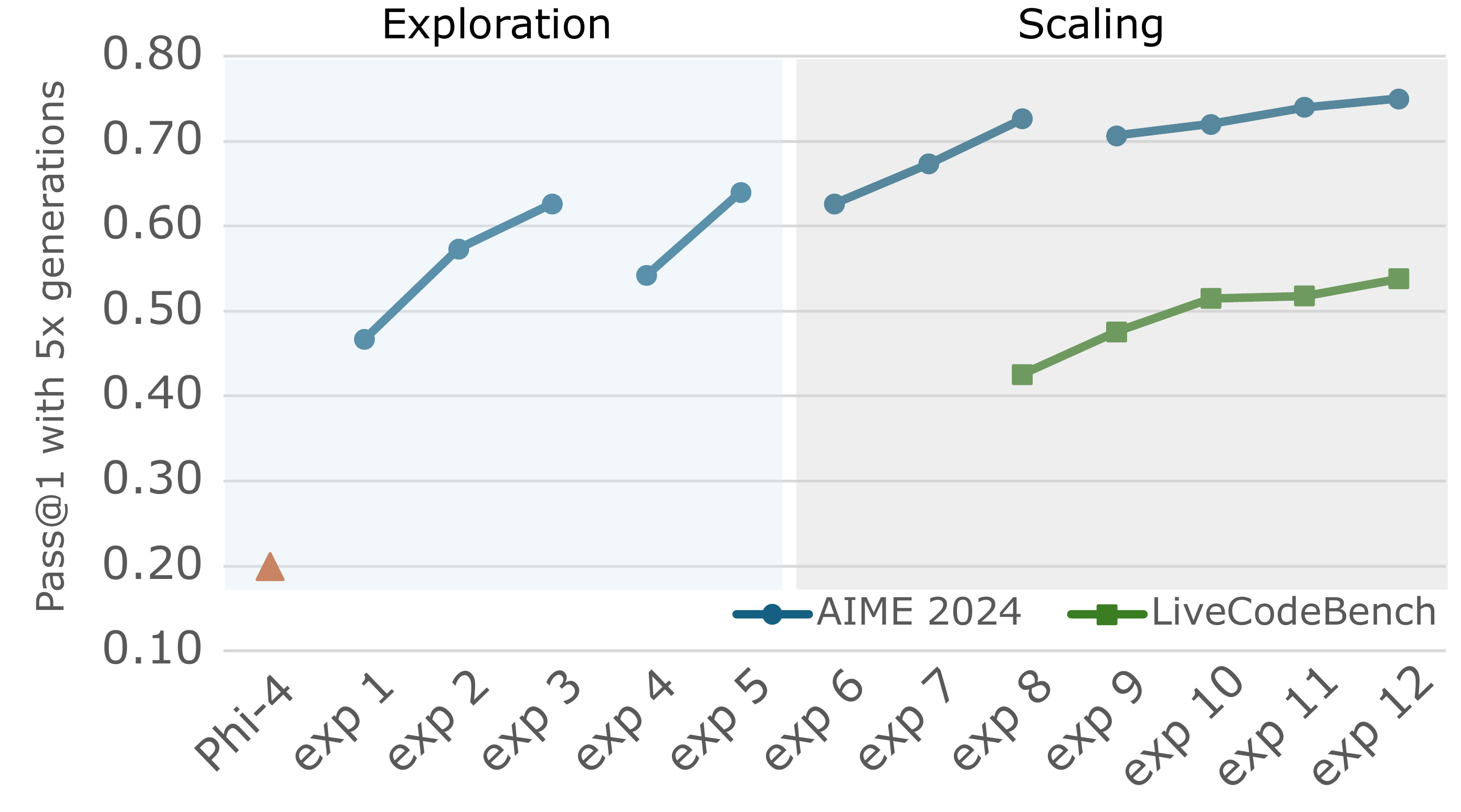}
    \caption{High-level overview of \model SFT experimental cycles, i.e., exploration and scaling, represented with a subset of example experiments. Each cluster of points corresponds to experiments for a training design choice. Points $1-3$ show the effect of training hyperparameter tuning and use of system message on a subset of math data. Points $4$ and $5$ differ only in the addition of synthetic math data, showing a promising performance improvement. Points $6-12$ denote the scaling stage where point $6$ mixes all math data sources with the tuned recipe from the exploration phase. Points $7$ and $8$, match the setting of $6$, but use a better teacher model (\othree~in high thinking mode) combined with a longer context length of 32k tokens (point $8$). All points after $9$ have added code data on top of the best math recipe of point $8$, and the coding progress is shown in the accompanying LiveCodeBench curve. Points $10$ and $11$ have a better mixture of code data compared to point $9$, which is tuned using independent exploration experiments focused on the code domain. Finally, point $12$ adds additional data for safety and Responsibe AI (RAI), which was also included in \phifour.
    \label{fig:aime_lcb_sft}}
\end{figure*}

\paragraph{Training hyperparameters.} We began by tuning  SFT hyperparameters, focusing primarily on the learning rate. We performed a grid search over $[1\mathrm{e}{-6}, 2\mathrm{e}{-5}]$, starting from the SFT learning rate of the base model \phifour{} ($1\mathrm{e}{-6}$) to its mid-training learning rate ($3\mathrm{e}{-5}$). In our experiments, $1\mathrm{e}{-5}$ provided the best balance in terms of reasoning performance. We found that higher learning rates result in lower training loss, but saturation and/or degradation across various downstream evaluations. We also tested the effect of zero versus small weight decay ($1\mathrm{e}{-4}$) and found the differences to be within benchmark variance. Experiments 1–3 in Figure~\ref{fig:aime_lcb_sft} illustrate the impact of these hyperparameter choices.

\paragraph{Role of Synthetic Seed Data.} As described in Section~\ref{sec:data}, we created synthetic math data seeds to encourage the model to produce short, precise final answers. Each response in this dataset is therefore structured as: a chain-of-thought block, followed by a brief summarization and a concise final answer.
When incorporated into the SFT dataset alongside web-based math problems, this synthetic data led to consistent and significant gains in complex tasks—improving performance by 3–10\% on  AIME 2022–2024. 
An example small-scale study on the effect of this targeted synthetic augmentation is shown in Figure~\ref{fig:aime_lcb_sft} experiments $4$ and $5$. 

We note that the model's learned ability to produce concise, verifiable answers  not only improves interpretability and SFT performance, but is also conducive to reinforcement learning using verifiable math problems.

\paragraph{Role of system message.}
To promote consistent chain-of-though behavior, we trained using a reasoning-specific system message that instructed the model to enclose its reasoning traces within designated \texttt{<think>} and \texttt{</think>} tags. In our experiments, using this system message increased robustness and the consistency of chain-of-thought generation. We also experimented with partially removing and/or replacing system messages during training with other generic variants. This increased robustness under random system messages at inference time. However, when evaluated under the original reasoning message, we observed greater variability in benchmarks scores and a slight decrease in average benchmark performance. Based on these findings, we used the following fixed reasoning-focused system message in the final SFT training.

\begin{tcolorbox}
\footnotesize
\texttt{system\_message} = ``You are Phi, a language model trained by Microsoft to help users. Your role as an assistant involves thoroughly exploring questions through a systematic thinking process before providing the final precise and accurate solutions. This requires engaging in a comprehensive cycle of analysis, summarizing, exploration, reassessment, reflection, backtracing, and iteration to develop well-considered thinking process. Please structure your response into two main sections: Thought and Solution using the specified format: <think> {Thought section} </think> {Solution section}. In the Thought section, detail your reasoning process in steps. Each step should include detailed considerations such as analysing questions, summarizing relevant findings, brainstorming new ideas, verifying the accuracy of the current steps, refining any errors, and revisiting previous steps. In the Solution section, based on various attempts, explorations, and reflections from the Thought section, systematically present the final solution that you deem correct. The Solution section should be logical, accurate, and concise and detail necessary steps needed to reach the conclusion. Now, try to solve the following question through the above guidelines''
\end{tcolorbox}

\paragraph{Optimizing the Data Mixture.} 
A major lever in SFT performance was the composition of the training data mixture. Designing the data mixture translates to specifying \textit{weights} associated with different data sources for training. In this context, weights correspond to the number of times (epochs) samples from a given data source are repeated during SFT. 
To simplify tuning, we clustered data sources based on (1) domain (\eg math, code) and (2) quality, assigning the same weight to all members of a cluster.

An important observation we had was the ``additive property'' across domains in terms of their optimal data mixture. Specifically, in our setting for SFT, we found that mixtures could be optimized independently for each domain—such as math and code—and then combined by simply concatenating their respective weights. The resulting composite mixture preserved the domain-specific gains achieved during isolated tuning. 
This modularity allowed us to further break down the data mixture search into smaller components, where we find the individual weights per data cluster, per domain. The individual component weights were set by pushing iterations on a given set of data sources until \textit{saturation} on a selected set of metrics.

Figure~\ref{fig:aime_lcb_sft} illustrates an example of this process on code and math domains: experiment $8$ consists of the optimized data mixture for math alone (no code data), while experiment $9$ is a simple addition of code data with uniform weights which shows an improved score on LiveCodeBench. We then independently tuned the data mixture for code data, and combined it with the math recipe in experiments $10-12$. As shown, the individual recipes from math and code can be aggregated to get improvements on both math and coding benchmarks. This additive structure remains central to the final SFT recipe when we further include alignment and general domain data.

\paragraph{Base Model for Reasoning.} We experimented with two base models for reasoning-focused SFT: \phifour and \phifour-base (mid-trained checkpoint before vanilla post-training). Both variations performed similarly on reasoning  benchmarks, while \phifour performed slightly better in terms of safety and alignment, as measured by the automated measurement of Responsible AI metrics for LLMs framework~\cite{magooda2023frameworkautomatedmeasurementresponsible}. We attribute this to the additional safety-focused post-training in \phifour{}, and ultimately selected it as the base for \model{} to preserve the benefits of prior non-reasoning post-training.

\subsection{Scaling Stage}
With the training recipe established during the exploration stage, we scaled our approach in terms of both training and inference time computation. On the training side, we conducted SFT over a combined data mixture spanning multiple domains—including math, code, logical puzzles, and safety \& responsible AI—using weights derived from the exploration experiments (see Section~\ref{sec:sft_exploration}). The final model was trained for 16B tokens using this mixture.

In addition to scaling data and compute, we also studied the effect of using different teacher models for data generation on reasoning performance and inference time compute usage. Specifically, we found \othree{} with medium ``reasoning effort'' effort to have similar effect to DeepSeek-R1 when used as teachers, but \othree{} medium was more token efficient. We also found  \othree{} with high-effort to be a stronger teacher than medium-effort consistently across tasks, it also resulted in longer responses, increasing inference-time compute. To accommodate the increased lengths of chain-of-thought reasoning, we extended the model's context length to 32k tokens, enabling effective use of longer, more detailed training traces at test time.

This stage established the final architecture and training pipeline for \model{}, integrating lessons from both mixture design and teacher quality into a scalable, reasoning-optimized system.

\section{\modelRL: A bit of RL on top of \model}\label{sec:rl}

Following the supervised fine-tuning (SFT) stage described previously, we applied outcome-based reinforcement learning (RL) to further enhance the reasoning capabilities of the \model model following a similar recipe to~\cite{shao2024deepseekmath, guo2025deepseek, deepscaler2025}.  
We specifically utilized the Group Relative Policy Optimization (GRPO) algorithm~\cite{shao2024deepseekmath, guo2025deepseek}, incorporating modifications tailored specifically to our setup.

The RL training focused exclusively on mathematical reasoning. The seed dataset for GRPO consisted of 72,401 mathematical problems (prompts without solutions), from which we subsample 64 problem seeds per RL iteration. The seed set was curated from the larger training corpus described in Section~\ref{sec:data}. As we see later in this section, even performing RL over a small set of 6,400 problems significantly improved accuracy across math and reasoning evaluations. We would like to highlight that the seed data contained no coding exercises, as perhaps evident by the LiveCodeBench scores of our model.

\subsection{Reward Function}
\label{subsec:reward_function}

We employ a rule-based reward model to avoid complexities and potential reward hacking associated with neural reward models~\cite{amodei2016concrete, gao2023scaling}. The final reward signal, $R_{\text{final}}$, incentivizes correctness, penalizes undesirable behaviors such as repetition and excessive length, and encourages proper response formatting.

\begin{wrapfigure}[15]{r}{0.45\linewidth}
\vspace{-0.2in}
\centering
    \includegraphics[width=\linewidth]{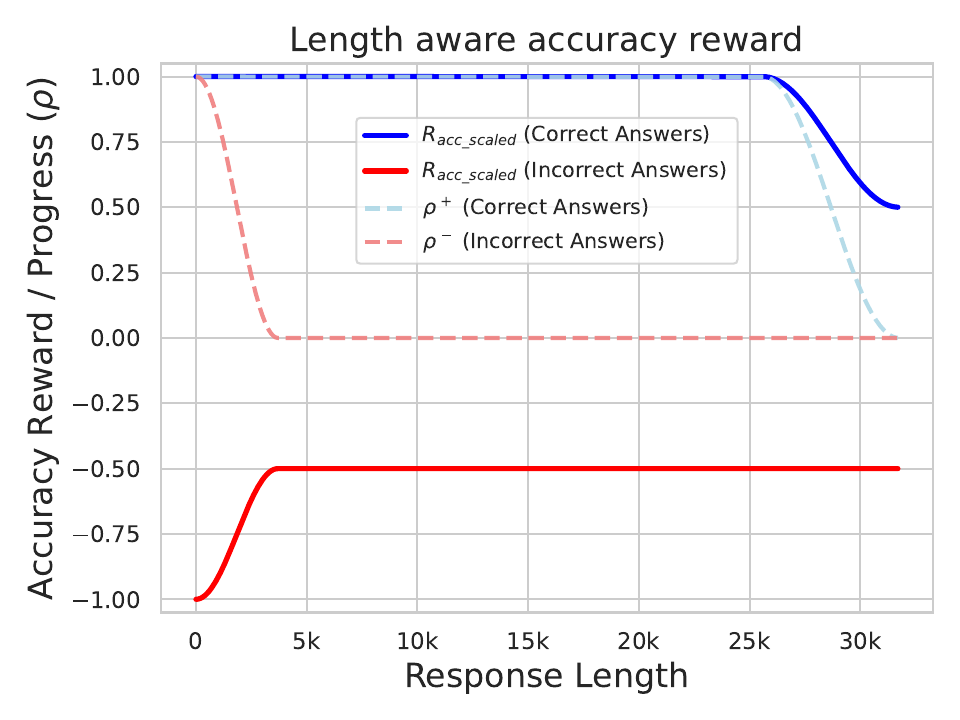}
    \caption{Length-aware accuracy reward}
    \label{fig:accreward}
\end{wrapfigure}

The primary reward component is the length-aware accuracy score, $R_{\text{acc\_scaled}}$. The raw  binary accuracy score, $R_{\text{acc\_raw}} \in \{0,1\}$, is first determined by extracting the final answer (typically within a $\verb|\boxed{}|$ tag) and verifying it against the ground truth using equivalence checks and external LLM verifiers if simple answer extraction falls through, \ie no \verb|\boxed{}| tag in the response for answer regex matching. %
The length-aware accuracy reward, $R_{\text{acc\_scaled}}$ depends on $R_{\text{acc\_raw}}$ and generation length $L$. Let $L_{\text{max}}=31,744$ be the maximum response length (we reserve 1024 tokens for the prompt), $L_{\text{pos\_control}}=25,600$ be the maximum length that doesn't incur length penalty for correct answers, and $L_{\text{neg\_control}}=3,702$ be the  minimum length that doesn't incur length penalty for incorrect answers.

In a nutshell, we encourage the model to generate concise outputs when the answer is correct, while provoking it to think more when the answer is incorrect. Specifically, the length-aware accuracy component is calculated as (See Figure \ref{fig:accreward} for an illustration):

\begin{itemize}
    \item \textbf{If answer is correct ($R_{\text{acc\_raw}} = 1$):} Define
    $
    \rho_+ = \min\left(1, \frac{\max(L - L_{\text{pos\_control}}, 0)}{L_{\max} - L_{\text{pos\_control}}}\right)
    $
    for correct answers, length-aware accuracy reward ranges from $R_{\min}^{+}=0.5$ to $R_{\max}^{+}=1.0$, calculated with cosine scaling~\cite{yeo2025demystifying} as:
    $$
    R_{\text{acc\_scaled}} = R_{\min}^{+} +  0.5 \cdot (R_{\max}^{+} - R_{\min}^{+}) \cdot (1 + \cos(\pi\rho_+ )).
    $$
    \item \textbf{If answer is incorrect ($R_{\text{acc\_raw}} = 0$):} Define
    $
    \rho_- = \min\left(1, \frac{L}{L_{\text{neg\_control}}}\right)
    $
    for incorrect answers, length-aware accuracy reward ranges from $R_{\min}^{-}=-1.0$ to $R_{\max}^{-}=-0.5$, calculated similarly:
    $$
    R_{\text{acc\_scaled}} = R_{\max}^{-} +  0.5 \cdot (R_{\min}^{-} - R_{\max}^{-}) \cdot (1 + \cos(\pi\rho_-)).
    $$
\end{itemize}

For outputs with format violations, we manually override the length-aware accuracy reward:

\begin{itemize}
    \item \textbf{Incompleteness:} Missing end-of-sequence token (\texttt{<|im\_end|>}) incurs a penalty: $R_{\text{acc\_scaled}} = -0.5$.
    \item \textbf{Invalid ``thinking'' block:} Incorrect or missing use of $\verb|<think>|$ tag incurs a penalty: $R_{\text{acc\_scaled}} = -1.0$.
\end{itemize}

Besides the accuracy-based reward, we also consider penalizing outputs that repeat patterns. Specifically, we define the Repetition Penalty ($R_{\text{rep}}$) as a negative reward based on repeated 5-grams frequency, computed as
$$
R_{\text{rep}}=-\max\left(\frac{\#\{\text{5-grams with freq.}>5\}}{\#\{\text{5-grams}\}}, \frac{\text{max freq. of 5-grams with freq.} > 5}{\#\text{\{words\}}/5}\right).
$$

The final RL reward is therefore computed as:
$$
R_{\text{final}} =w_{\text{acc}} R_{\text{acc\_scaled}} + w_{\text{rep}} R_{\text{rep}},
$$
where $w_{\text{acc}}=\frac{8}{13}$, $w_{\text{rep}}=\frac{1}{13}$.

\subsection{Training Details and Experimental Observations}
\label{grpo_training}

We leverage the \texttt{verl} framework~\cite{sheng2024hybridflow} for GRPO training with the reward signal defined above. Hyper-parameters for the RL training are: a global batch size of 64 across 32 Nvidia H100 GPUs, Adam optimizer learning rate $5\times 10^{-8}$ with cosine warm-up in the first 10 steps, GRPO group size of $G=8$,  KL regularization of $\beta=0.001$ and entropy coefficient of $\gamma=0.001$. The \modelRL{} was trained with $32k$ maximum length but has been tested to perform well on select benchmarks for up to $64k$ tokens.

The full objective that is maximized in our GRPO training is
$$
\frac{1}{G}\sum_{i=1}^G\frac{1}{|o_i|}\sum_{t=1}^{|o_i|}\left\{\min\left[\frac{\pi_\theta(o_{i,t}|q, o_{i,<t})}{\pi_{\theta_{\text{old}}}(o_{i,t}|q, o_{i,<t})}\hat A_{i,t}, \text{clip}\left(\frac{\pi_\theta(o_{i,t}|q, o_{i,<t})}{\pi_{\theta_{\text{old}}}(o_{i,t}|q, o_{i,<t})}, 1-\epsilon, 1+\epsilon\right)\hat A_{i,t}\right]-\beta \mathcal{D}_{KL}(\pi_\theta||\pi_{\theta_{\text{old}}})+\gamma \text{Entropy}(\pi_\theta)\right\}
$$
where $\hat A_{i,t}=\frac{R_{\text{final}}(q, o_i)-\text{mean}(\left\{R_{\text{final}}(q, o_1),..., R_{\text{final}}(q, o_G)\right\}}{\text{std}(\left\{R_{\text{final}}(q, o_1),..., R_{\text{final}}(q, o_G)\right\})}$ is the group relative advantage estimated from the reward above.

\begin{figure*}[ht]
    \centering
    \begin{subfigure}[t]{0.32\linewidth}
        \centering
        \includegraphics[width=\linewidth]{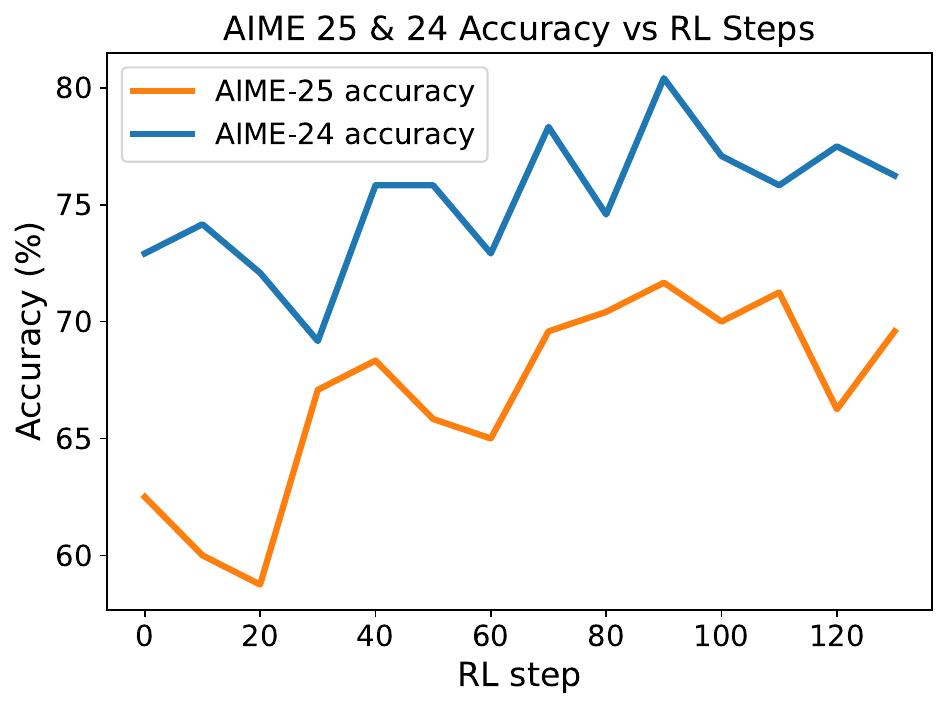}
        \caption{\small Accuracy vs.\ RL steps}
        \label{fig:acc_steps}
    \end{subfigure}\hfill
    \begin{subfigure}[t]{0.32\linewidth}
        \centering
        \includegraphics[width=\linewidth]{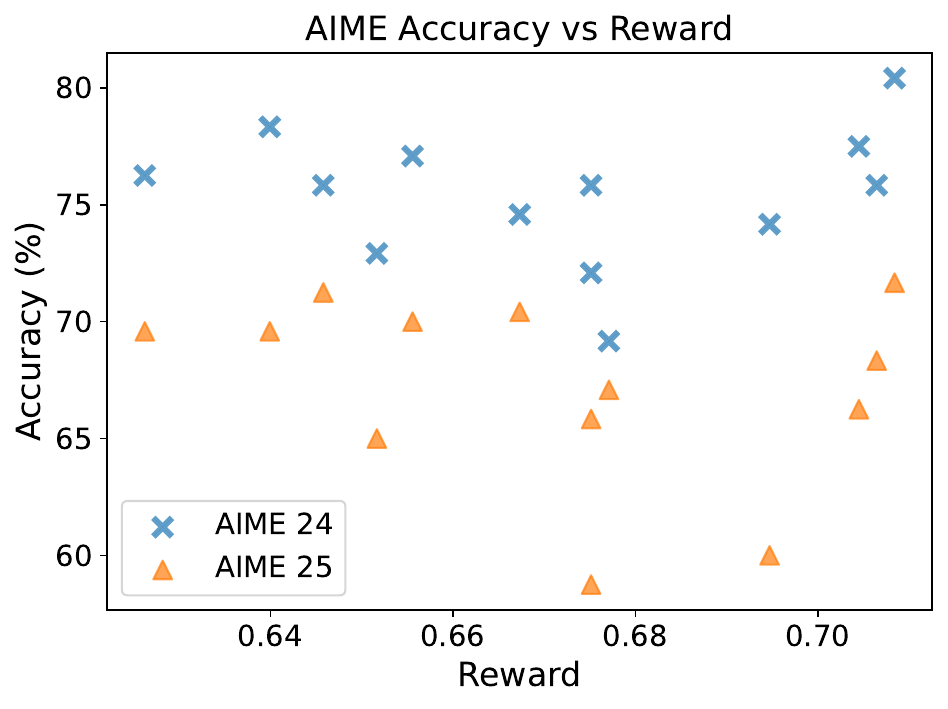}
        \caption{\small Accuracy vs.\ reward}
        \label{fig:acc_reward}
    \end{subfigure}\hfill
    \begin{subfigure}[t]{0.32\linewidth}
        \centering
        \includegraphics[width=\linewidth]{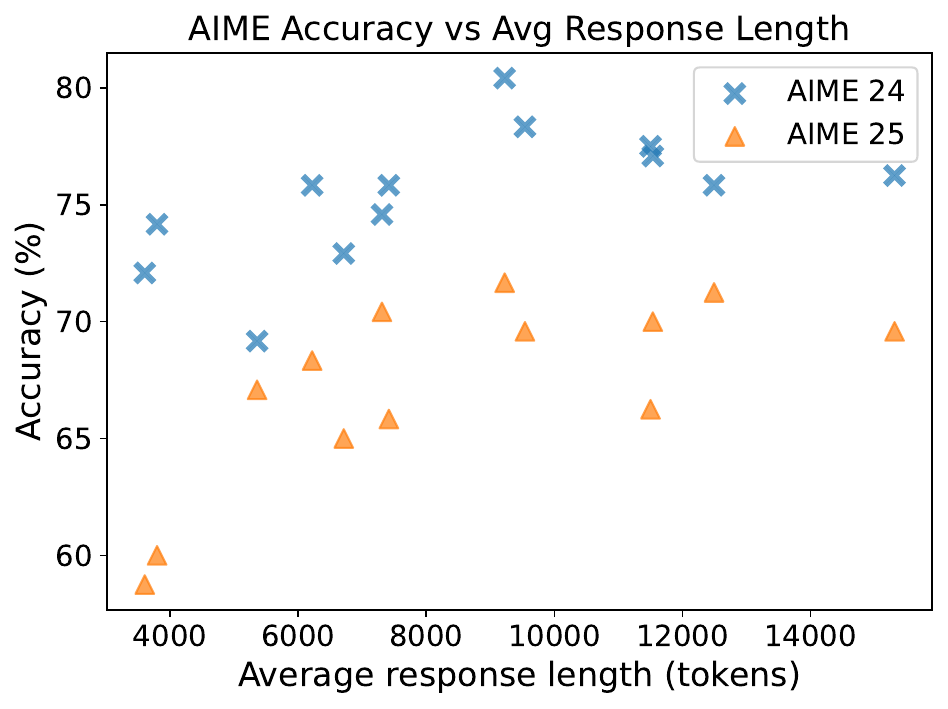}
        \caption{\small Accuracy vs.\ length}
        \label{fig:acc_length}
    \end{subfigure}

    \begin{subfigure}[t]{0.32\linewidth}
        \centering
        \includegraphics[width=\linewidth]{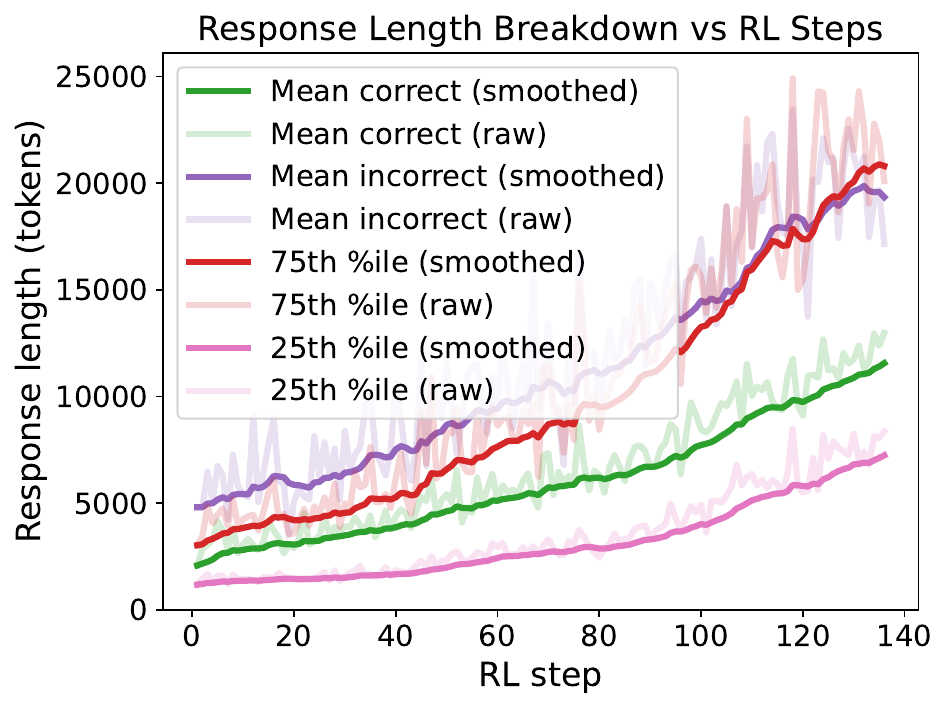}
        \caption{\small Length growth vs.\ RL steps}
        \label{fig:length_constrast}
    \end{subfigure}\hfill
    \begin{subfigure}[t]{0.32\linewidth}
    \centering
    \includegraphics[width=\linewidth]{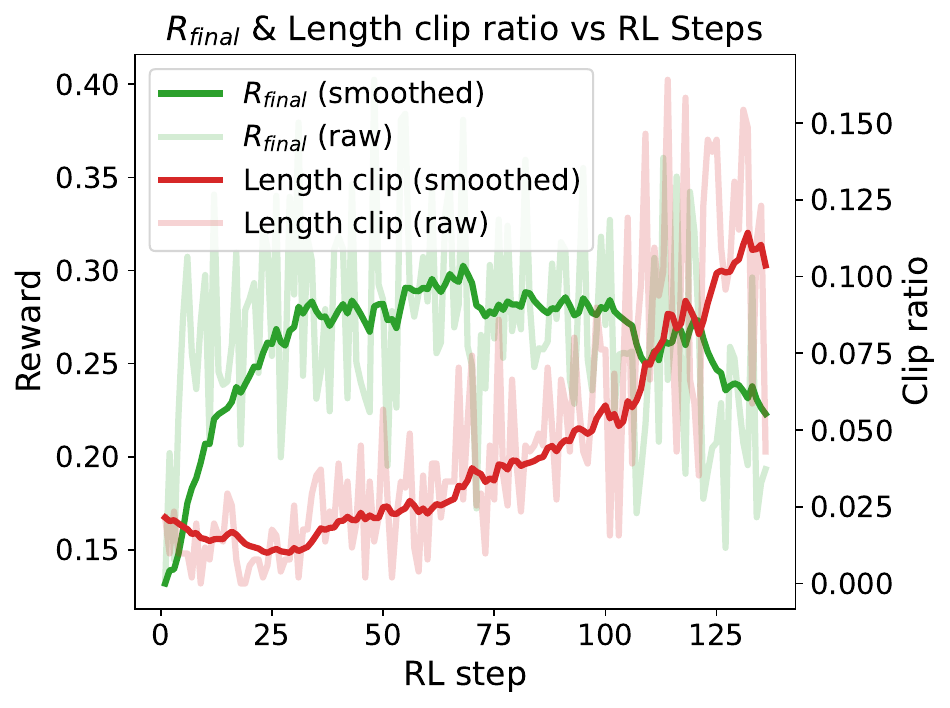}
    \caption{\small Final reward and response length clip‑ratio vs RL steps}
    \label{fig:reward_clip}
    \end{subfigure}\hfill
    \begin{subfigure}[t]{0.32\linewidth}
        \centering
        \includegraphics[width=\linewidth]{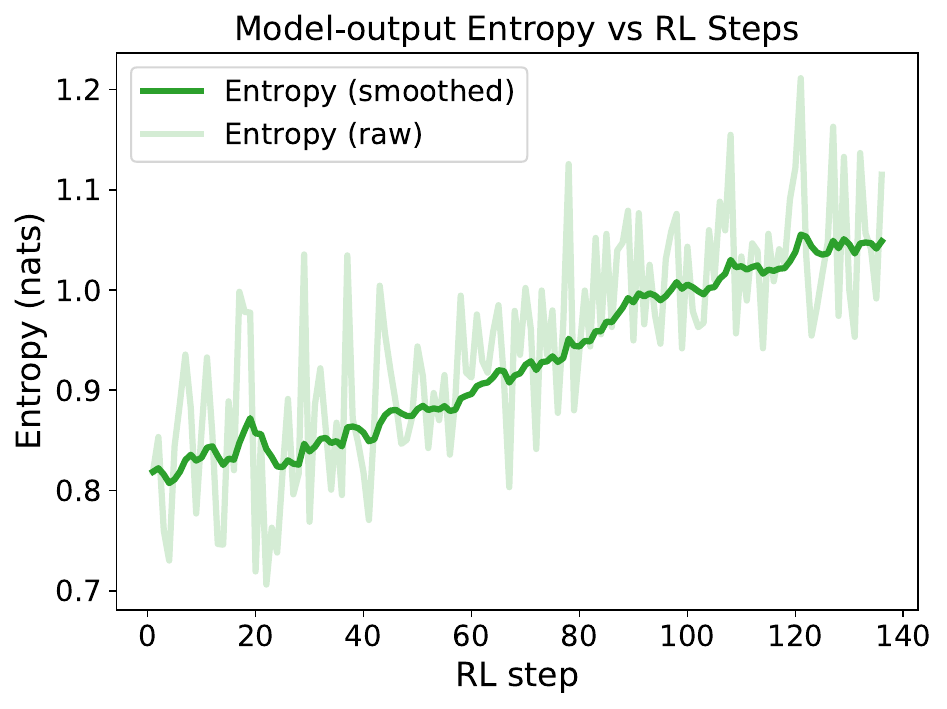}
        \caption{\small Entropy  vs.\ RL steps}
        \label{fig:entropy_steps}
    \end{subfigure}\hfill
    \vspace{-.4em}
    \caption{Behaviour of \modelRL{} during the first $125$ GRPO updates.}
    \label{fig:rl_triptych}
\end{figure*}

 We select as our RL checkpoint  the model with the best observed AIME 2024 score, which is the model trained for 90 steps, over only $\sim6k$ examples (and 8 trajectories of responses per example). We share some of our findings on GRPO training dynamics of \modelRL in Figure \ref{fig:rl_triptych}. 

Starting from a strong SFT model, \ie \model, additional GRPO training for only 90 steps boosts AIME performance by more than $10\%$ (Figure \ref{fig:acc_steps}). Further training for more steps does not translate to additional gains, hinting the potential of an already strong SFT model is near its performance ceiling. A caveat to this observation is the fact that we clip responses beyond $31k$ output tokens during GRPO\footnote{Outputs that are longer than $31k$ are clipped to their first $31k$ tokens, as we reserve $1k$ tokens for the prompt.}, which limits the extent to which GRPO can help. 

We find that throughout GRPO training, the duration of the response is strongly correlated with the performance of the model on AIME, as shown in Fig.~\ref{fig:acc_length}. Moreover, AIME scores seem weakly correlated with reward, for example, see Fig.~\ref{fig:acc_reward}, despite the fact that the model is trained with reward signals primarily from accuracy (Figures \ref{fig:acc_reward} and \ref{fig:acc_length}). 

The effect of growing response length is desired during training with GRPO, as the model learns to spend more test time compute before answering a question, inherently improving its reasoning ability. Figure \ref{fig:length_constrast} further reveals this effect due to our reward model design\footnote{The maximum possible $R_{\text{final}}=8/13\approx 0.62$ by our reward function design.}, where generation length of incorrect answers grows faster than correct ones as the model is rewarded higher to think more before answering when its current answer is incorrect. 

In fact, further improvements to GRPO could potentially be achieved through rejection sampling based solely on response length, particularly for responses significantly exceeding the median length. As illustrated in Fig.~\ref{fig:length_constrast}, during our training runs, responses within the bottom 25th percentile of length increase similarly to the average length of correct responses across RL iterations. In contrast, incorrect responses tend to grow in length more rapidly with each iteration, aligning closely with the 75th percentile of overall response lengths. This divergence suggests that length-based rejection sampling may enhance model efficiency by selectively moderating overly extensive, typically incorrect outputs.

Meanwhile, it is worth noting that due to the maximum sequence length constraint, incorrect answers might not always get corrected once they use up all $31k$ of maximum allowed generation tokens before the model gets a chance to produce a final answer in the end with the \verb|\boxed{}| tag, thus reward plateaus as clipping of excessively long generations goes up (Figure \ref{fig:reward_clip}). 

Despite enforcing length clipping during training, we observe that the model consistently maintains healthy entropy levels, suggesting sustained exploration within its solution space (Figure~\ref{fig:entropy_steps}). We hypothesize that enabling the model to support even longer context windows—such as $64k$ tokens, potentially through interpolation techniques similar to those used to extend context length from $16k$ to $32k$ during SFT or alternative RoPE interpolation methods~\cite{chen2023extending, li2023functional, deepscaler2025}—could yield additional benefits in GRPO training. We leave the exploration of this extended-context approach for future work.

\begin{table}[t]
\centering
\small
\begin{tabular}{@{}l@{}!{\vrule}llllll@{}l@{}}
\toprule
\textbf{Model} & \textbf{AIME 24} & \textbf{AIME 25} & \textbf{HMMT} & \textbf{OmniMath} & \textbf{GPQA-D} & \textbf{LCB} & \textbf{Codeforces}\\
& & &{Feb 2025} & & &  $8/24-1/25$\\
\midrule
\model & 74.6 (5.1) & 63.1 (6.3) & 43.8 (6.2) & 76.6 (0.5) & 67.1 (2.7) & 53.8 & {\underline{1736}}\\
\modelRL & {\bf 81.3} (1.8) & {\bf 78.0} (4.6) & {\bf 53.6} (6.3) & \underline{81.9} (0.1) & \underline{69.3} (2.1) & 53.1 &1723 \\
OpenThinker2‑32B. & 58.0 & 58.0 & --- & --- & 64.1 & --- & --- \\
QwQ 32B & \underline{79.5} & 65.8 & \underline{47.5} & --- & 59.5 & $\underline{63.4}_{\{8/24-2/25\}}$ & ---\\
EXAONE-Deep-32B & 72.1 & 65.8 & --- & --- & 66.1 & $59.5_{\{9/24-2/25\}}$  & ---\\
DeepSeek‑R1‑Distill‑70B & 69.3 (2.7) & 51.5 (5.8) & 33.3 & 63.4 (0.4) & 66.2 (2.4) &57.5 & 1633 \\
DeepSeek‑R1 & 78.7 (3.8) & \underline{70.4} (4.3) & 41.7 & {\bf85.0} (0.6) & {\bf 73.0} (1.7) & \bf{65.9}  & \bf{2029}\\
\hline
o1‑mini & 63.6 & 54.8 & 38.0 (6.2) & 60.5 & 60.0 & 53.8 & 1650 \\
o1 & 74.6 (6.5) & 71.4 (5.7) & 48.3 & 67.5 (0.9) & 76.7 (1.8) & 63.4 & 1891 \\
o3‑mini-high & 88.0 (5.5) & 82.5 (4.9) & 67.5 & {\bf 74.6} (5.1) & 77.7 (0.6) & 68.8 & \textbf{2130} \\
Claude‑3.7‑Sonnet & 55.3 (3.0) & 53 (5.8) & 31.7 & 54.6 (0.9) & 76.8 (1.3) & 52.6 & --- \\
Gemini‑2.5‑Pro & {\bf 92} & {\bf 86.7} & {\bf 82.5} & --- & {\bf 84} & {\bf 69.1}  & --- \\
\bottomrule
\end{tabular}

\caption{Average Pass@1 accuracy (\%) of models on selected reasoning benchmarks. Bold denotes best model per benchmark and model class (i.e., open and closed model weights), and underline denotes the second best.
We report standard deviation in parentheses for all results we produced using Eureka~\cite{balachandran2025inferencetimescalingcomplextasks} and MathArena's~\cite{balunovic_srimatharena_2025} judge/scoring function for HMMT.\protect\footnotemark
}

\label{tab:reasoning_benchmarks_results}
\end{table}

\footnotetext{We use 50 and 64 repetitions for AIME 2025 and HMMT respectively (for statistical robustness) and five repetition for all other benchmarks. Scores for all baselines for LiveCodeBench are from \cite{guo2025deepseek} or LiveCodeBench leaderboard. Sores for HMMT 2025 are from MathArena leaderboards. All other scores for OpenThinker2~\cite{openthoughts}, QwQ~\cite{qwq32b}, EXAONE-Deep~\cite{exaone-deep} and Gemini-2.5-Pro are reproduced from their corresponding reports. Claude-3.7-Sonnet is evaluated with thinking enabled. Codeforces evaluation for Phi models are performed using procedure described in \ref{sec:benchmark_description} while all other Codeforces numbers are from their corresponding reports. Empty cells indicate results not yet reported.}
\section{Evaluation}
\label{sec:eval}
We evaluate our models along two complementary axes: reasoning-specific capabilities and general-purpose capabilities. The primary distinction between these benchmark categories lies in the extent to which the solution to the problems could benefit from step-by-step problem-solving, which is a behavior explicitly encouraged by our training methods. Nevertheless, there exist several general-purpose capabilities that incorporate some form of simpler reasoning among other more salient expected behaviors, for which extended scratchpads with reasoning traces may also help with the final performance. Several of the general-purpose benchmarks may also contain more complex prompts related to math and constraint satisfaction (e.g. MMLUPro, FlenQA etc.). The following empirical results analyze both in order. For both parts, we draw from a rich and diverse set of benchmarks that are still challenging for most state-of-the-art models, with less emphasis on over saturated benchmarks.

\subsection{Reasoning Benchmarks}
\label{sec:benchmark_description}

Among the benchmarks discussed in this report, AIME, MATH, GPQA Diamond, and LiveCodeBench are widely adopted in recent technical reports accompanying major model releases~\citep{O3mini,jaech2024openai,guo2025deepseek}. AIME comprises problems from the American Invitational Mathematics Examination, spanning the years 1983 through 2025, while GPQA includes graduate-level science questions authored by domain experts in biology, physics, and chemistry. The performance on these benchmarks are discussed in Figure~\ref{fig:phi4-main}.

\begin{figure}[t!]
    \centering
    \includegraphics[width=1.0\linewidth]{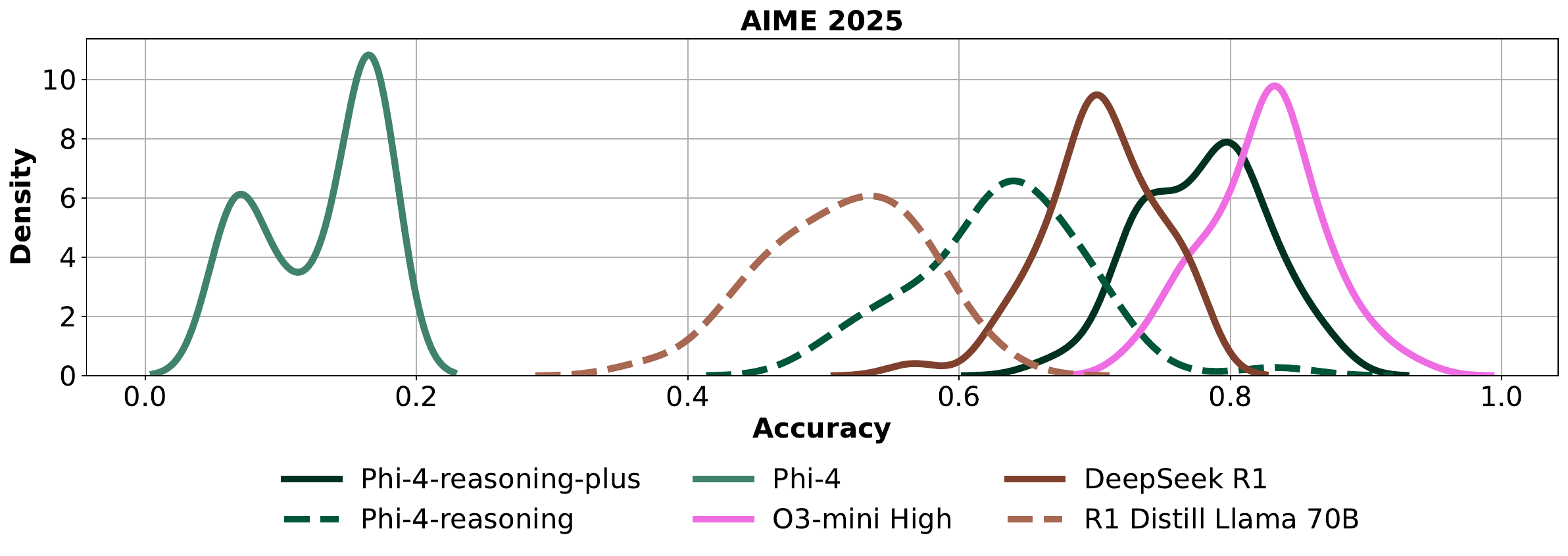}
    \caption{Distribution of pass@1 accuracy on AIME 2025, approximated by kernel density estimation using 50 independent runs with the same prompt and temperature (0.8). We observe a high accuracy variance for all models (\ronedistill ranges from 30\% to 70\%, while \othree's accuracy range is from 70\% to 100\%) suggesting that model comparisons among models using a single run results can be unreliable. The accuracy distribution further indicates the competitive performance of \modelRL, largely intersecting with \othree's distribution and being almost disjoint from the \ronedistill's distribution.}

    \label{fig:aime_50runs}
 \end{figure}

The subset of AIME for the year 2025 is particularly interesting as it  was released after the finalization of training data for \model{}, ensuring it remains fully contamination-free. However, this benchmark contains only 30 problems, which makes evaluation particularly sensitive to sampling variance—especially at higher decoding temperatures commonly used in reasoning models. For all models including ours and those in the OpenAI and DeepSeek family, average-of-5 results from two independent runs can differ significantly by up to 5-10 percentage points on AIME 2025, thus making side-by-side comparison of models unreliable. To mitigate this issue and increase the statistical robustness of results, we report pass@1 accuracy averaged over 50 independent runs in Table~\ref{tab:reasoning_benchmarks_results} and Figure~\ref{fig:phi4-main}. Further detailed analysis and visualizations of AIME 2025 performance, including comparisons between \model{} and \modelRL{} are provided in Figure~\ref{fig:aime_50runs} and Section~\ref{sec:acc_distribution_aime25}.

To assess reasoning ability more broadly, we adopt a comprehensive suite of benchmarks from~\cite{balachandran2025inferencetimescalingcomplextasks}. Omni-MATH~\citep{gao2024omni} includes over 4000 olympiad-level problems with rigorous human annotations, covering a wide range of topics and problem types. We also include two new benchmarks, 3SAT and TSP~\citep{balachandran2025inferencetimescalingcomplextasks} for studying the ability of models to solve NP-hard problems using symbolic and combinatorial reasoning~\citep{papadimitriou2003computational,hartmanis1982computers}. In addition, we evaluate on BA-Calendar~\citep{butt2024benchagents}, a calendar planning benchmark that requires models to find a common time slot among participants while considering constraints beyond availability, such as time zones, buffer time, priority, etc. Finally, we include two spatial reasoning benchmarks: Maze and SpatialMap~\citep{wang2024picture}. Maze consists of multiple choice questions such as counting the number of turns or determining the spatial relationships between two points in a given maze, and we use the $10\times10$ version of the benchmark. SpatialMap evaluates relational reasoning by asking about spatial relations between objects or counts satisfying geometric constraints. %
 
\begin{figure}[t]
    \centering
    \begin{subfigure}[b]{0.3\textwidth}
        \centering
        \includegraphics[width=\textwidth]{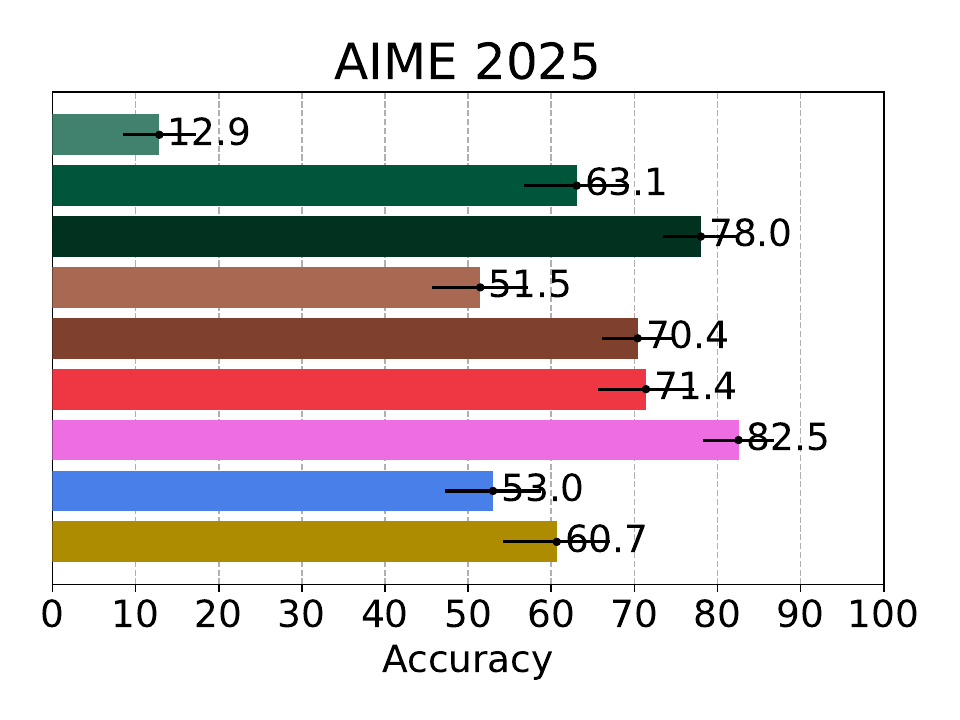}
    \end{subfigure}
    \begin{subfigure}[b]{0.3\textwidth}
        \centering
        \includegraphics[width=\textwidth]{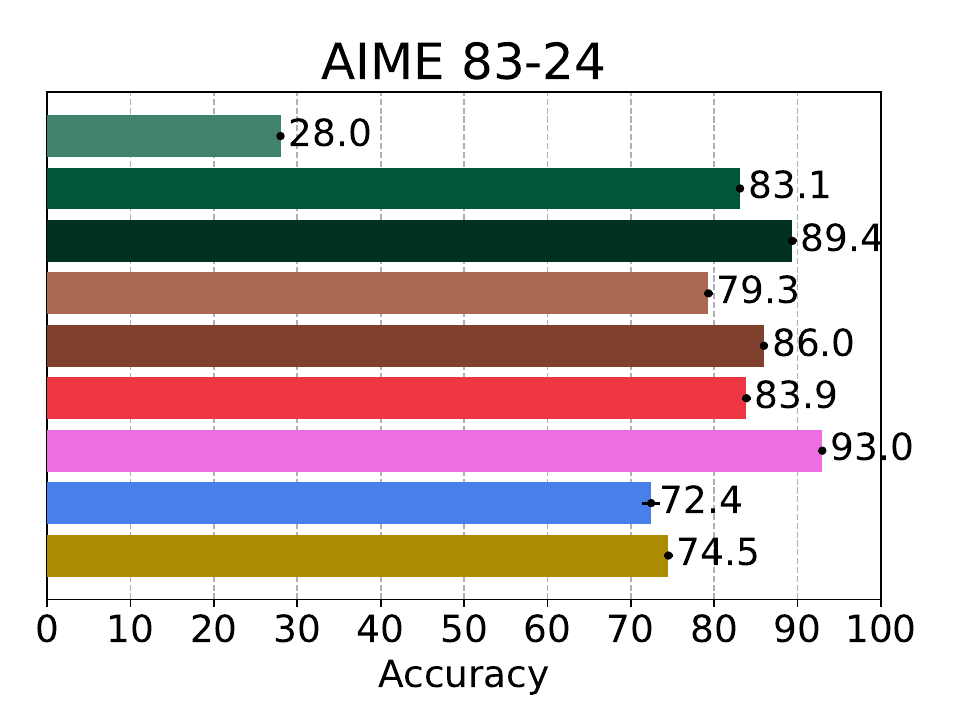}
    \end{subfigure}
    \begin{subfigure}[b]{0.3\textwidth}
        \centering
        \includegraphics[width=\textwidth]{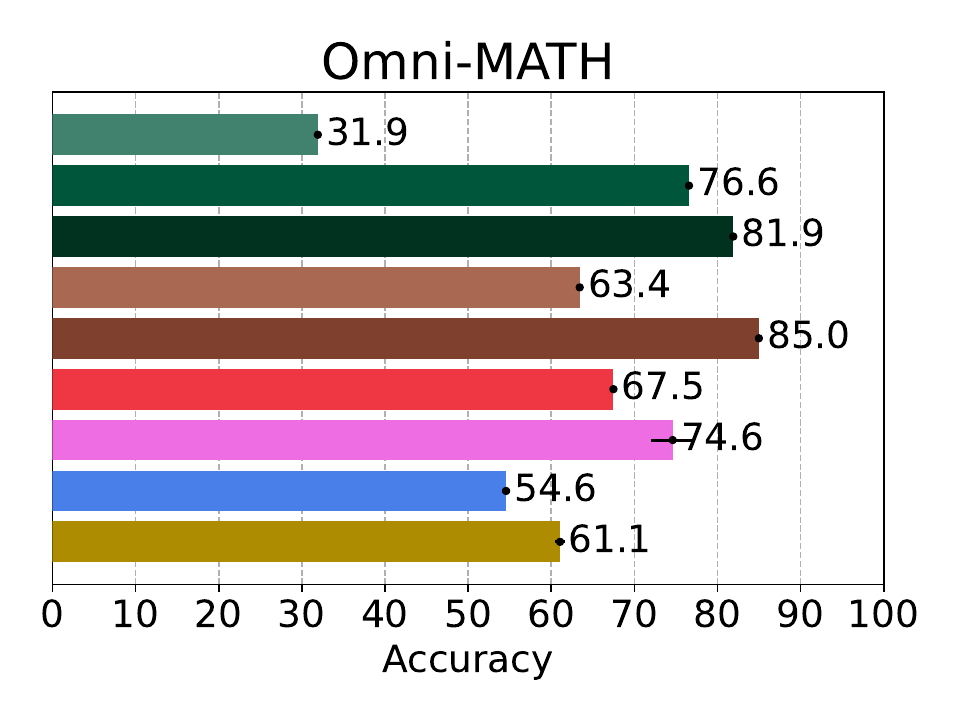}
    \end{subfigure} 
    \begin{subfigure}[b]{0.3\textwidth}
        \centering
        \includegraphics[width=\textwidth]{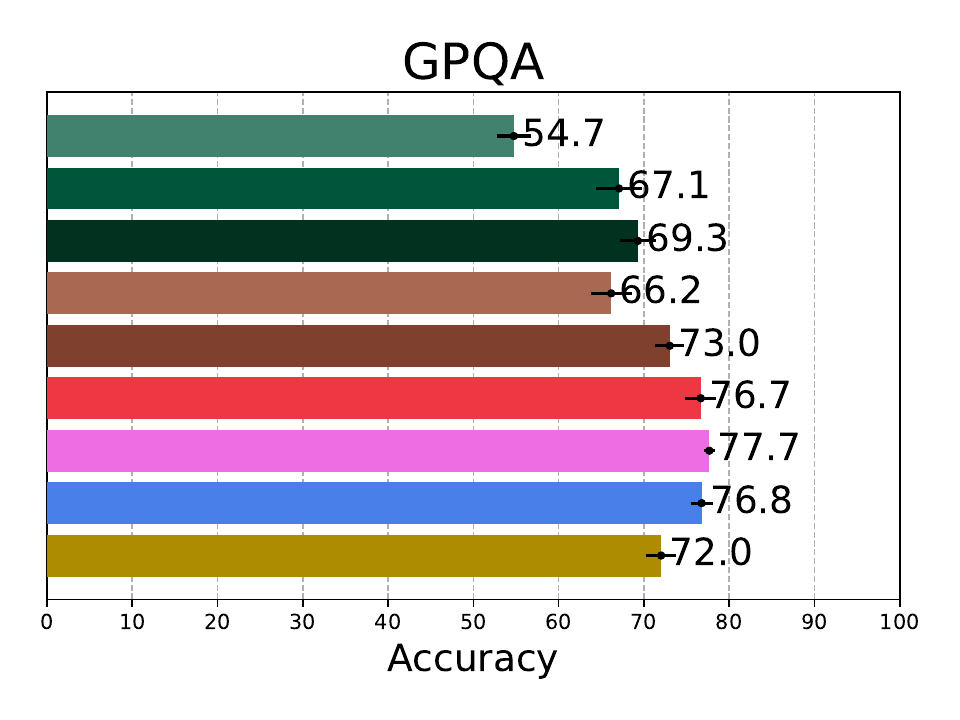}
    \end{subfigure} 
    \begin{subfigure}[b]{0.3\textwidth}
        \centering
        \includegraphics[width=\textwidth]{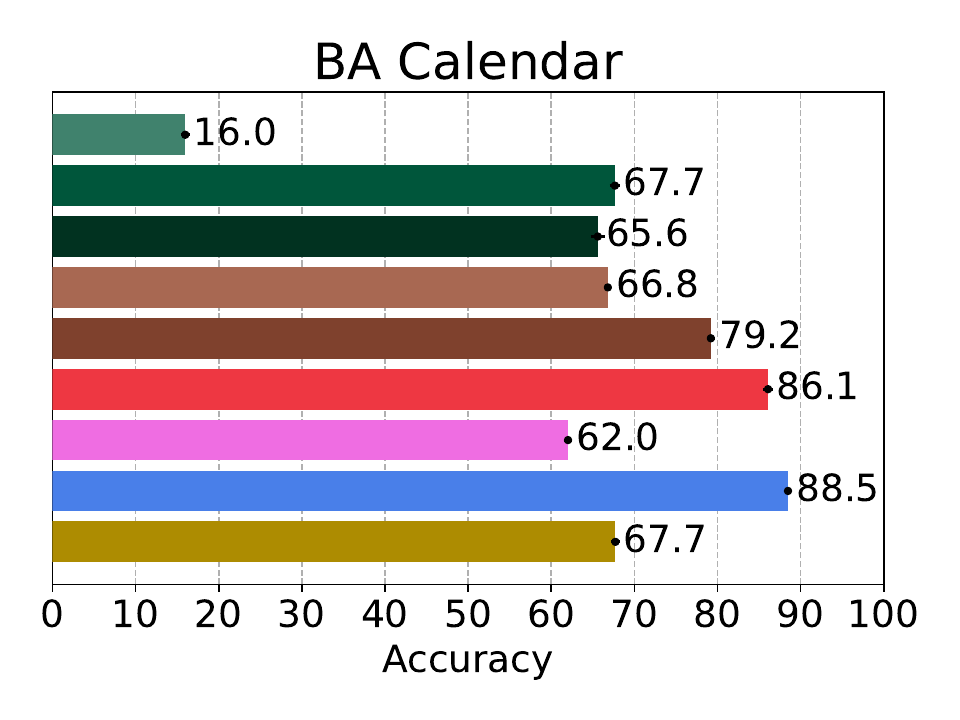}
    \end{subfigure}
        \begin{subfigure}[b]{0.3\textwidth}
        \centering
        \includegraphics[width=\textwidth]{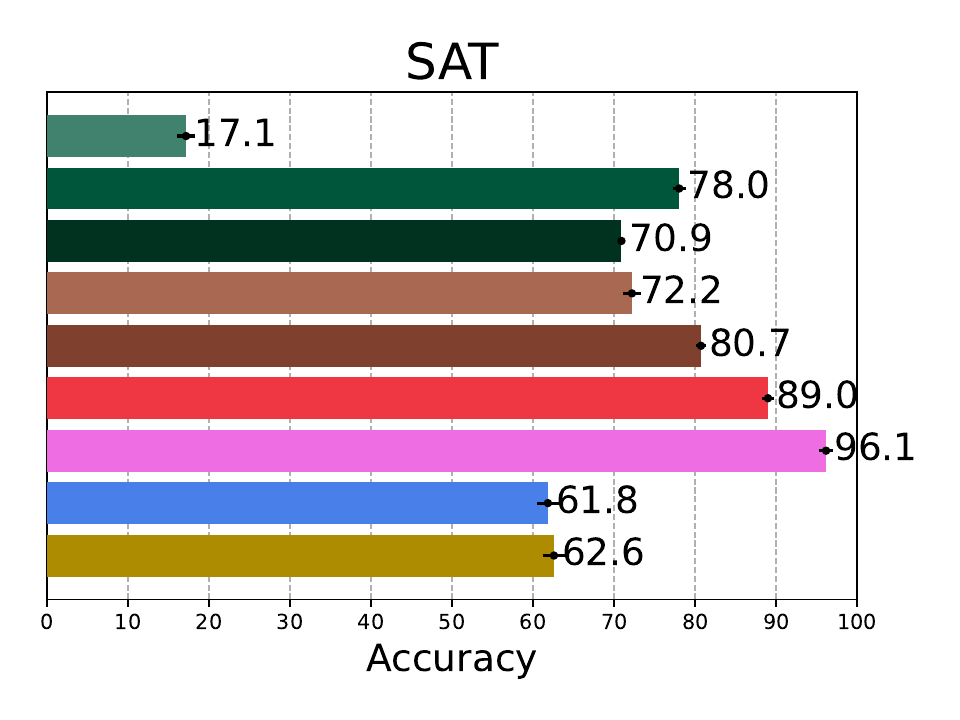}
    \end{subfigure} 
    \begin{subfigure}[b]{0.3\textwidth}
        \centering
        \includegraphics[width=\textwidth]{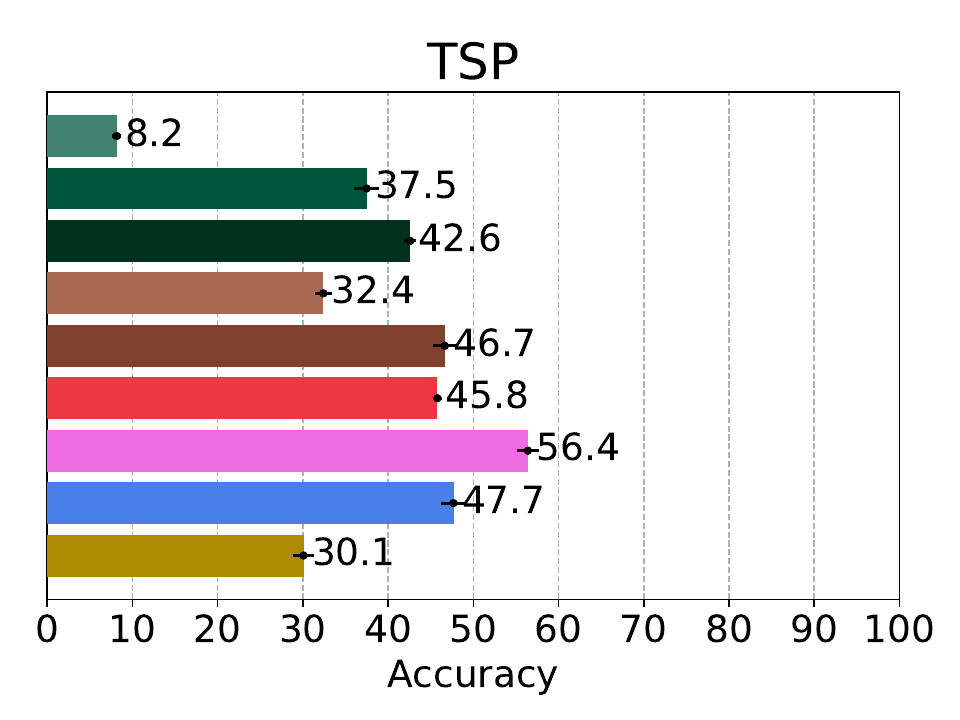}
    \end{subfigure}     
    \begin{subfigure}[b]{0.3\textwidth}
        \centering
        \includegraphics[width=\textwidth]{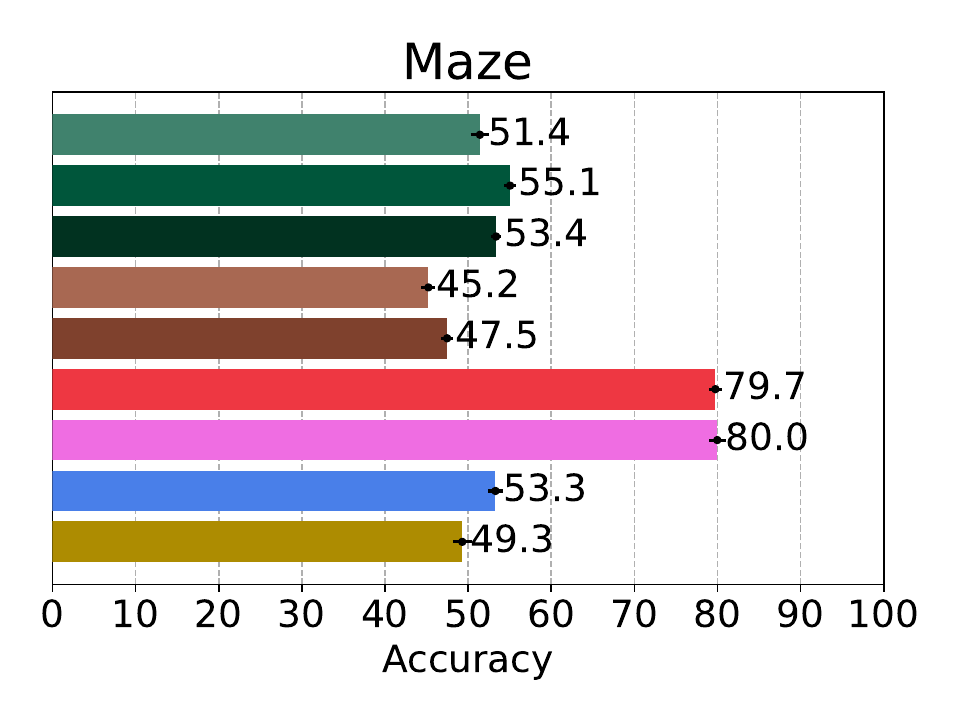}
    \end{subfigure}    
    \begin{subfigure}[b]{0.3\textwidth}
        \centering
        \includegraphics[width=\textwidth]{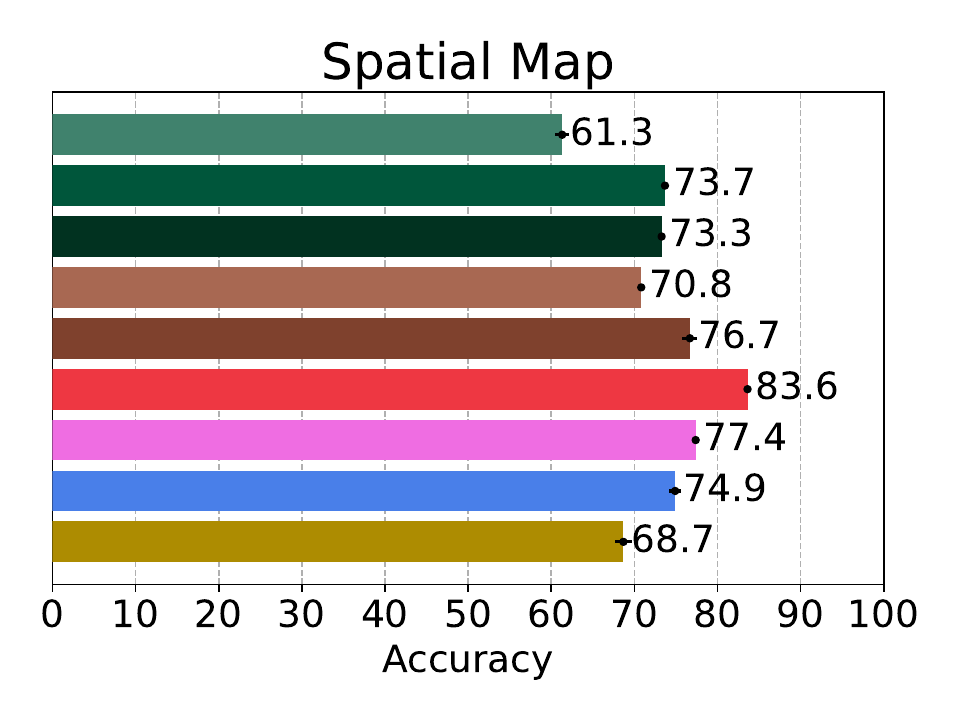}
    \end{subfigure}     
    \begin{subfigure}[b]
    {\textwidth}
        \centering
\includegraphics[width=\textwidth]{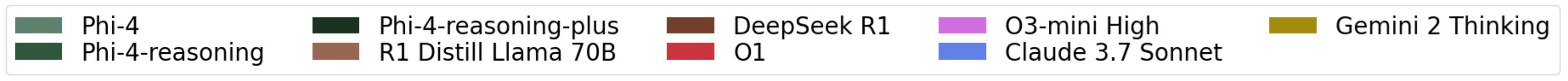}
    \end{subfigure}
    \caption{Average Pass@1 model performance across eight reasoning tasks across five independent runs (generations).  We use larger number of generations for AIME 2025 and GPQA for more statistical robustness.\protect\footnotemark }
    \label{fig:all_in_one}
\end{figure}
\footnotetext{We use 50 and 64 generations for AIME 2025 and GPQA respectively and provide more analysis of variance in Figure~\ref{fig:aime_50runs}}

For all these results, we use the implementation pipelines provided in the Eureka ML Insights repository\footnote{https://github.com/microsoft/eureka-ml-insights} to evaluate all models. Eureka ML Insights is a reusable and open evaluation framework for standardizing evaluations of large foundation models beyond single-score reporting and rankings. 
Note that the evaluation results on the baseline models were recently reported in~\cite{balachandran2025inferencetimescalingcomplextasks}, we reuse the same evaluation logs for baseline comparisons as well as the same evaluation methodology for studying \model and \modelRL. 

The full list of reasoning benchmarks used in this report is summarized in Table~\ref{tab:datasets_reasoning_repro}. The performance of \model and \modelRL{} on this comprehensive suite of benchmarks along with select baselines are shown in Figure~\ref{fig:all_in_one}.

\subsubsection{Baseline models} 
We compare our models with selected state-of-the-art models as described in Table~\ref{tab:models_repro}, including \oone, \othree-high, \rone, \ronedistill, \claude, and \geminiflash. We use temperature 0.8 for the Phi models, 0.6 for models in the DeepSeek family as recommended in the corresponding model cards on HuggingFace, and otherwise use either 1.0 or the default temperature setting in the model APIs. In terms of maximum token length, we aim to allow as many tokens as possible to all models, within limits that do not present other experimental hurdles such as timeouts from the API. For both our models we use the system message described in Section 3. We use the same CoT prompt templates for all models on all benchmarks, which explicitly require models to think step by step and then provide an answer. For \oone{} evaluations we use a plain non-CoT prompt template because of policy violation refusals usually triggered by CoT requests. 

In addition to the baselines in Figure~\ref{fig:all_in_one}, we report comparison with evaluations in newer baselines including OpenThinker2~\cite{openthoughts}, QwQ~\cite{qwq32b}, EXAONE-Deep~\cite{exaone-deep}, and Gemini-2.5-Pro \cite{GeminiFlash} in Table~\ref{tab:reasoning_benchmarks_results}.

\subsubsection{Accuracy distribution on AIME 2025: Beyond single-score analyses} 
\label{sec:acc_distribution_aime25}
Most existing comparisons of reasoning models on AIME 2025 use the average accuracy computed on a single pass over the dataset. However, LLMs have exhibited large generation nondeterminism, \ie they may produce substantially different answers given the same prompts and inference parameters (such as temperature and max tokens)~\cite{balachandran2024eureka,balachandran2025inferencetimescalingcomplextasks,hochlehnert2025sober}. For larger non-reasoning models, nondeterminism can occur even at very low temperature (even zero), and the phenomenon is also almost always expected for reasoning models which are expected to diversify the inference paths and also highly recommended to be run at high temperatures ($0.6\sim1.0$). Given that AIME 2025 also only contains 30 questions, nondeterminism renders the accuracy-based analysis questionable. 

To account for the stochastic nature of such experiments, we study the accuracy distribution on AIME 2025, approximated by kernel density estimation using 50 independent runs with the same prompt and temperature (see Table~\ref{tab:models_repro} for temperature details). We have found several interesting observations as shown in Figure \ref{fig:aime_50runs}. First, we observe a high accuracy variance for all models. For example, accuracy of answers generated by \ronedistill ranges from 30\% to 70\%, while \othree's accuracy range is from 70\% to 100\%. This suggests that any comparison among models using a single run can easily produce misleading conclusions. Second, models on the two extremes of average accuracy demonstrate more robust accuracy. For example, \modelRL and \phifour have relatively narrower accuracy ranges compared to \ronedistill and \model. Third, the accuracy distribution further indicates the competitive performance of \modelRL, largely intersecting with \othree's distribution and being almost disjoint from \ronedistill's distribution. Figure~\ref{fig:all_in_one} also shows the average pass@1 accuracy for several models including ours along with the standard deviation. Note that given the very small size of the data, this picture can look dramatically different from experiments with fewer runs (e.g., 5) because there may exist variance even between two different sets of 5 independent runs. This behavior also explains differences and discrepancies between different and concurrent evaluation works that may report different scores across five runs~\cite{hochlehnert2025sober,petrov2025proof} which deviate from what model cards report at release time.

We hope that these results will serve as insights for more robust quantitative analyses and as a motivation for moving beyond single-score and single-run accuracy reportings. Throughout this report, we also aimed to scale up our quantitative analysis across a diversified set of benchmarks that have a larger number of samples between 800-4500, except AIME 24 \& 25, HMMT and GPQA which are commonly present in most technical reports of recently released models. Larger benchmarks also enable more disaggregated error analysis that allows for analyzing model performance across different subgroups of data, finding common error patterns, and potentially behavioral correlations between models~\cite{nushi2018towards,barocas2021designing,moayeri2024unearthing}.

At the same time, it is still important to evaluate models on smaller but very challenging benchmarks (\eg HMMT, USAMO, AIME) if that analysis includes several runs, standard deviation and statistical tests, and  preferably qualitative insights. For example, tools like MathArena~\cite{balunovic_srimatharena_2025} and our work on Eureka ML Insights~\cite{balachandran2024eureka} practice the process of revealing not only the scores but also complete evaluation logs per benchmark prompt and per run. In particular, for reasoning models, further work is needed to better understand the variance in model behavior and to study the properties of longer solution scratchpads.

\begin{figure}[bt]
    \centering
    \includegraphics[width=0.8\linewidth]{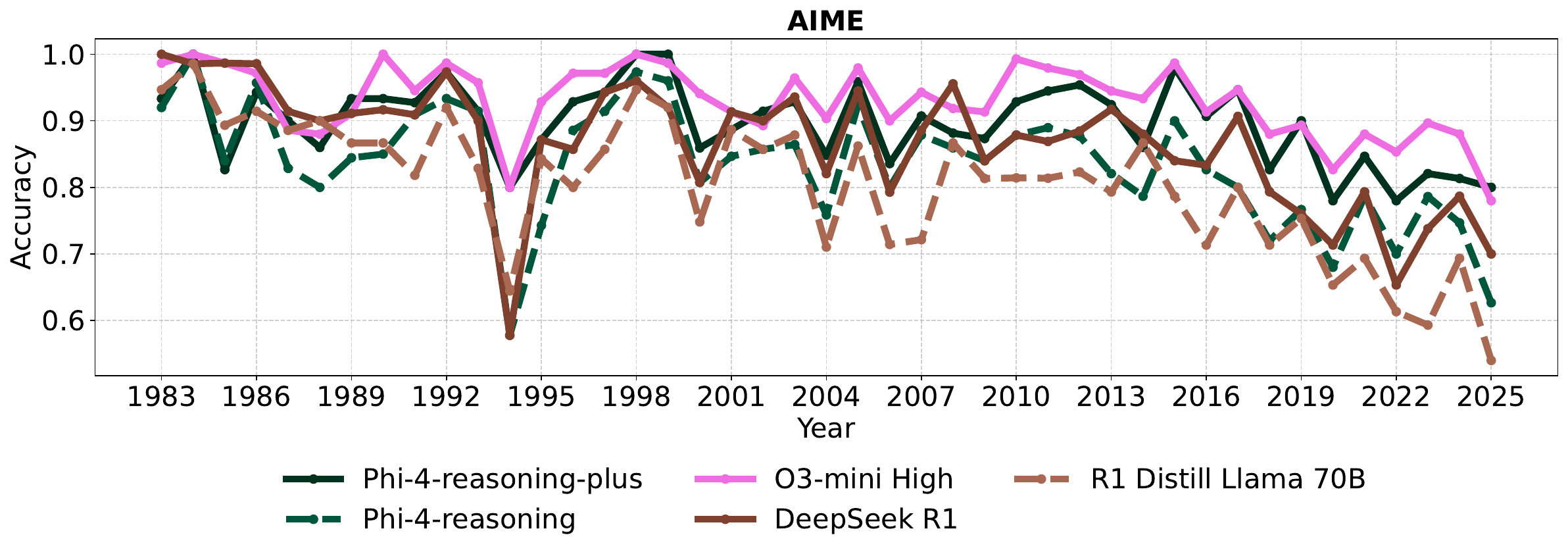}
    \caption{Performance breakdown by years (from 1983 to 2025) for AIME on 5 independent runs. There is a large performance variance across different years. For example, most models perform substantially worse in 1994 and 2025. }
    \label{fig:aime_parallel_8324}
\end{figure} 
\subsubsection{Main findings} 

We present the main findings from the results on the broader set of reasoning benchmarks presented in
Figures~\ref{fig:all_in_one} and Table~\ref{tab:reasoning_benchmarks_results}. We also present complementary analysis that support additional discussions in this section.

\paragraph{Reasoning benchmark performances.} \model and \modelRL both present major improvements over the earlier \phifour{} model on a wide range of reasoning tasks including math and scientific reasoning, coding, algorithmic problem solving, and planning. In particular, the two new models improve by 50\% accuracy on math benchmarks (AIME and Omni-Math) and by over 25\% on coding (LiveCodeBench). Surprisingly, these models also improve by 30\%-60\% on algorithmic and planning problems (Traveling Salesman, Satisfiability, Calendar Planning) which demonstrates increased generalizability on domains that were not targeted in the fine-tuning or RL training for reasoning.

\model and \modelRL are comparable with or have better accuracy than R1 models (\rone and \ronedistill) and \oone/\othree{} models on math reasoning, despite being smaller models with only 14B parameters.  While \oone/\othree{} have very strong performance on AIME which is a popularly reported and small-sized benchmark, the efficacy reduces when considering a more diverse benchmark like Omni-MATH as they struggle in domains like Discrete Math and Geometry (see Figure \ref{fig:omni_math_const_acc}). \model and \modelRL show strong generalization to these diverse math settings as well. This showcases the benefits of rich data synthesis processes based on high-quality seeds and scalable distillation processes, further supported by enhanced exploration at RL stage. They also outperform \claude and \geminiflash on all tasks except GPQA and Calendar Planning.

\modelRL marks important advantages over \model on math (also targeted during the RL stage). More specifically, \modelRL is 15\% more accurate than \model on average across 50 runs on AIME 2025 (Figure~\ref{fig:all_in_one}), and 5\% more accurate on Omni-Math and TSP. This is less pronounced on algorithmic reasoning, planning, coding, and spatial understanding tasks, encouraging further work that enables exploration and verification for a broader set of tasks. For example, we observe that even though the generations of \modelRL are on average across tasks 1.5x longer, on tasks like Calendar Planning, 3SAT, Maze, and SpatialMap, the lengthened traces do not lead to higher accuracy. Nevertheless, a deeper analysis by difficulty level presented in Figure~\ref{fig:tsp__calendar_omnimath_token_accuracy_difficulty} shows that \modelRL is still able to offer improvements for the easy-to-mid levels of difficulty. Accuracy drops towards higher difficulty levels are common for all models, even the most capable ones.

To evaluate the models coding ability, we report LiveCodeBench (2024-08 – 2025-01) and Codeforces (using the \texttt{143} problems from contest IDs 1505 through 1536 inclusive) benchmarks in Table~\ref{tab:reasoning_benchmarks_results}. For Codeforces, we allow each model ten independent submission attempts for every problem and finally compute the Elo rating. To validate comparison against other models, we evaluate the same Codeforces protocol on \ronefourteenb and receive Elo score of \texttt{1481}. Our numbers were within 1 accuracy point for LiveCodeBench and within 20 Elo rating points for CodeForces as reported in~\cite{guo2025deepseek}.
For all other models, we report the Codeforces numbers as published in their corresponding reports for reference, however, they may not be directly comparable due to the lack of public information on their exact evaluation procedure.

The evaluation also reveals several opportunities for improvement for both models we contribute in this report, and for other models evaluated alongside. As previously noted in recent work~\citep{balachandran2025inferencetimescalingcomplextasks}, improvements in scientific topics like biology and chemistry are smaller than for math and physics (Figure~\ref{fig:gpqa_const_acc}). Even within math, all models have lower accuracy on sub topics like discrete math, indicating areas for improvement (Figure~\ref{fig:omni_math_const_acc}). Interestingly, models also follow similar, potentially correlated patterns of errors across years in AIME competitions between 1983-2025 (Figure~\ref{fig:aime_parallel_8324}). Most prominently, all models' performance drops over time and for recent years.

\begin{figure}[ht!]
    \centering
     \begin{subfigure}[b]{0.32\textwidth}
         \centering
         \includegraphics[width=\textwidth]{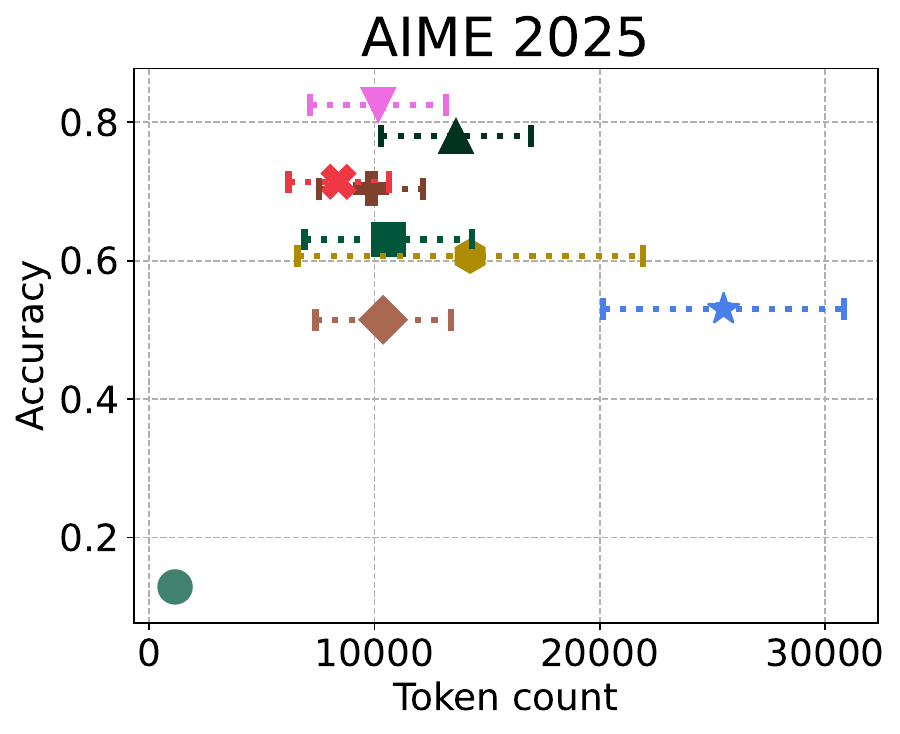}
     \end{subfigure}
        \begin{subfigure}[b]{0.32\textwidth}
        \centering
        \includegraphics[width=\textwidth]{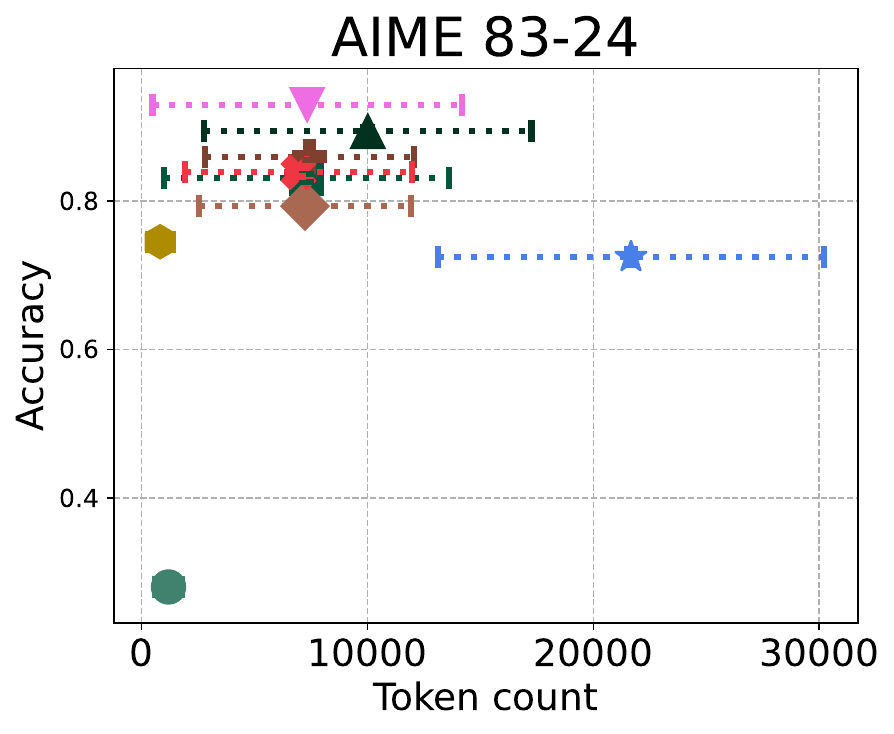}
    \end{subfigure}
    \begin{subfigure}[b]{0.32\textwidth}
        \centering
        \includegraphics[width=\textwidth]{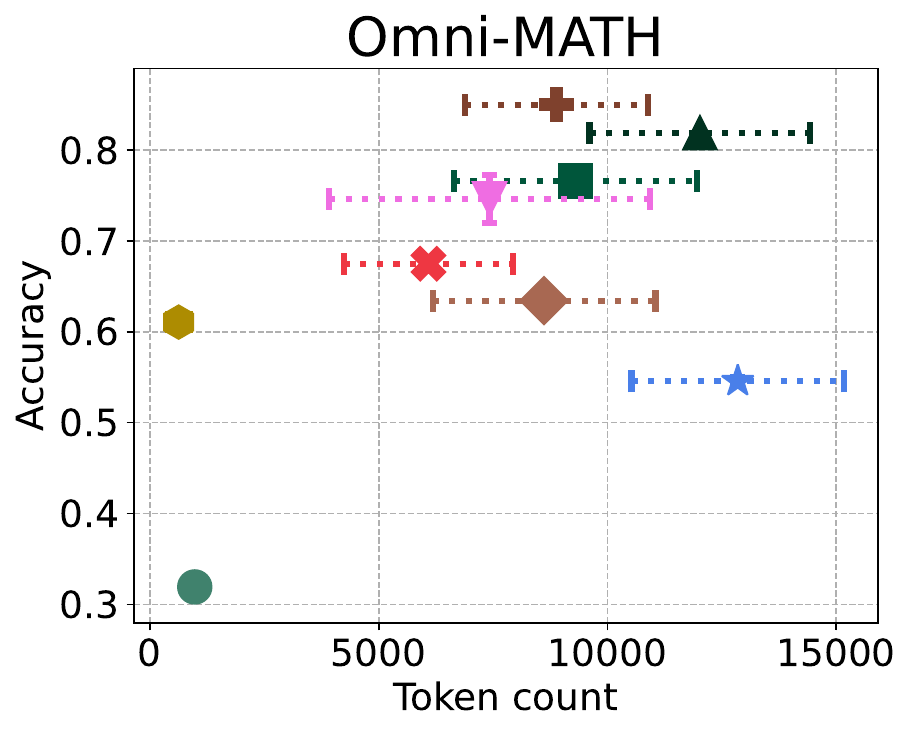}
    \end{subfigure} \\
    \begin{subfigure}[b]{0.32\textwidth}
        \centering
        \includegraphics[width=\textwidth]{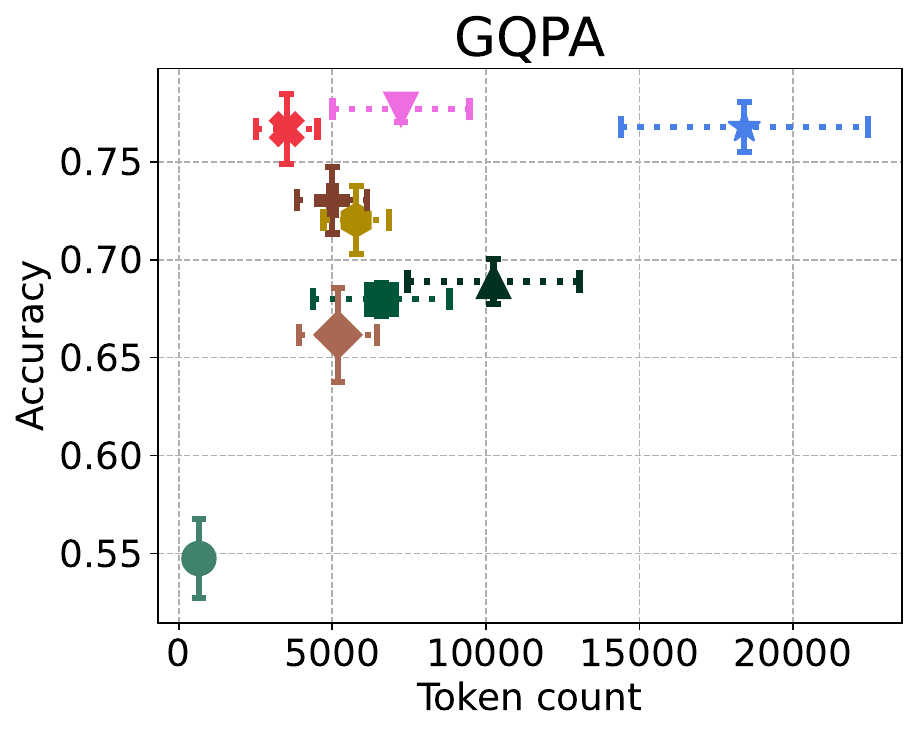}
    \end{subfigure} 
    \begin{subfigure}[b]{0.32\textwidth}
        \centering
        \includegraphics[width=\textwidth]{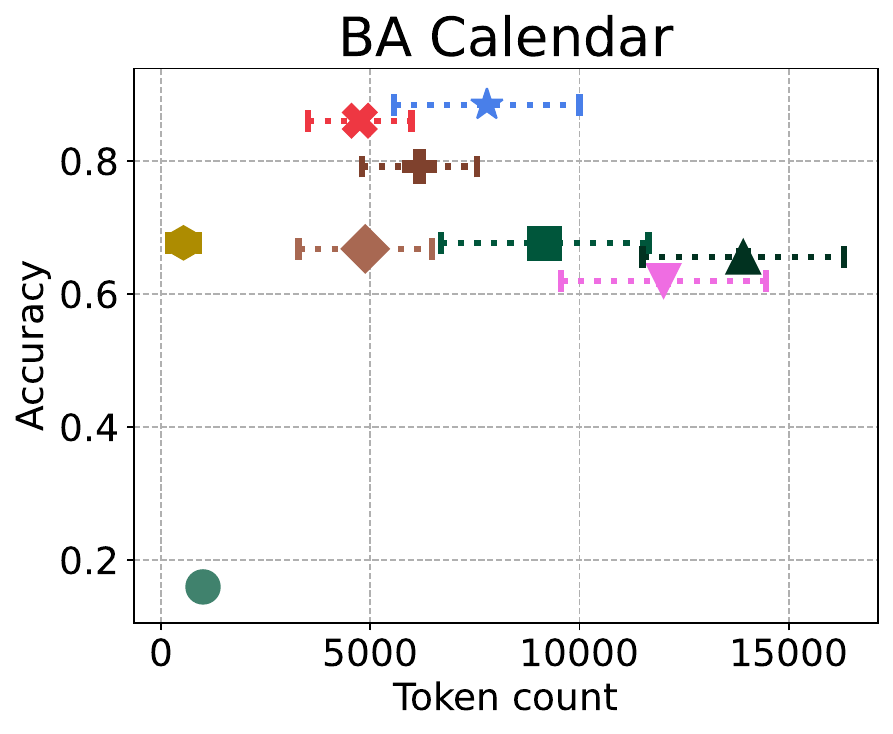}
    \end{subfigure} 
    \begin{subfigure}[b]{0.32\textwidth}
    \centering
        \includegraphics[width=\textwidth]{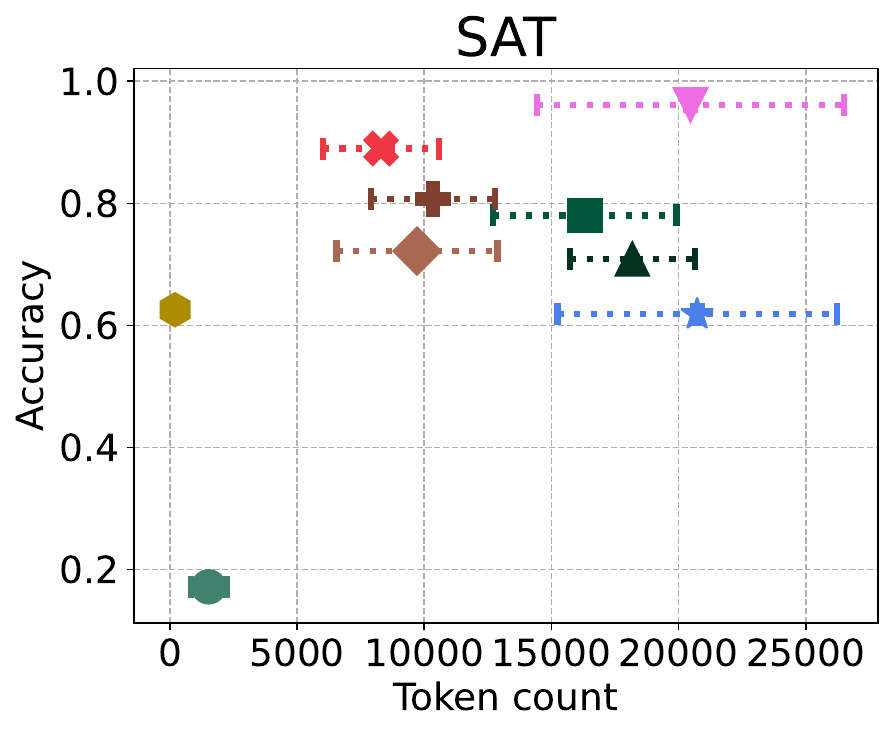}
    \end{subfigure} 
    \begin{subfigure}[b]{0.32\textwidth}
        \centering
        \includegraphics[width=\textwidth]
        {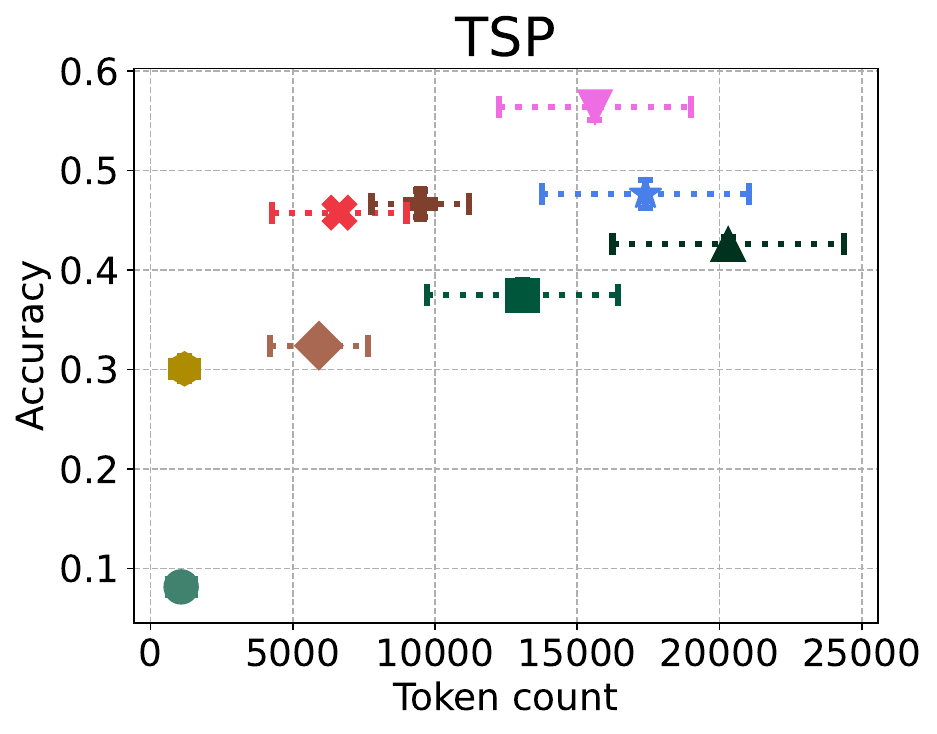}
    \end{subfigure} 
    \begin{subfigure}[b]{0.32\textwidth}
        \centering
        \includegraphics[width=\textwidth]{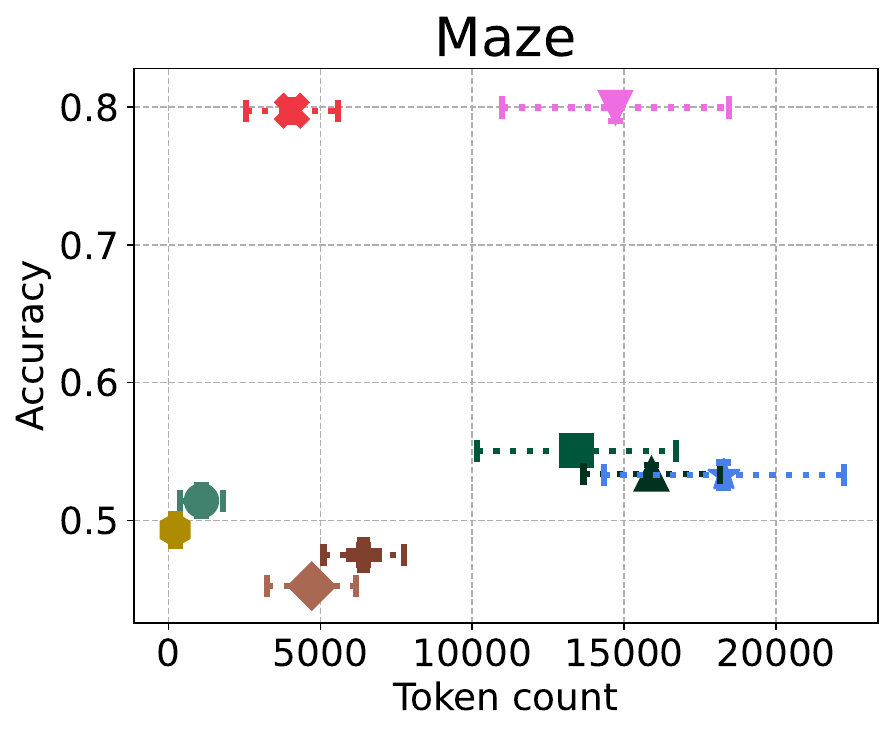}
    \end{subfigure}
    \begin{subfigure}[b]{0.32\textwidth}
        \centering
        \includegraphics[width=\textwidth]{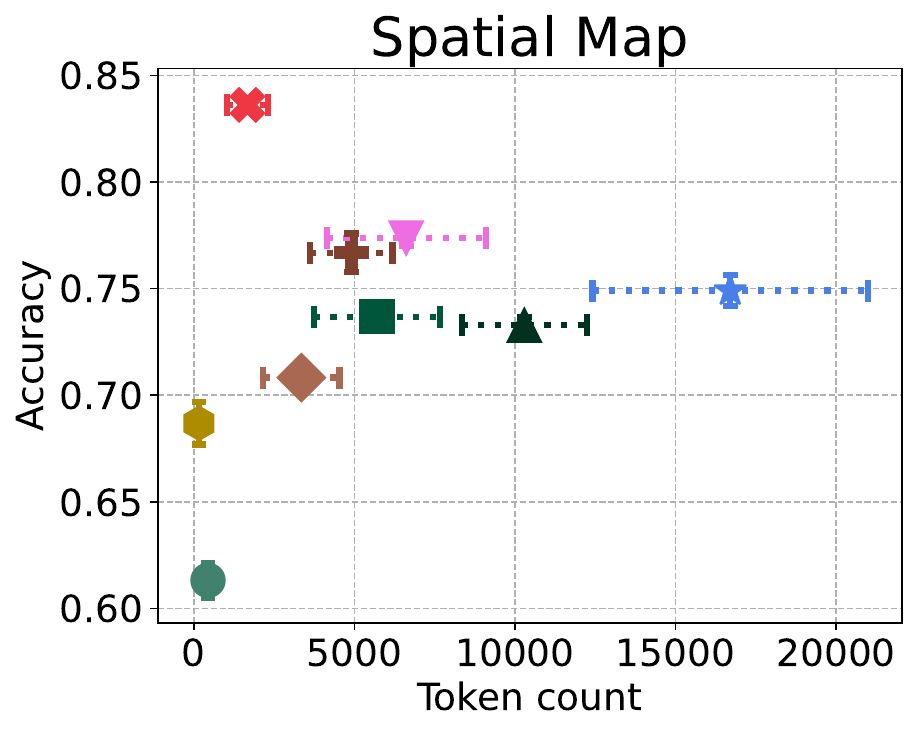}
    \end{subfigure}    
     \begin{subfigure}[b]{\textwidth}
        \centering
\includegraphics[width=1.0\textwidth]{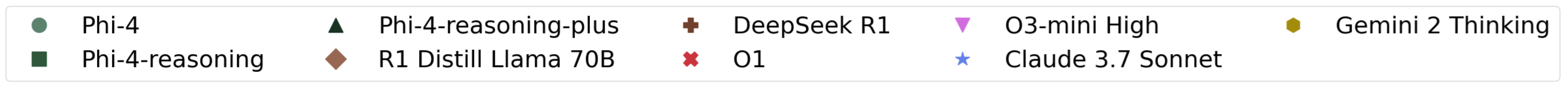}
    \end{subfigure}
    \caption{Tradeoff between accuracy and token usage for all benchmarks. The standard deviation for accuracy (vertical, filled line) is computed across 5 different repetitions (50 for AIME 2025). The standard deviation for token usage (horizontal, dotted line) is computed by first taking the standard deviation per data instance, and then averaging by the size of the benchmark, to show the variability per instance.
    }
    \label{fig:all_in_one_accuracy_tokens}
\end{figure}

\paragraph{Performance vs. token usage tradeoffs.} Results on accuracy vs. token usage tradeoffs are shown in Figure~\ref{fig:all_in_one_accuracy_tokens}. In terms of token length, on average over reasoning benchmarks \modelRL's generations are 1.5x longer than \model, and 1.15x longer than \othree. \model token lengths are similar to \othree. The horizontal error bars in the chart show the standard deviation per instance (prompt), which means that standard deviation was first computed for each prompt, and then averaged across all prompts in the benchmark. This is to show generation length variability for the same prompt, which is comparable for all models.

\subsubsection{Average vs. best-of-5 performance}
\begin{figure}[t]
    \centering
    \begin{subfigure}[b]{0.49\textwidth}
        \centering
        \includegraphics[width=\textwidth]{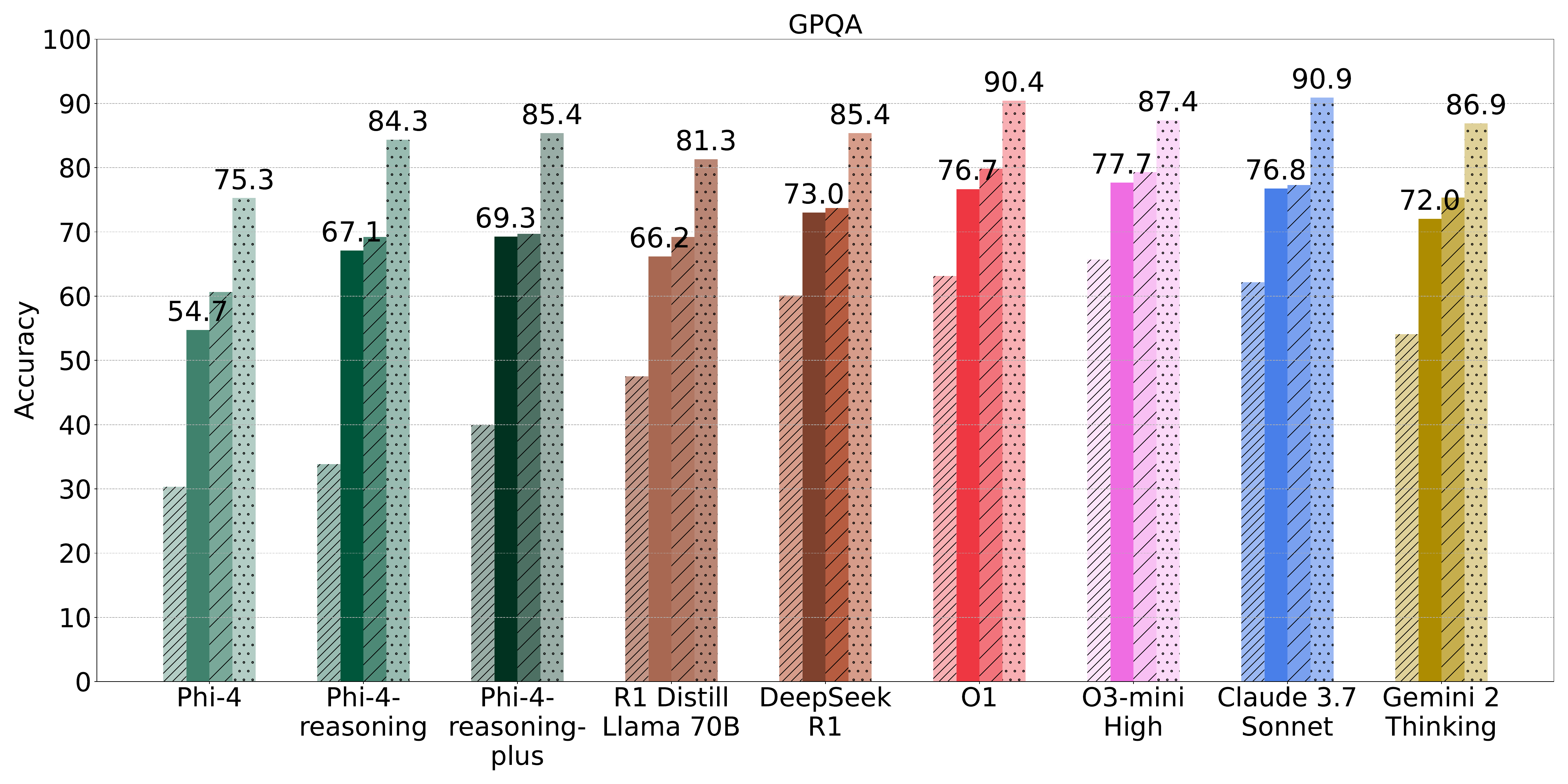}
    \end{subfigure}
    \begin{subfigure}[b]{0.49\textwidth}
        \centering
        \includegraphics[width=\textwidth]{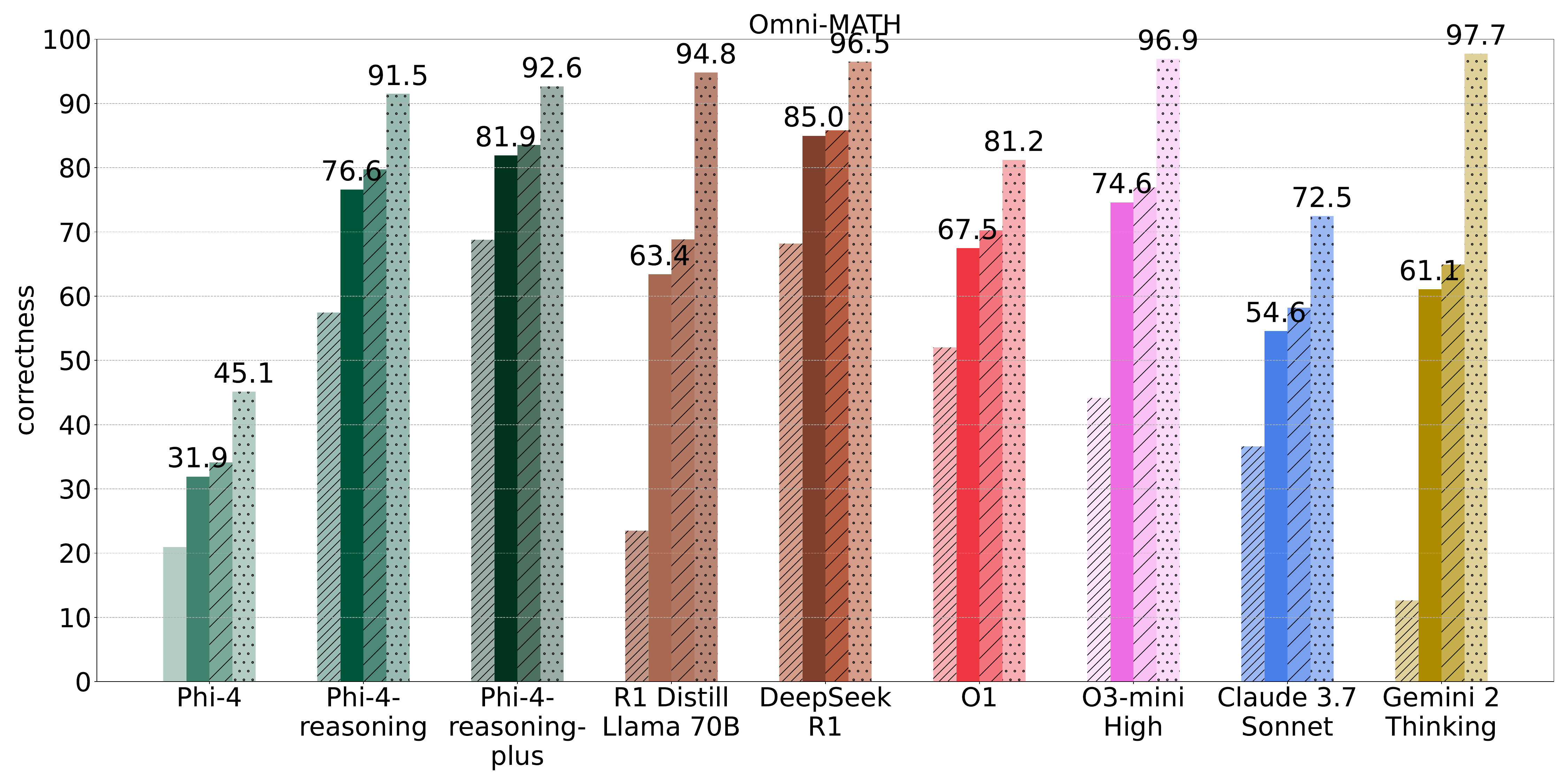}
    \end{subfigure}
    \begin{subfigure}[b]{0.49\textwidth}
        \centering
        \includegraphics[width=\textwidth]{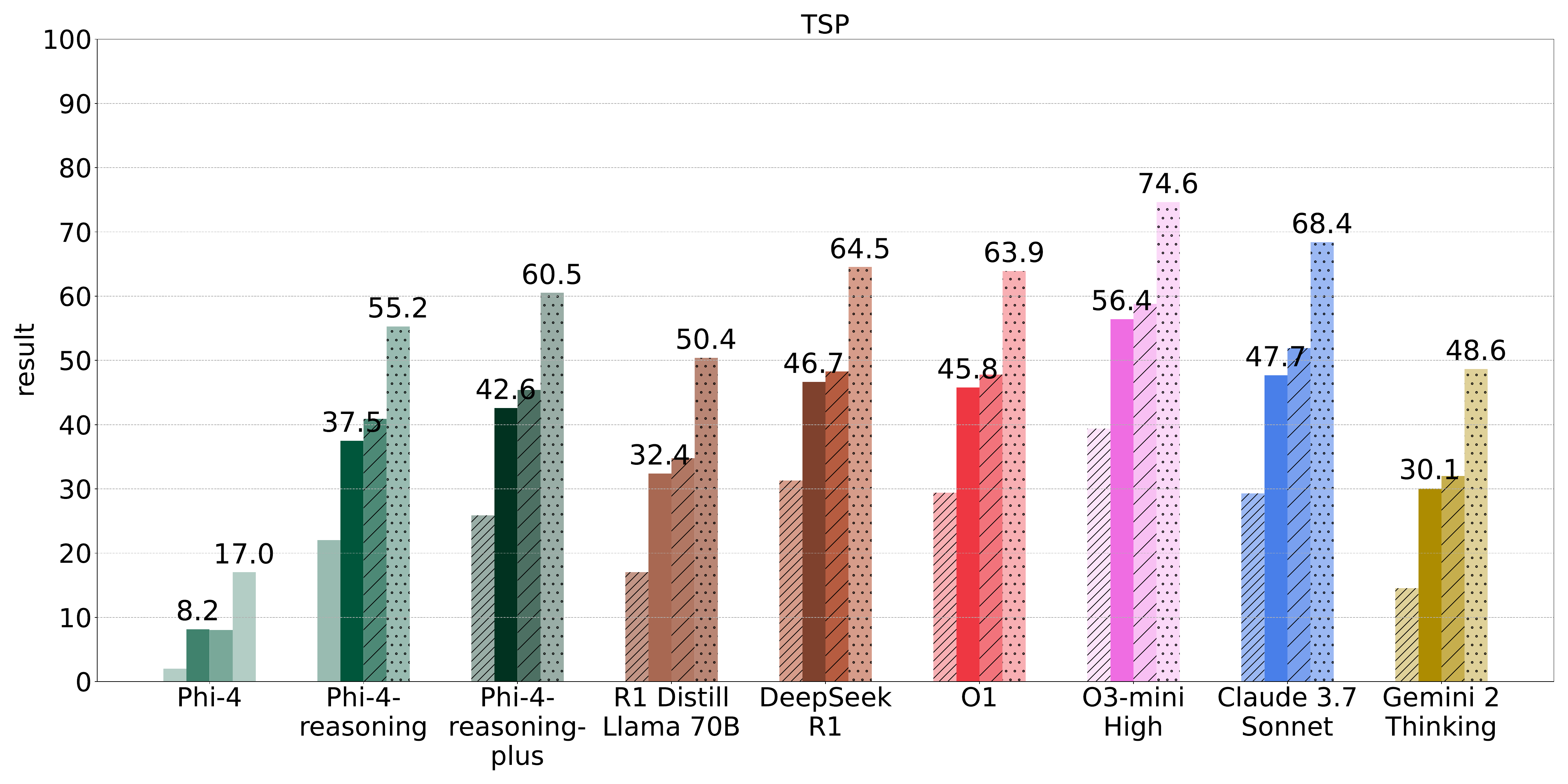}
    \end{subfigure}
    \begin{subfigure}[b]{0.49\textwidth}
        \centering
        \includegraphics[width=\textwidth]{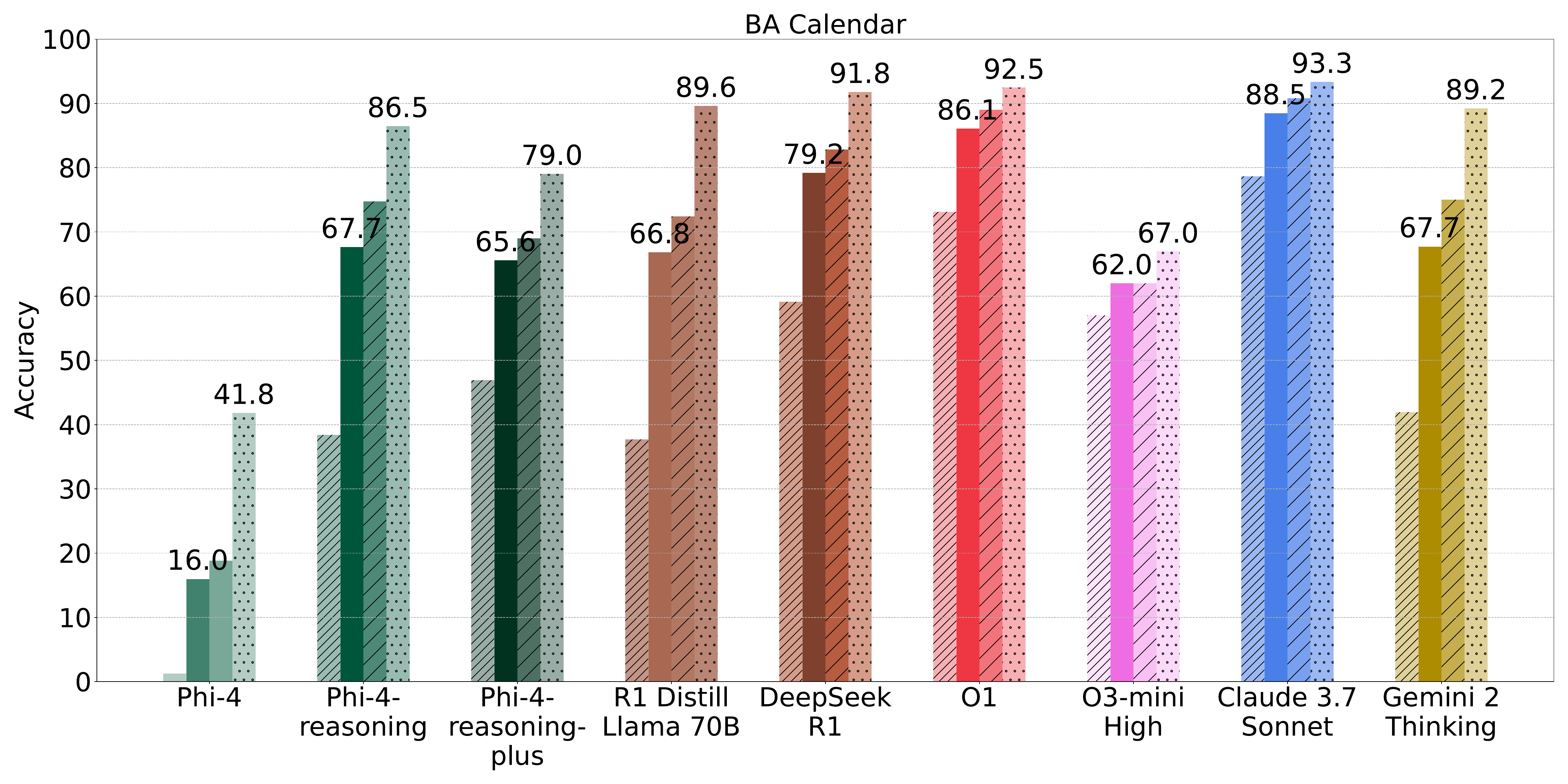}
    \end{subfigure}
    \vspace{-15pt}
    \begin{subfigure}[b]
    {\textwidth}
        \centering
\includegraphics[width=0.5\textwidth]{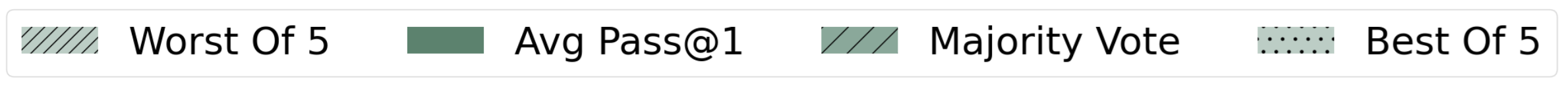}
\includegraphics[width=\linewidth]{figures/model_legend.png}
    \end{subfigure}
    \vspace{-5pt}
    \caption{Results on reasoning benchmarks with different aggregations on 5 independent runs: worst of 5, average pass@1, majority vote, and best of 5. An extended version of this figure can be found in appendix Figure~\ref{fig:main_inf_scaling}.}
    \label{fig:reduced_main_inf_scaling}
\end{figure}

Figures~\ref{fig:reduced_main_inf_scaling} and \ref{fig:main_inf_scaling} show different aggregation scores for five runs on all benchmarks. Comparisons between average accuracy scores across 5 runs and best-of-5 scores show that, similarly to all other reasoning models, there may exist even better trajectories in our models' generations revealing potential for further progress. This also affirms the importance of using improved verifiers at training time that can extract model capabilities which we cannot yet access. While these insights are encouraging, extracting such capabilities without spending N-times more inference compute on models that are already expensive at inference time, remains an open question for future work. Reliably improving accuracy across many samplings from the same prompt remains a caveat for all models, including ours. This is an aspect that can impact user and developer experience on repeated prompts, where determinism and predictable performance are both important.

\subsection{General-purpose Benchmarks}\label{sec:non_reasoning_benchmarks}
\begin{table}[t]
\centering
\small
\begin{tabular}{l!{\vrule}ccc!{\vrule}cc}
\toprule
\textbf{Model} & \textbf{\phifour} & \makecell{\textbf{\model}\\} & \makecell{\textbf{\modelRL}\\} & \textbf{\othree} & \textbf{\gptfouro} \\
\hline
\textbf{FlenQA [3K-token subset] \citep{flenqa}} & 82.0  & \underline{97.7}  & \textbf{97.9} & 96.8 & 90.8  \\
\textbf{IFEval Strict \cite{ifeval}} & 62.3  & 83.4  & \underline{84.9} & \textbf{91.5} & 81.8  \\
\hline
\textbf{ArenaHard \cite{li2024crowdsourced}} & 68.1$\star$ & 73.3  & \underline{79.0} & \textbf{81.9} &69.0$\star$ \\
\textbf{HumanEvalPlus \cite{liu2023codegeneratedchatgptreally}}  & 83.5 & \underline{92.9} & 92.3 & \textbf{94.0} &84.9\\
\textbf{MMLUPro \cite{wang2024mmluprorobustchallengingmultitask}} & 71.5 & 74.3 & \underline{76.0} & \textbf{79.4} &73.5\\
\hline
\textbf{Kitab \citep{AbdinGCLYPNN24}} &  &  &  & & \\
\hspace{10pt}No Context - Precision & 19.3 & 23.2 & 27.6 & \underline{37.9} & \textbf{53.7}  \\
\hspace{10pt}With Context - Precision & 88.5  & \underline{93.8} & 93.6 & \textbf{94.0} & 84.7  \\
\hspace{10pt}No Context - Recall& \underline{8.2}  & 4.9 & 6.3 & 4.2 & \textbf{20.3}  \\
\hspace{10pt}With Context - Recall& 68.1  & 74.8 & \underline{75.4} & \textbf{76.1} & 69.2  \\
\hline
\textbf{Toxigen Discriminative \citep{toxigen}} &   &   &  &  &   \\
\hspace{10pt}Toxic category & 72.6  & \underline{86.7}  & 77.3 & 85.4 & \textbf{87.6}  \\
\hspace{10pt}Neutral category & \underline{90.0}  & 84.7  & \textbf{90.5} & 88.7 & 85.1  \\
\hline
\textbf{PhiBench 2.21 ~\citep{abdin2024phi4technicalreport}}& 58.2 & 70.6 & \underline{74.2} & \textbf{78.0} &73.1\\
\bottomrule
\end{tabular}
\caption{Average pass@1 accuracy of models across general-purpose benchmarks evaluated averaged over five generations. For the results in this table \model and \modelRL{} were evaluated at temperature 0.8, while \phifour{} was evaluated at temperature of 0.0. For ArenaHard$\star$, there is a discrepancy for \phifour{} and GPT-4o compared to the numbers reported in the \phifour{} paper~\cite{abdin2024phi4technicalreport} due to changing the backend LLM-judge. Bold and underlined numbers denote the best and second best scores, respectively, per benchmark.}
\label{tab:results_general_purpose}
\end{table}
In addition to reasoning benchmarks, we also report results in standard benchmarks to ensure training models on reasoning does not degrade their general abilities. First, we use the benchmarks from the \phifour report~\cite{abdin2024phi4technicalreport}. Specifically, we use OpenAI’s simple-evals framework (including prompts, temperature, and extraction) for evaluating the model on MMLU \cite{hendrycks2021measuringmassivemultitasklanguage} and MGSM \cite{shi2022languagemodelsmultilingualchainofthought}. Furthermore, we evaluate our models on MMLU-pro \cite{wang2024mmluprorobustchallengingmultitask}, HumanEvalPlus \cite{liu2023codegeneratedchatgptreally}, and ArenaHard \cite{li2024crowdsourced}. Next, we also use a set of general-purpose benchmarks from the Eureka ML Insights repository, namely FlenQA \citep{flenqa}, Toxigen \citep{toxigen}, Kitab \citep{AbdinGCLYPNN24} and IFEval \cite{ifeval}\footnote{Evaluations of \phifour on Eureka ML Insights were done at temperature=0 and reported for a single run all benchmarks except IFEval which is smaller than the other benchmarks, for which we ran the evaluation three times (for all models) and report the average accuracy.}. Finally, we use PhiBench, our internal
collection of evaluations \cite{abdin2024phi4technicalreport}. See Table \ref{tab:results_general_purpose} for results.

FlenQA~\citep{flenqa} consists of 12K questions ranging in length from 250 to 3000 tokens with True/False labels. This benchmark was designed to isolate the effect of input length on LLMs' performance using multiple versions of the same task, extended with padding of different lengths, types and locations. The task itself involves making inferences on top of two logical statement needles introduced in context, and answering a question that requires both statements. Each prompt is padded with paragraphs sampled from other instances of the same task, or paragraphs sampled from Book Corpus, with key information presented at various locations in the context (at the beginning, at the end, in the middle, or at random locations). Results on this benchmark show that reasoning models, \model, \modelRL, and \othree, are robust to longer inputs compared to conventional models \phifour and \gptfouro, and they are also not affected by the location of key information in the context (See appendix Figure \ref{fig:flenqa_context}). Improvements are potentially linked to the fact that reasoning models are trained to better handle longer context and self-reflection on such context. 

\begin{figure}[t]
\centering
    \includegraphics[width=0.46\textwidth]{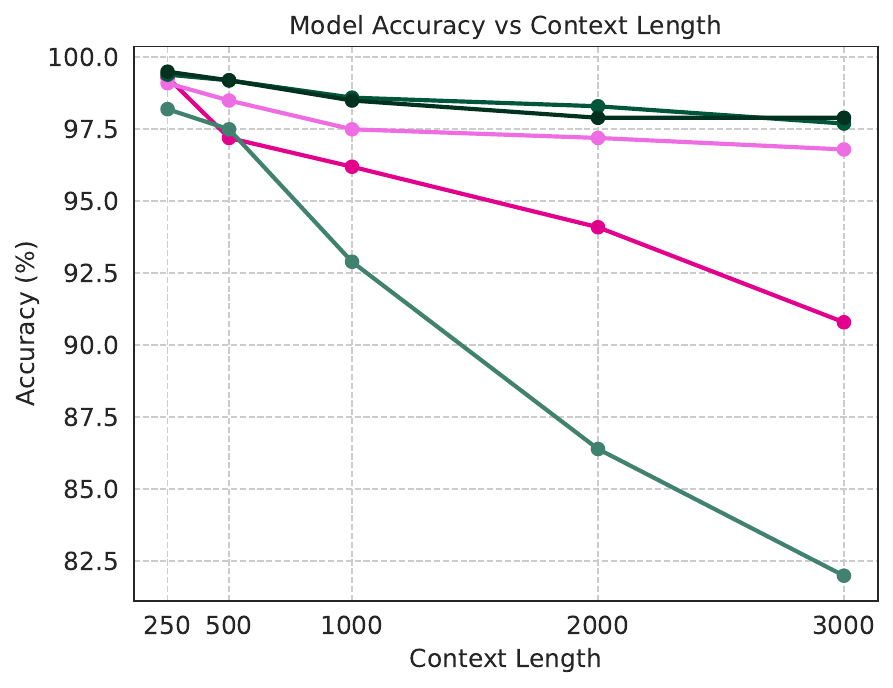}
    \includegraphics[width=0.48\textwidth]{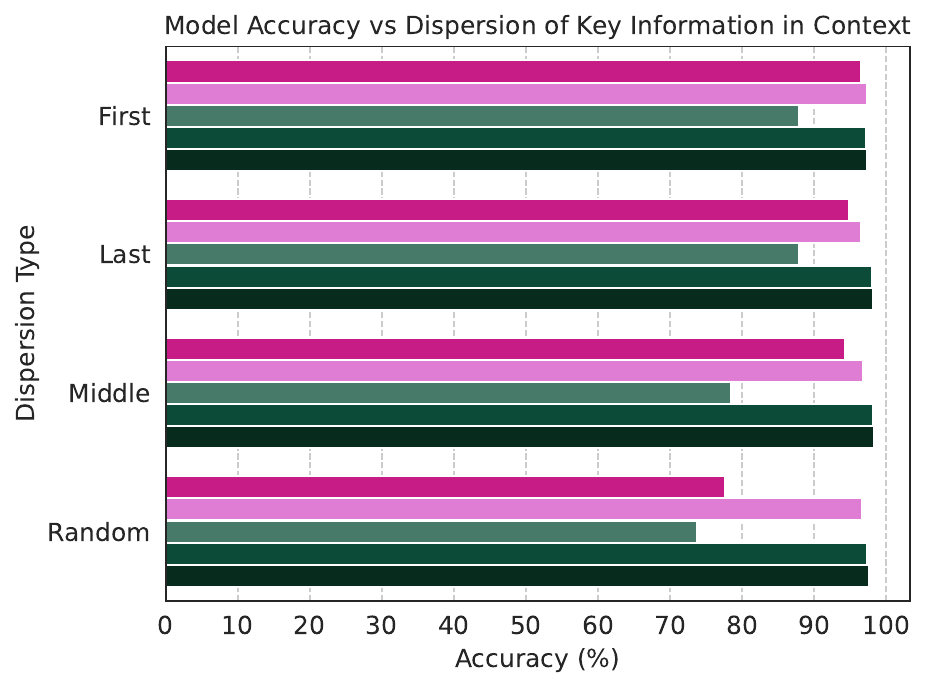}\\
 \includegraphics[width=0.9\textwidth]{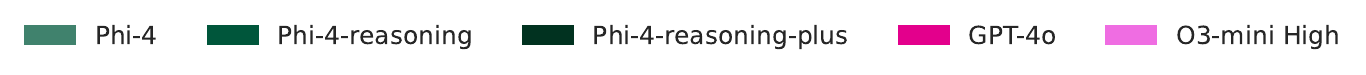}
        \label{fig:flenqa_overall}
    \caption{FlenQA - (Left) Effect of context length on language models' performance. All models suffer a degradation in accuracy with increased context length. Reasoning models are more robust than conventional models in this regard. (Right) Accuracy of language models in the longest context setting (3000 tokens), with key information placed at the beginning (\textit{first}), at the end (\textit{last}), in the middle (\textit{middle}), or dispersed at random locations (\textit{random}) of the context. In the first three dispersion settings, the two pieces of key information are adjacent, while in the \textit{random} setting they are presented separately. Conventional models are remarkably more challenged in the \textit{random} setting compared to other settings while reasoning models are not affected by the dispersion of information.}
\label{fig:flenqa_context}
\end{figure}

Kitab~\cite{AbdinGCLYPNN24} is a challenging information retrieval benchmark containing queries with constraint filters (e.g. List all books written by Isabel Allende, written between 2000-2010.). We performed Kitab evaluations either using the model's parametric knowledge only (no context) or with grounding in a RAG-style setting (with context), only on the subset of queries that contains a single book constraint. The addition of reasoning capabilities to the Phi family models generally seems to improve precision and degrade recall in the no-context setting, while improving both metrics to be almost on-par with \othree when retrieval context is provided. Information retrieval and factuality with only parametric knowledge remain challenging for our models, potentially due to their significantly smaller scale. This remains an avenue for further improvement for all models (even for the larger ones), for teaching them how and when to retrieve information from other sources.

IFEval \citep{ifeval} includes instruction-based prompts that involve instructions amenable to objective verification of compliance. The addition of reasoning capabilities to the Phi family significantly improves performance on this benchmark to even surpass \gptfouro's performance. 

Our reasoning models also lead to improvements over the \phifour model of more than 10\% on ArenaHard, HumanEvalPlus, and our private PhiBench 2.21 data, which is tailored to evaluate diverse abilities that the team found critical to \phifour development~\citep{abdin2024phi4technicalreport}. There is also a 3\%-5\% improvement on MMLUPro. As we mentioned earlier in this section, while these benchmarks are more for general purpose evaluation, it is not entirely incidental that we observe improvements on these too. These benchmarks still combine simpler forms of reasoning as part of the main task, although reasoning is not the main skill to test. Some of them even include simpler math problems and constraint satisfaction queries. It is however encouraging to see that \model and \modelRL bring well-rounded, general improvements on a highly diverse set of measurements.

\subsection{Safety Evaluation}

We developed \model in accordance with Microsoft’s Responsible AI principles\footnote{https://www.microsoft.com/en-us/ai/principles-and-approach}. To assess model safety, we used the following benchmarks:

Automated RAI Measurement Framework: We used a framework
for automated measurement of Responsible AI metrics for LLMs~\cite{magooda2023frameworkautomatedmeasurementresponsible}. In this framework, one LLM poses as a user and engages in a synthetic conversation with the LLM under test. Another LLM then
evaluates the responses to measure its tendency to violate Responsible AI guidelines for the following categories: (1) How many times the user succeeded in Jailbreaking the model
under test? (2) How many times the model under test generates Potentially Harmful
Content? (3) How many times the model leaks Intellectual Property (IP)? These categories are divided into
multiple sub-categories described in~\cite{magooda2023frameworkautomatedmeasurementresponsible}. Results show that \model shows minor regression compared to the base model \phifour.

Toxigen: Toxigen \citep{toxigen} is a large-scale dataset consisting of toxic and neutral statements about 13 groups of people with a focus on implicit hate speech about minority groups. The dataset is balanced with an equal number of toxic and neutral prompts for each identity group. This allows us to evaluate erasure as well as toxicity detection: that is where the identity mention of specific groups of people is treated as a signal of toxicity or hate. Results indicate that as we add reasoning capabilities to models, we are able to improve performance on only one of the toxic or neutral categories at a time, indicating that detecting toxicity without causing erasure is still a challenging problem for all models. Nevertheless, we still observe modest improvements even in this task, with \model offering a better balance (than \modelRL and \phifour) between toxic vs. neutral content detection accuracy. Lower imbalance is also a preferable behavior for applications such as content filtering and moderation. Figure~\ref{fig:toxigen_d_model_comp} in the appendix shows a more detailed view of model accuracy on Toxigen, reported for 13 different demographic groups. Improvements from \model show that the model has also narrowed some of the group-based discrepancies, previously observed in \phifour. This is a positive direction towards increased group-based fairness. Both models also improve upon the \phifour model.

Despite the above measurements, appropriate evaluation of output generated by reasoning models remains a challenge mainly due to the fact that current LLM judges and tools have not yet been optimized for long, step-by-step traces and may struggle with the extended length and the non-linear generations. For example, reasoning models may tend to repeat the problem, create counterfactuals of the problem statement or edge cases. In this process, judges may falsely trigger safety measures or otherwise miss important assumptions and biases that models might have due to the entangled nature of the language~\citep{jiang2025safechain}. Future research is needed to mature the practices in these areas by potentially improving safety-oriented verification methods and by decomposing and simplifying the evaluation of long traces.

Finally, the above measurements focus on measuring engagement and refusal in jailbreak scenarios and on classification-style toxic language detection. We acknowledge that it is also important to evaluate models, including ours, in more open-form and benign scenarios, beyond jailbreaks and adversarial scenarios. These evaluations would enable a better understanding of how biases and harms may occur in benign, real-world tasks.
\section{Limitations}
\model inherits limitations from its base model. For example, the \phifour model primarily supports English text, with performance declining for other languages and less represented English varieties compared to standard American English. Despite safety measures, the model may still perpetuate stereotypes, misrepresent groups, or produce inappropriate content due to biases in training data. Additionally, as we show in our evaluations on general-purpose benchmarks for factuality, the model can generate inaccurate or outdated information that sounds plausible. For coding, \phifour is mainly trained on Python using common packages, and users should manually verify API uses if scripts involve other languages or packages.

Additionally, \model also exhibits common limitations of other reasoning language models such as requiring more computational time and resources due to their reasoning processes, resulting in slower response times compared conventional LLMs. Additionally, reasoning models may produce responses that contradict their own reasoning chains, potentially leading to inaccuracies. The ability to understand and monitor the reasoning steps for more transparency is still an active area of research. 

The \model model, while powerful, also has notable limitations, particularly with its context length of 32k tokens. This constraint can be limiting for more complex tasks that require extensive context to generate accurate and coherent responses. Additionally, the model sometimes generates responses that exceed its context window size, leading to truncation and loss of important information. This limited context window size also impacts the model's performance in multiturn interactions, as it may struggle to maintain continuity over extended conversations. Furthermore, the supervised fine-tuning (SFT) training data is limited to STEM, code, and safety, while the reinforcement learning (RL) data is limited to math. Although there are signs of generalization to other domains, this limitation may affect the model's performance on different  contexts.
\section*{Author Contributions}
\textbf{Data and Supervised Finetuning}: Mojan Javaheripi, Arindam Mitra, Sahaj Agarwal, Caio César Teodoro Mendes, Olli Saarikivi, Marah Abdin and Suriya Gunasekar  \\
\textbf{Reinforcement Learning}: Yue Wu, Harkirat Behl, Guoqing Zheng, Vaish Shrivastava and Dimitirs Papailiopoulos \\
\textbf{Evaluation and Analysis}: Vidhisha Balachandran, Lingjiao Chen, Neel  Joshi,  Vibhav Vineet, Safoora Yousefi and Besmira Nushi \\
\textbf{Infrastructure and Release}: Yash Lara, Gustavo de Rosa, Piero Kauffmann and Shital Shah \\
\textbf{Project Lead}: Ahmed Awadallah

\section*{Acknowledgements}
We would like to thank Eric Price, Shivam Garg, John Langford for discussion and feedback and Ece Kamar, Rafah Hosn, Doug Burger, Peter Lee, Asha Sharma, Chelsea Carter, Rob Young and Weishung Liu for their support throughout the project.
\bibliographystyle{plainnat} %
{
\small
\bibliography{custom}
}
\clearpage
\appendix
\section{Benchmarking Details}
\begin{table}[h]
  \centering
  \footnotesize
    \begin{tabular}{lccc}
    \toprule
    \bfseries{Model} & \bfseries{temp.} & \bfseries{max token} & \bfseries{reasoning} \\
    \hline
    \phifour~\citep{abdin2024phi4technicalreport} & 0.8$^\dagger$     & 4,096 & n \\
    \model &  0.8      & 32,768$^*$ & y \\
    \modelRL &  0.8      & 32,768$^*$ & y \\
    \hline
    \ronedistill \citep{guo2025deepseek} & 0.6   & 65,536 & y \\
    \rone \citep{guo2025deepseek} & 0.6   & 65,536 & y \\
    \hline
    \gptfouro \textsf{2024-05-13} \citep{hurst2024gpt} & 1.0     & 4,096  & n \\
    \oonemini \textsf{2024-09-12}~\cite{jaech2024openai} & NA & NA & y\\
    \oone\xspace \textsf{2024-12-17} \citep{jaech2024openai}& NA    & NA    & y \\
    \othree\xspace \textsf{2025-01-31(high)} \citep{O3mini} & NA    & NA    & y \\
    \hline
    \claude \textsf{2025-02-19} \citep{Claude37Sonnet} & 1.0     & 32,768 & y \\
    \hline
    \geminiflash \textsf{Exp 2025-01-21} \citep{GeminiFlash}& 1.0     & 32,768 & y \\
    \bottomrule
    \end{tabular}%
     \caption{List of models studied in this report and corresponding temperature and maximum token limits used for all experiments. $^*$For \model and \modelRL evaluations on AIME, HMMT, GPQA, and Codeforces we use 65,536 as the maximum number of tokens for generation without changing any RoPE parameters. We note that neither model has properly trained on this length. All other evaluations for \model and \modelRL use 32,768. $^{\dagger}$For \phifour we use temp=0.8 for the reasoning benchmarks, and 0.0 for the general-purpose benchmarks.}
  \label{tab:models_repro}%
\end{table}%
\begin{table}[h]
  \footnotesize
    \centering
    \begin{tabular}{ll}%
    \toprule
    \bfseries{Dataset (\#prompts)} & \bfseries{Link} \\
    \hline
    AIME 25~\citep{AIME25} (30) &   \url{https://huggingface.co/datasets/lchen001/AIME2025}  \\
    \hline
    AIME 83-24~\citep{AIME8324} (949) & \url{https://huggingface.co/datasets/lchen001/AIME1983_2024} \\
    \hline
    HMMT February 2025 (30) & \url{https://huggingface.co/datasets/MathArena/hmmt_feb_2025} \\
    \hline
    Omni-MATH~\citep{gao2024omni} (4428) & \url{https://huggingface.co/datasets/KbsdJames/Omni-MATH}     \\
    \hline
    GPQA$\mathbin{\Diamond}$~\citep{rein2024gpqa} (198) & \url{https://huggingface.co/datasets/Idavidrein/gpqa} \\
    \hline
    BA-Calendar~\citep{butt2024benchagents} (2000) & \url{https://huggingface.co/datasets/microsoft/ba-calendar}    \\
    \hline
    TSP-Opt (new benchmark) (960) & To be released    \\
    \hline
    3SAT-Search (new benchmark) (800) & To be released    \\
    \hline
    Maze~\citep{wang2024picture} (1500) & \url{https://huggingface.co/datasets/microsoft/VISION_LANGUAGE}  \\
    \hline
    SpatialMap~\citep{wang2024picture} (1500) & \url{https://huggingface.co/datasets/microsoft/VISION_LANGUAGE}  \\
    \bottomrule
    \end{tabular}%
      \caption{List of reasoning benchmarks used in this report and where to find them.}
  \label{tab:datasets_reasoning_repro}%
\end{table}%
\newpage
\section{Additional Results}
\begin{figure*}[h!]    
     \begin{subfigure}[b]{\textwidth}    
     \centering
    \includegraphics[width=0.47\linewidth]{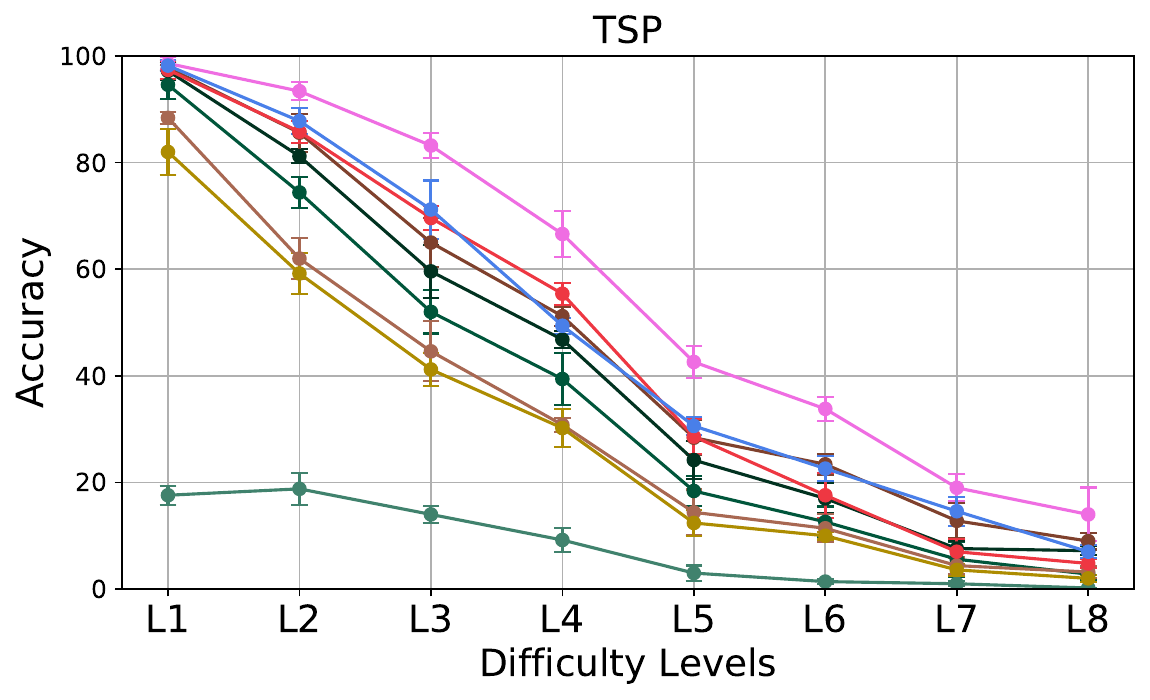}
    \includegraphics[width=0.47\linewidth]{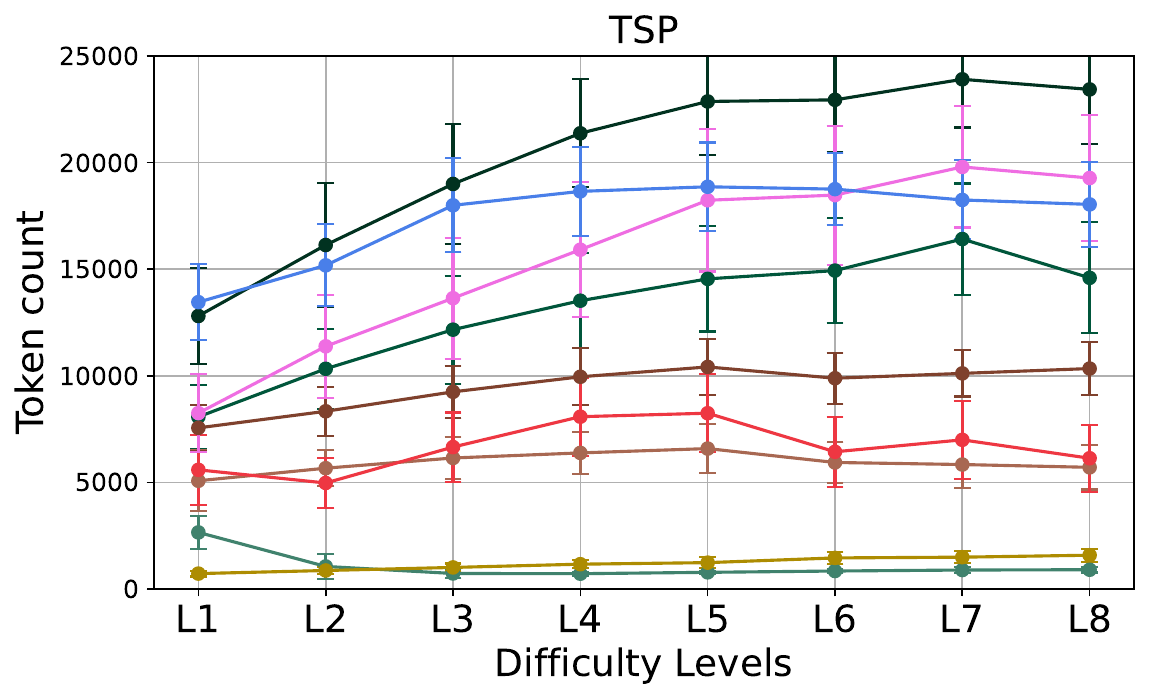}
    \end{subfigure}
    \begin{subfigure}[b]{\textwidth}
        \centering
        \includegraphics[width=0.47\textwidth]{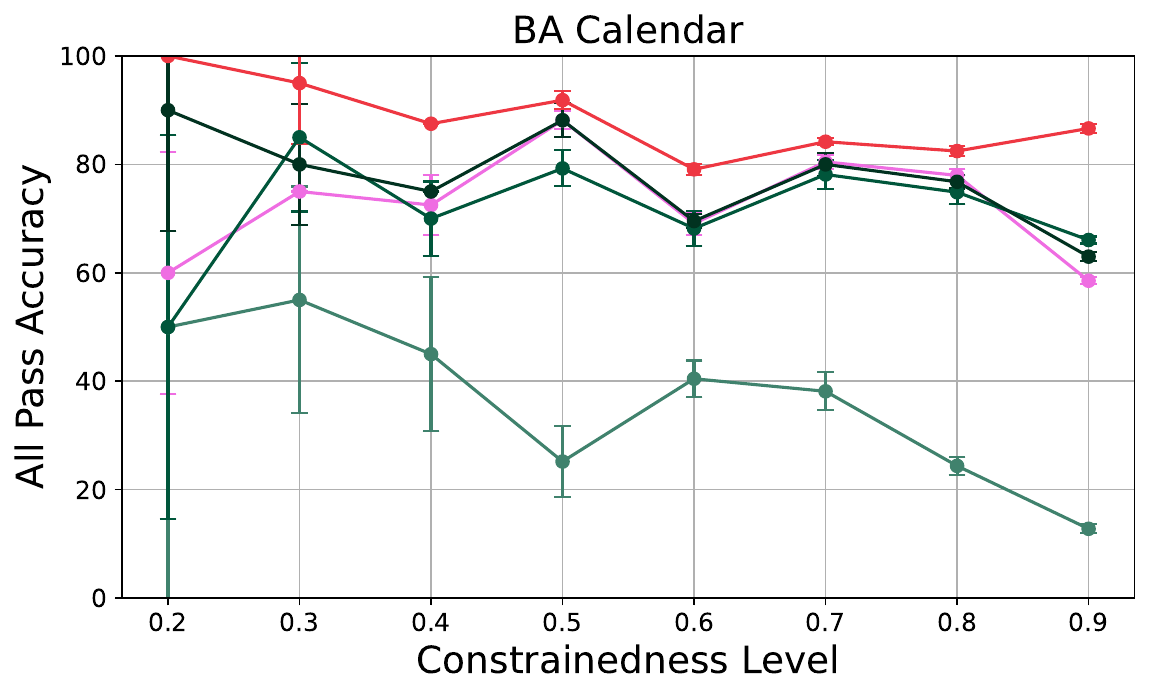}
        \includegraphics[width=0.47\textwidth]{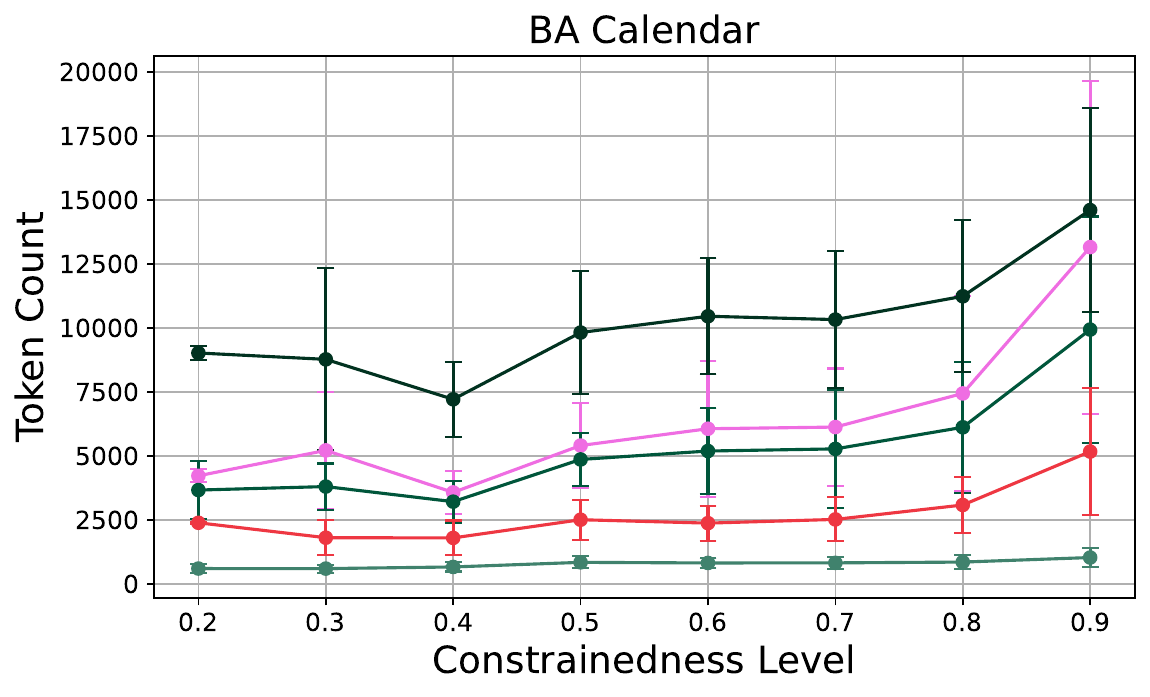}
    \end{subfigure}
    \begin{subfigure}[b]{\textwidth}
        \centering
        \includegraphics[width=0.47\textwidth]{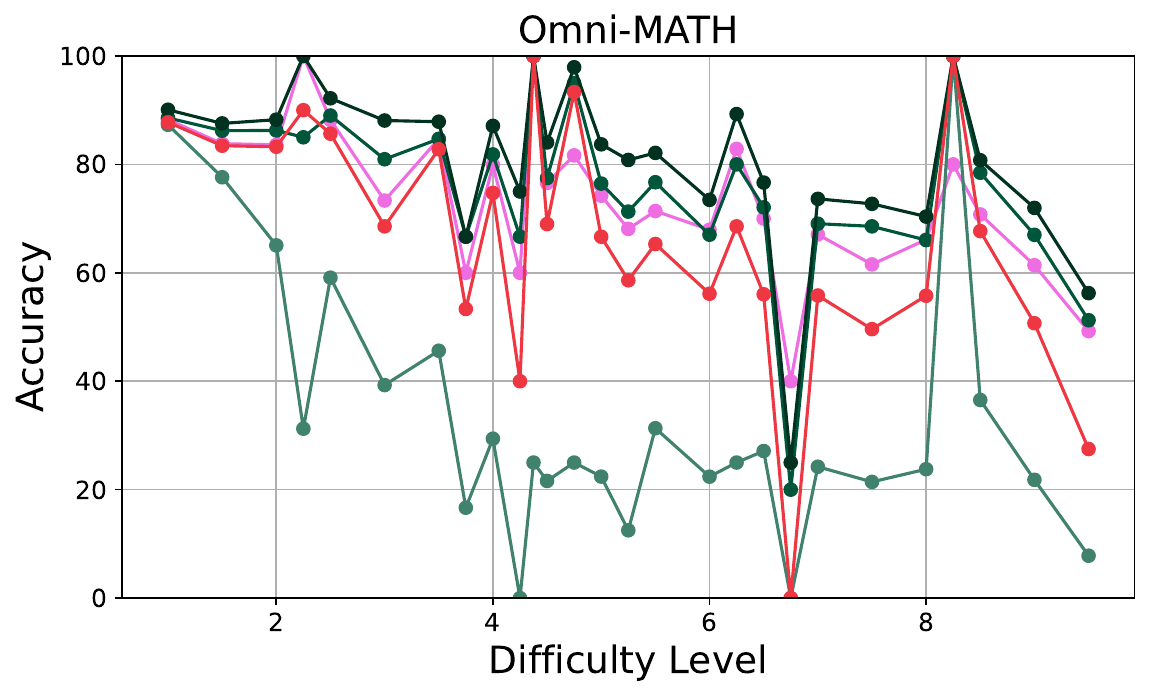}
        \includegraphics[width=0.47\textwidth]{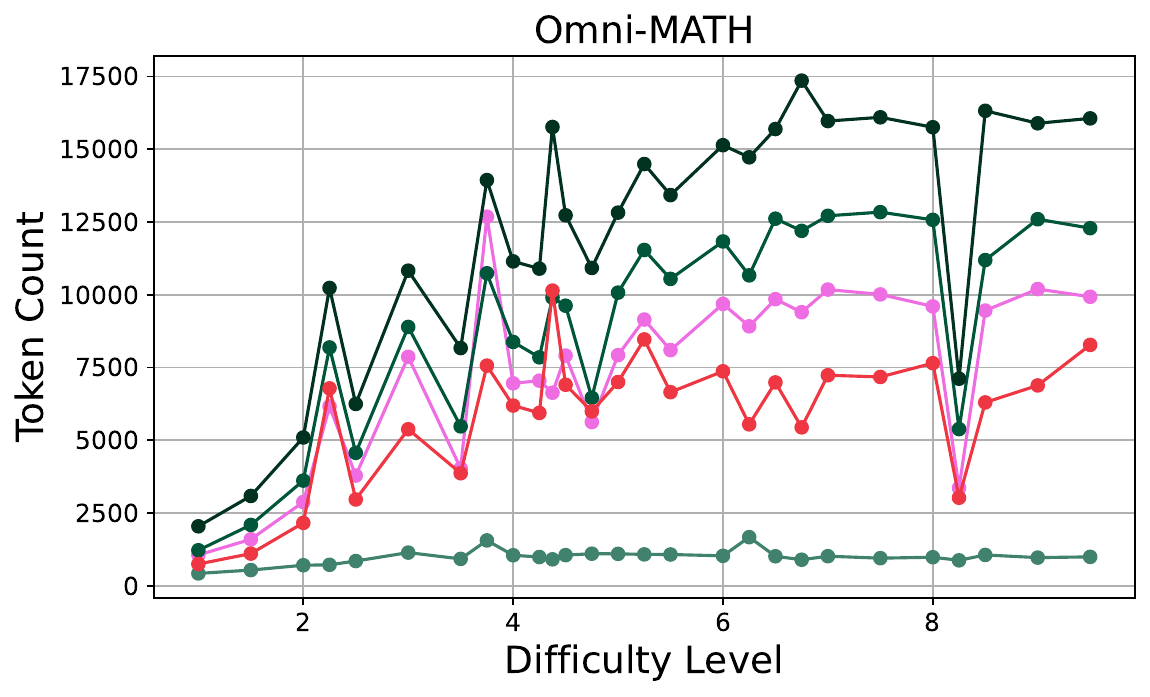}
    \end{subfigure}
        \begin{subfigure}[b]
    {\textwidth}
        \centering
\includegraphics[width=\textwidth]{figures/model_legend.png}
    \end{subfigure}
    \caption{TSP, BA-Calendar, and Omni-Math accuracy and token usage with difficulty levels. Standard deviation for token usage is computed across different parallel repeats. For Omn-Math, we exclude error bars and some of the models to improve readability. Note the difficult tags in Omni-MATH can be nosiy and imperfect, and the size of the data available per level is imbalanced. This leads to fluctuations between adjacent difficulty levels. Similarly, we do not show all models for the BA-Calendar charts.}
    \vspace{-2pt}
    \label{fig:tsp__calendar_omnimath_token_accuracy_difficulty}
\end{figure*}

\begin{figure}[t]
    \centering
    \includegraphics[width=0.8\textwidth]{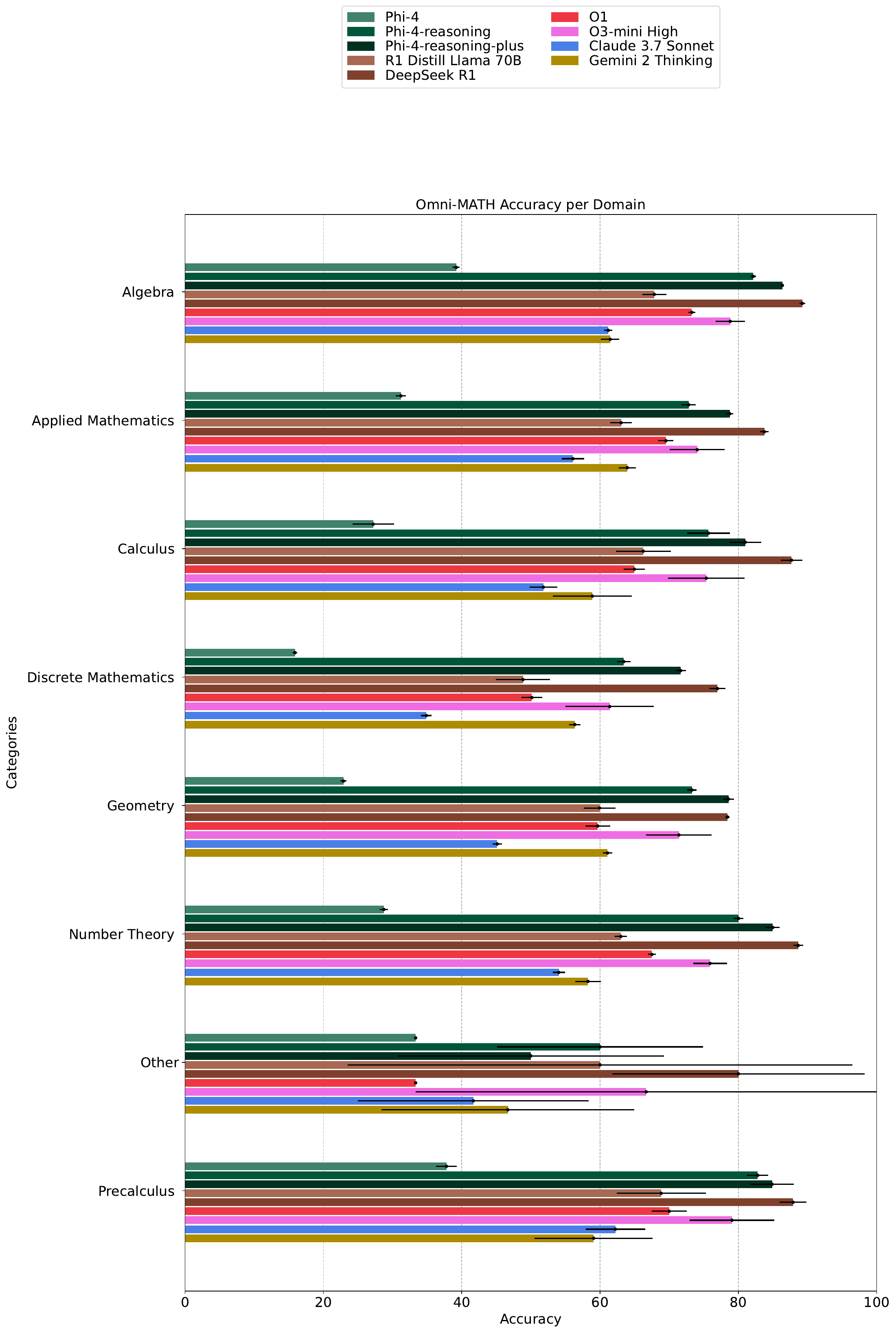}
    \caption{Omni-MATH topic-level accuracy. While general model trends are consistent across different topics, all models display lower performance on problems related to discrete mathematics and geometry.}
    \label{fig:omni_math_const_acc}
\end{figure}

\begin{figure}[t]
    \centering
    \includegraphics[width=0.45\textwidth]{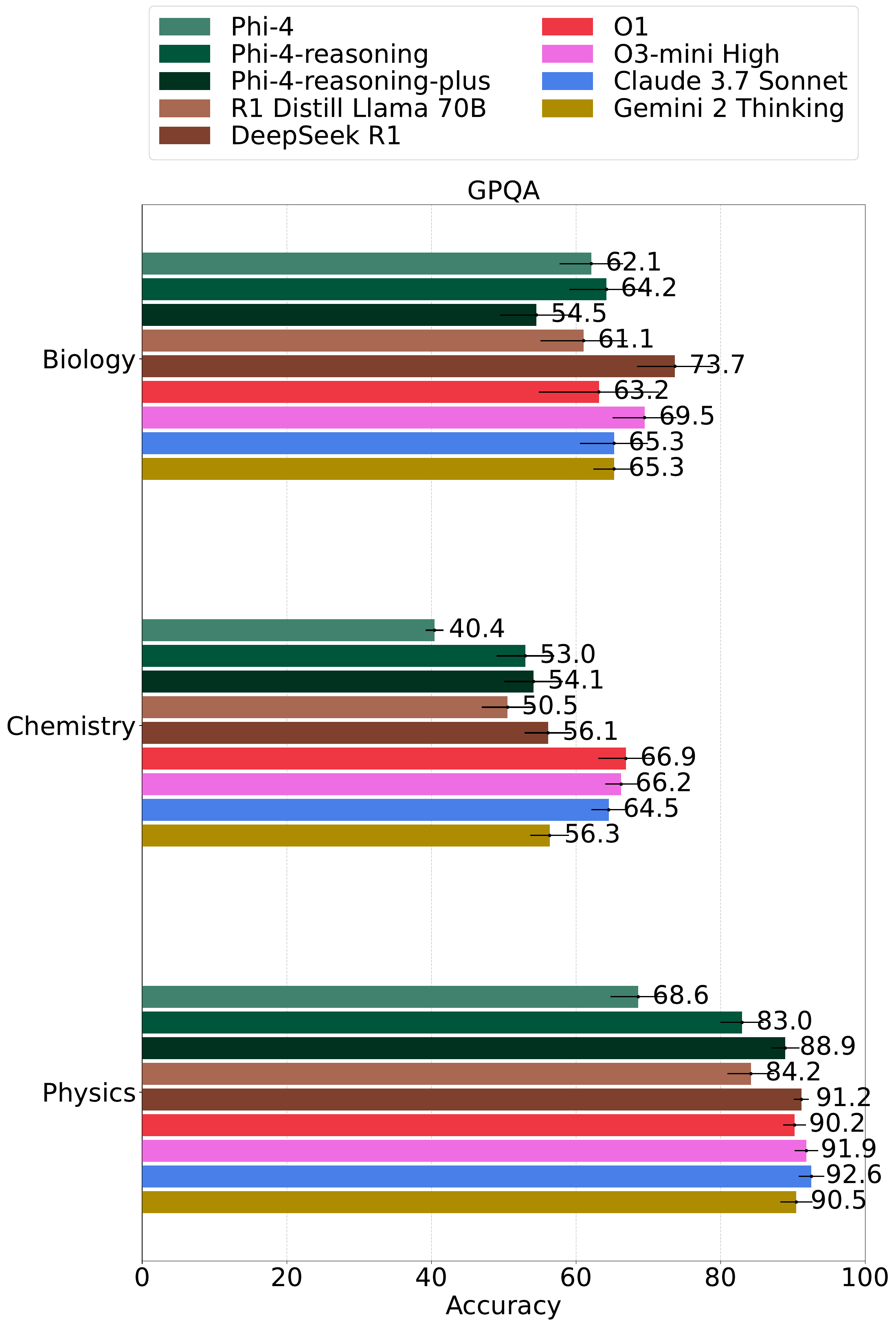}
    \includegraphics[width=0.45\textwidth]{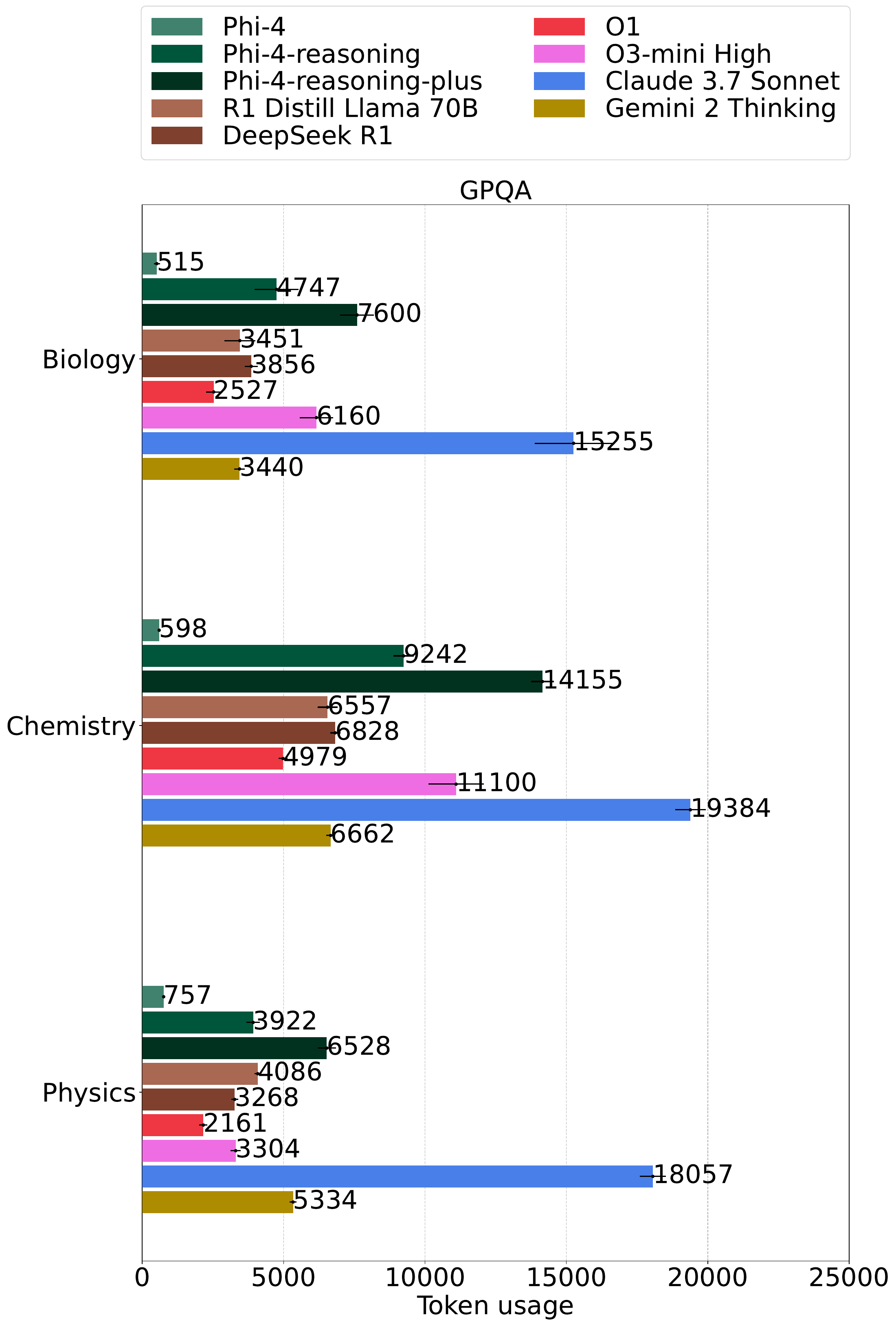}
    \caption{GPQA accuracy and token usage by high-level domain. Standard deviations for token usage are computed across five repeats, within the same high-level domain. Improvements for all models on biology and chemistry are lower than on physics.}
    \label{fig:gpqa_const_acc}
\end{figure}

\begin{figure}[t]
    \centering
     \begin{subfigure}[b]{0.48\textwidth}
     \centering
         \includegraphics[width=\textwidth]{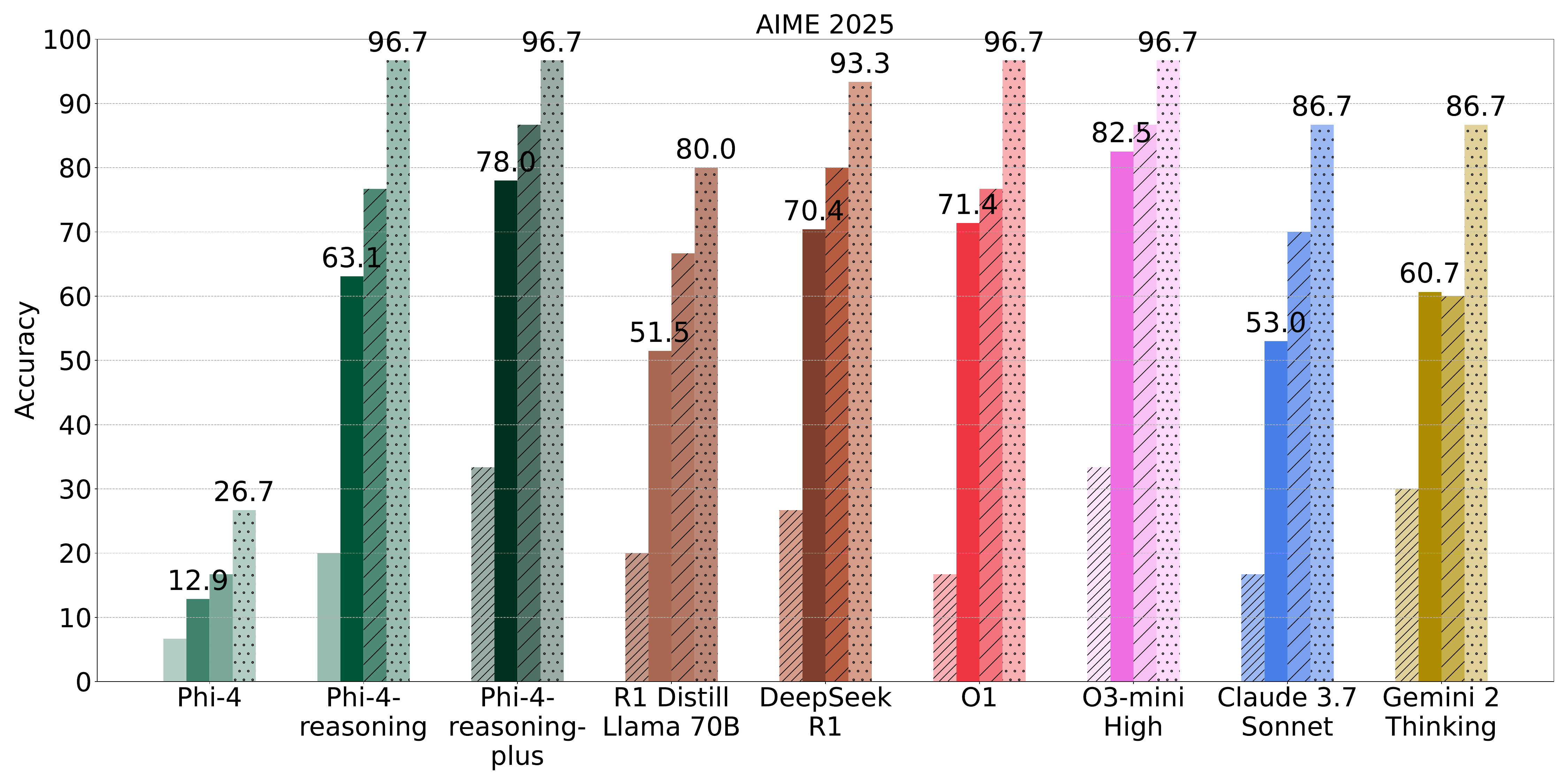}
     \end{subfigure}
        \begin{subfigure}[b]{0.48\textwidth}
        \centering
        \includegraphics[width=\textwidth]{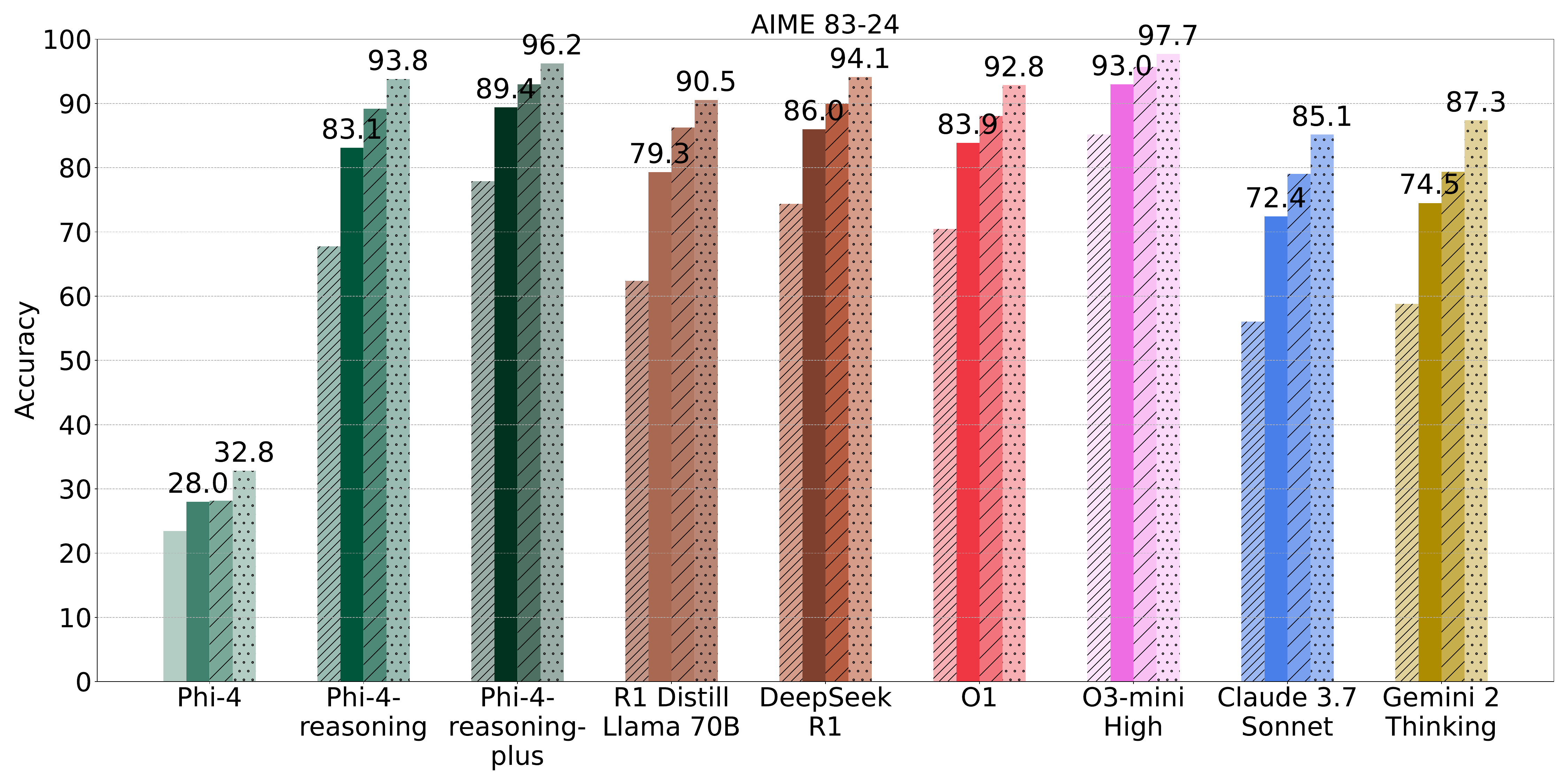}
    \end{subfigure}
    \begin{subfigure}[b]{0.48\textwidth}
        \centering
        \includegraphics[width=\textwidth]{figures/GPQA_charts/inf_analysis_report/ExactMatch_result_inference_comp_accuracy_bar_chart.pdf}
    \end{subfigure}
    \begin{subfigure}[b]{0.48\textwidth}
        \centering
        \includegraphics[width=\textwidth]{figures/Omni_Math/inf_analysis_report/OmniMath_correctness_inference_comp_accuracy_bar_chart.pdf}
    \end{subfigure}
    \begin{subfigure}[b]{0.48\textwidth}
        \centering
        \includegraphics[width=\textwidth]{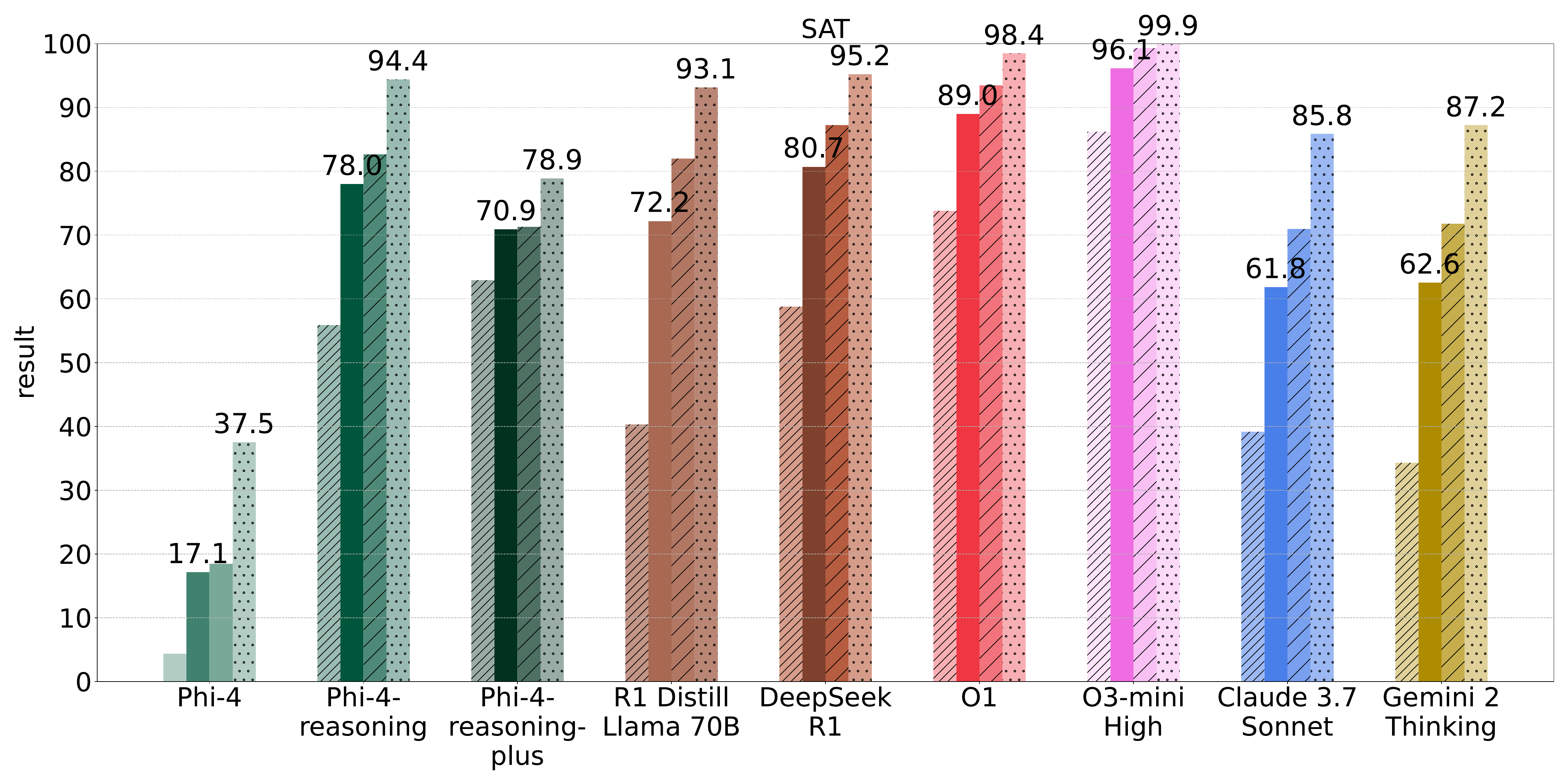}
    \end{subfigure}
    \begin{subfigure}[b]{0.48\textwidth}
        \centering
        \includegraphics[width=\textwidth]{figures/TSP_Analysis/bon_analysis_report/NPHardTSPMetric_comp_accuracy_bar_chart.pdf}
    \end{subfigure}
    \begin{subfigure}[b]{0.48\textwidth}
        \centering
        \includegraphics[width=\textwidth]{figures/BA_Calendar/inf_analysis_report/all_correct_inference_comp_accuracy.pdf}
    \end{subfigure}
    \begin{subfigure}[b]{0.48\textwidth}
        \centering
        \includegraphics[width=\textwidth]{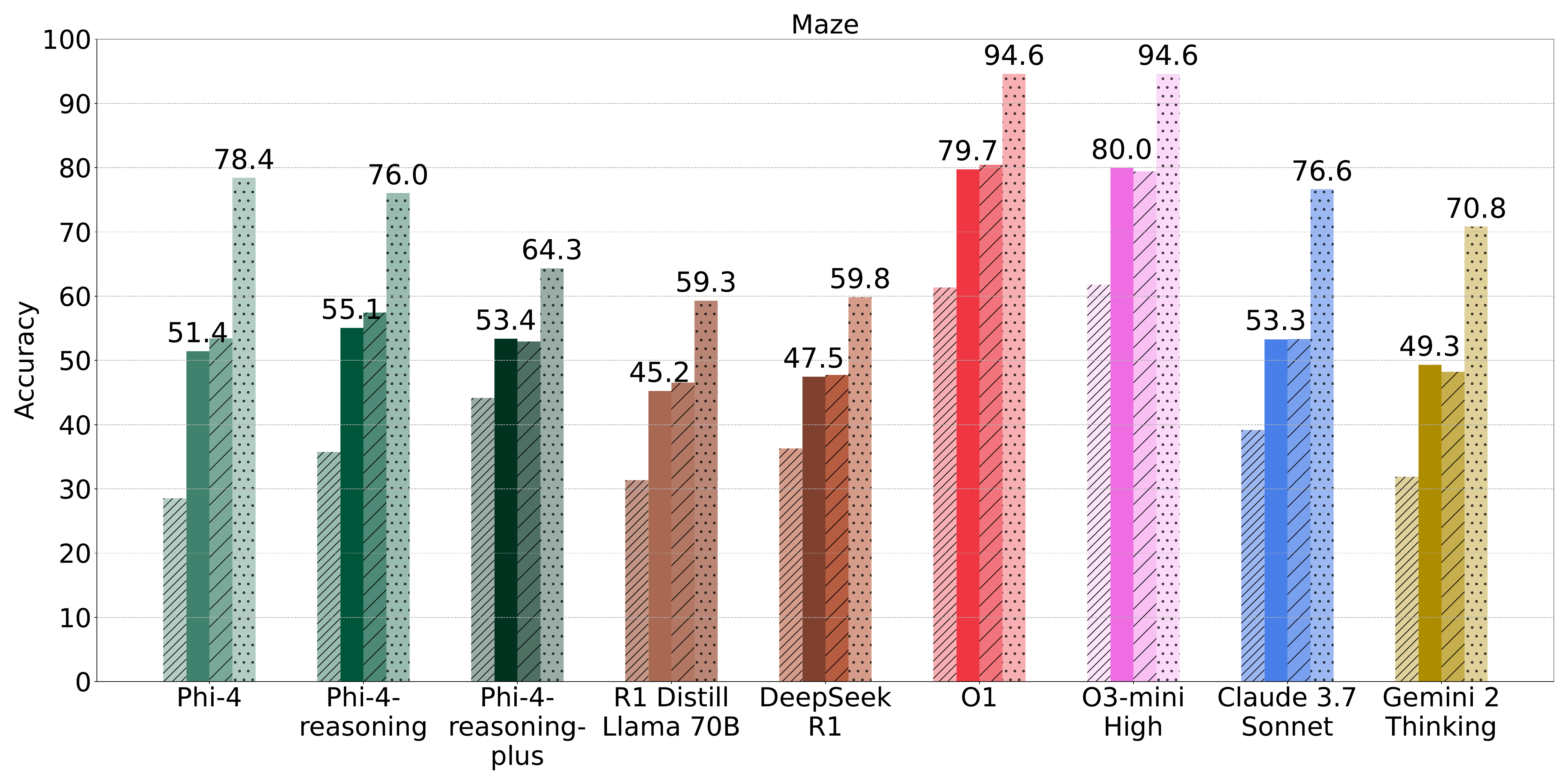}
    \end{subfigure} 
    \begin{subfigure}[b]{0.48\textwidth}
        \centering
        \includegraphics[width=\textwidth]{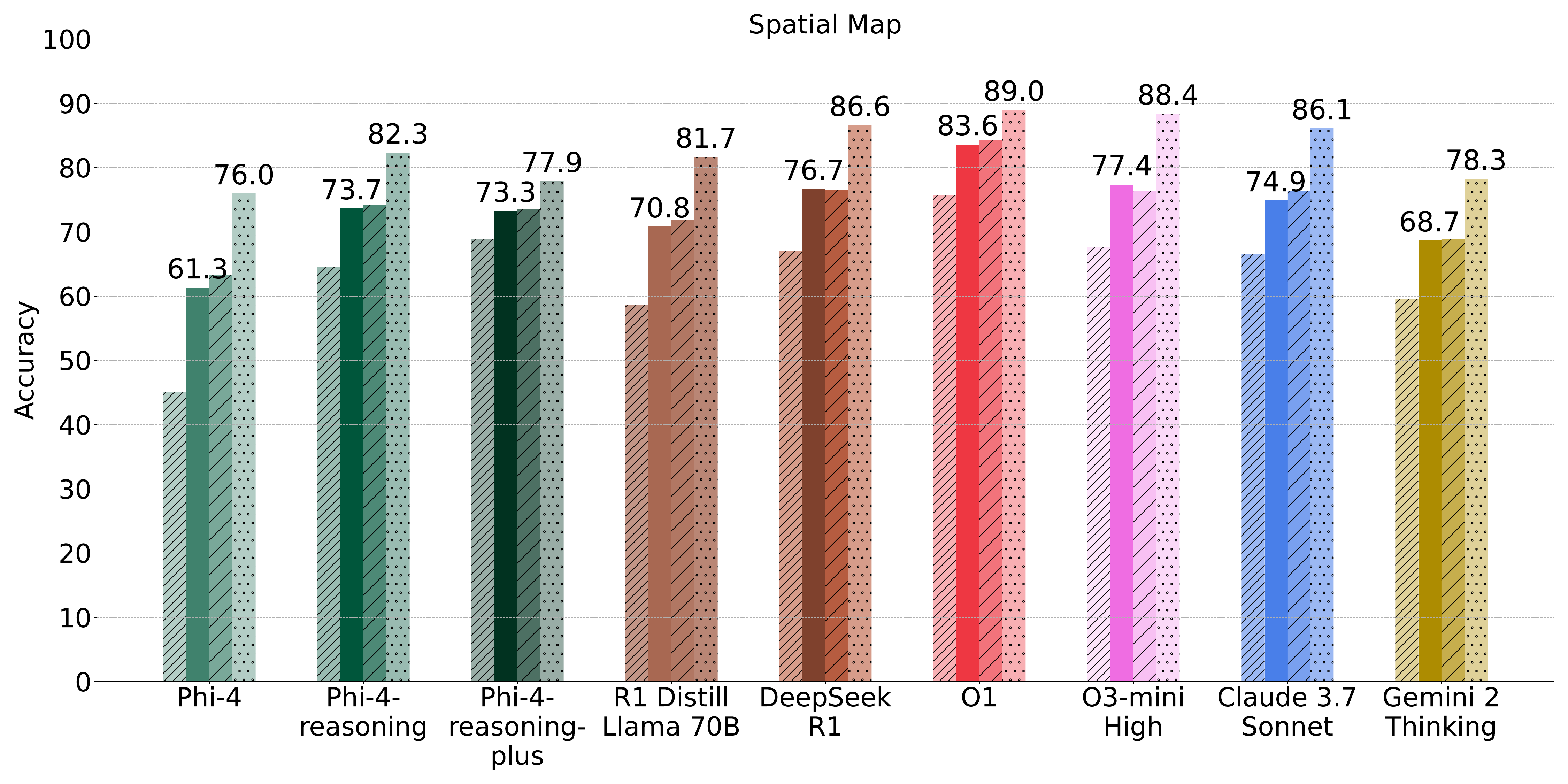}
    \end{subfigure}    
    \vspace{-15pt}
    \begin{subfigure}[b]
    {\textwidth}
        \centering
\includegraphics[width=0.5\textwidth]{figures/agg_legend.png}
        \label{fig:9}
    \end{subfigure}
    \vspace{-5pt}
    \caption{Results on reasoning benchmarks with different aggregations on 5 independent runs: worst of 5, average pass@1, majority vote, and best of 5. Across models and benchmarks we observe improvement in performance with best-of-5 indicating further room for improvement.}
    \label{fig:main_inf_scaling}
\end{figure}

\begin{figure}[t]
    \centering
    \begin{subfigure}[b]{0.38\textwidth}
\includegraphics[width=0.75\linewidth]{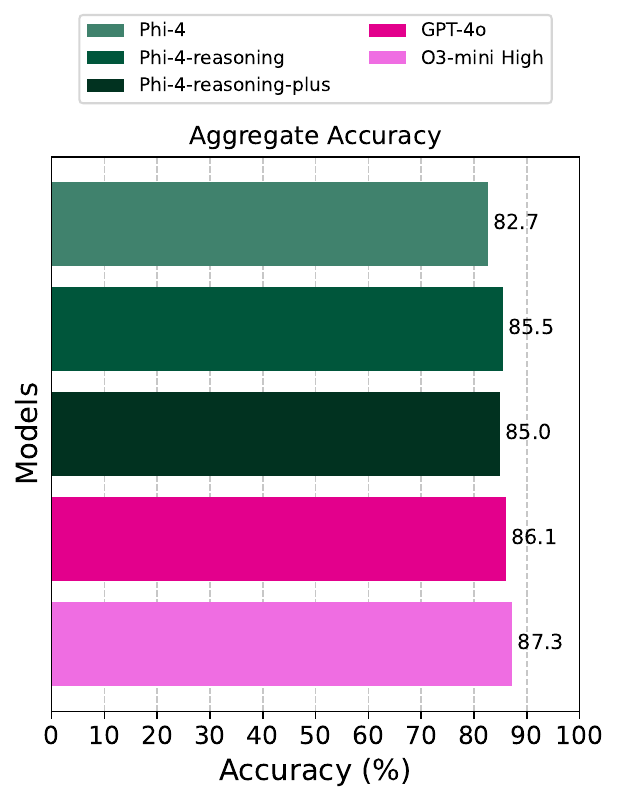}\\
    \includegraphics[width=0.75\linewidth]{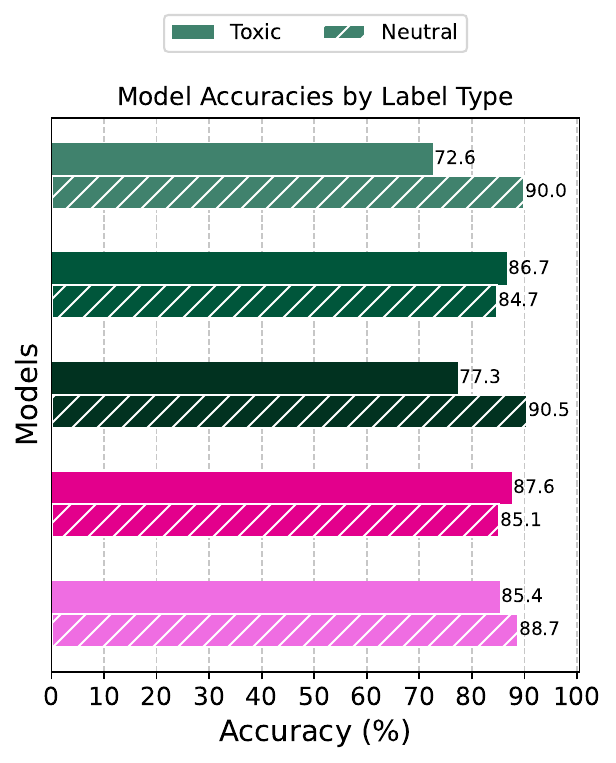}
    \caption{Toxigen discriminative evaluation across different models.
    \label{fig:toxigen_d_aggregate}}
    \end{subfigure}\quad
\begin{subfigure}[b]{0.57\textwidth}
    \centering
    \includegraphics[width=\linewidth]{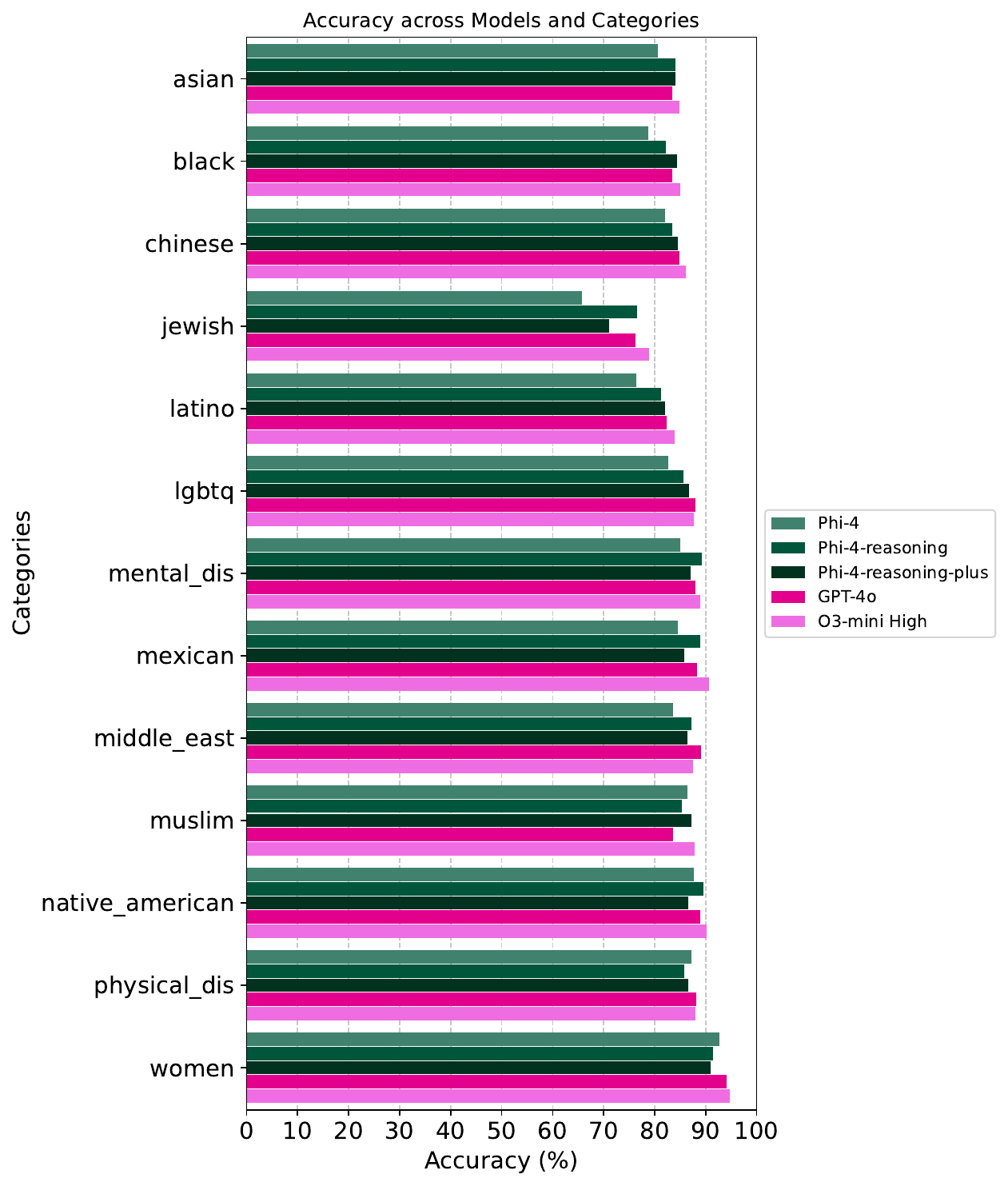}
    \caption{Model comparison across different categories in the discriminative evaluation setting of Toxigen.
    \label{fig:toxigen_d_model_comp}}
    \end{subfigure}
    \caption{Fine grained results on Toxigen. We observe modest improvements in this task of detecting toxic and neutral text, with \model offering a better balance between toxic vs. neutral content detection accuracy. Improvements from \model  has also narrowed some of the group-based discrepancies, previously observed in \phifour indicating increased group-based fairness.}
\end{figure}

\end{document}